%% file: main.tex
\documentclass[11pt]{article}
\pdfoutput=1

\usepackage{microtype}
\usepackage{graphicx}
\usepackage{booktabs} %
\usepackage{amssymb}
\usepackage{bm}
\usepackage{bbm}
\usepackage{amsmath}  %
\usepackage{amsthm}
\usepackage{xcolor}
\usepackage{textcomp}
\usepackage{multirow}
\usepackage{mathtools}
\usepackage[margin=1in]{geometry}
\usepackage{authblk}
\usepackage[numbers,sort&compress]{natbib}
\usepackage{xspace}
\usepackage{comment}
\usepackage{caption}
\usepackage{subcaption}
\usepackage{thmtools, thm-restate}  %
\usepackage{verbatim}
\usepackage{arydshln}
\usepackage[hang, flushmargin]{footmisc}

\usepackage{tikz}
\usetikzlibrary{arrows}
\usetikzlibrary{positioning}

\tikzset{
  treenode/.style = {align=center, inner sep=0pt, text centered,
    font=\sffamily},
  arn_n/.style = {treenode, circle, black, font=\sffamily\bfseries, draw=black,
    fill=white, text width=1.5em},%
  arn_r/.style = {treenode, circle, black, font=\sffamily\bfseries, draw=black,
    fill=white, text width=1.0em},%
  arn_x/.style = {treenode, rectangle, draw=black,
    minimum width=0.5em, minimum height=0.5em}%
}

\usepackage{hyperref}
\usepackage[capitalize,noabbrev]{cleveref}

\usepackage{algorithm}
\usepackage{algpseudocode}

\usepackage{bm}
\usepackage{enumitem}

\newcommand{\sysname}{\textsc{Parcae}\xspace}

\usepackage{pifont}
\newcommand{\cmark}{\ding{51}}
\newcommand{\xmark}{\ding{55}}

\newcommand*{\ShowNotes}{}

\input{macros.tex}

\newif\ifarxiv

\title{Parcae: Scaling Laws For Stable Looped Language Models}

\author[1,2]{Hayden Prairie}
\author[1]{Zachary Novack}
\author[1]{Taylor Berg-Kirkpatrick}
\author[1,2]{Daniel Y. Fu}

\affil[1]{University of California, San Diego}
\affil[2]{Together AI}

\usepackage{titling}
\date{}

\begin{document}

\setlength{\parindent}{0pt}
\setlength{\parskip}{0.5em}  %

\maketitle

\vspace{-5.5em}  %
\begin{center}
\texttt{\{hprairie,znovack,tberg,danfu\}@ucsd.edu}
\end{center}

\vspace{-1em}
\begin{abstract}
  \input{sections/abstract}
\end{abstract}

\input{sections/introduction}

\input{sections/background}

\input{sections/methodology}

\input{sections/parcae}

\input{sections/results}

\input{sections/conclusion}

\bibliographystyle{plainnat}
\bibliography{main}

\appendix

\input{appendix/notation}

\input{appendix/discussion}
\input{appendix/derivation}
\input{appendix/flop-estimate}
\input{appendix/training-algorithm}
\input{appendix/stability}
\input{appendix/loss-spikes}
\input{appendix/truncated}
\input{appendix/choosing}
\input{appendix/prelude-norm}
\input{appendix/train-parametric}
\input{appendix/test-parametric}
\input{appendix/eval-setup}
\input{appendix/fixed-expanded-results}
\input{appendix/scaling-laws-setup}
\input{appendix/definitions}
\input{appendix/hyperparameters}
\input{appendix/tokenizer}

\end{document}

%% file: macros.tex
\ifdefined\ShowNotes
  \newcommand{\colornote}[3]{{\color{#1}\bf{#2 #3}\normalfont}}
\else
  \newcommand{\colornote}[3]{}
\fi

\definecolor{darkred}{rgb}{0.7,0.1,0.1}
\definecolor{darkgreen}{rgb}{0.1,0.5,0.1}
\definecolor{cyan}{rgb}{0.7,0.0,0.7}
\definecolor{dblue}{rgb}{0.2,0.2,0.8}
\definecolor{maroon}{rgb}{0.76,.13,.28}
\definecolor{burntorange}{rgb}{0.81,.33,0}
\definecolor{royalpurple}{rgb}{0.47,.31,0.66}

\ifdefined\ShowNotes
  
\else
  
\fi

\newcommand{\meanrecurrence}{$\mu_{\text{rec}}$\xspace}
\newcommand{\meanbackward}{$\mu_{\text{bwd}}$\xspace}
\newcommand{\prelude}{$\mathcal{P}$\xspace}
\newcommand{\recurrent}{$\mathcal{R}$\xspace}
\newcommand{\coda}{$\mathcal{C}$\xspace}
\newcommand{\dt}{\Delta}
\newcommand{\A}{\bm{A}}
\newcommand{\B}{\bm{B}}
\newcommand{\C}{\bm{C}}

\newcommand{\dA}{\overline{\bm{A}}}
\newcommand{\dB}{\overline{\bm{B}}}

\newcommand{\R}{\mathbb{R}}

%% file: sections/abstract.tex
\noindent Traditional fixed-depth architectures scale quality by increasing training FLOPs, typically through increased parameterization, at the expense of a higher memory footprint, or data. 
A potential alternative is \emph{looped architectures}, which instead increase FLOPs by sending activations through a block of layers in a loop.
While promising, existing recipes for training looped architectures can be unstable, suffering from residual explosion and loss spikes.
We address these challenges by recasting looping as a nonlinear time-variant dynamical system over the residual stream. Via a linear approximation to this system, we find that instability occurs in existing looped architectures as a result of large spectral norms in their injection parameters.
To address these instability issues, we propose \emph{\textbf{Parcae}}, a novel \emph{stable}, looped architecture that constrains the spectral norm of the injection parameters via discretization of a negative diagonal parameterization. As a result, Parcae achieves up to 6.3\% lower validation perplexity over prior large-scale looped models.
Using our stable looped architecture, we investigate the scaling properties of looping as a medium to improve quality by increasing FLOPs in training and test-time.
For training, we derive predictable power laws to scale FLOPs while keeping parameter count fixed. Our initial scaling laws suggest that looping and data should be increased in tandem, given a fixed FLOP budget.
At test-time, we find that Parcae can use looping to scale compute, following a predictable, saturating exponential decay. 
When scaled up to 1.3B parameters, we find that Parcae improves CORE and Core-Extended quality by 2.99 and 1.18 points when compared to strong Transformer baselines under a fixed parameter and data budget, achieving a relative quality of up to 87.5\% a Transformer twice the size.

%% file: sections/introduction.tex
\section{Introduction}
\label{sec:intro}
Scaling laws have established that model performance improves predictably with increased FLOPs \citep{kaplan2020scalinglawsneurallanguage, hoffmann2022trainingcomputeoptimallargelanguage}, typically by increasing parameter count or training data.
These scaling laws suggest that FLOP-optimal training increases parameters and training data in tandem following empirical power laws.
As a result, the depth and width of state-of-the-art models have grown in an effort to scale with data, subsequently inflating the memory footprint to deploy these models \citep{dettmers2023case4bitprecisionkbit, lin2024awqactivationawareweightquantization}.

However, as inference deployments take on an increasingly large portion of compute \citep{touvron2023llamaopenefficientfoundation}, and deployments begin to move to the edge \citep{moon2024lpulatencyoptimizedhighlyscalable, narayan2025minionscostefficientcollaborationondevice}, there is increasing interest in scaling model quality without increasing parameters.
One mechanism to do this is layer-looped models, such as looped transformers \citep{dehghaniUniversalTransformers2019, geiping_scaling_2025, zhuScalingLatentReasoning2025}, which iteratively loop activations through a block of layers.
Initial results have been encouraging, with looped models matching the quality of larger fixed-depth architectures~\citep{geiping_scaling_2025, zhuScalingLatentReasoning2025}. Moreover, they show potential for latent reasoning~\citep{avi_learn_algorithm, yangLoopedTransformersAre2023} and per-token adaptive compute \citep{geiping_scaling_2025,mcleish_retrofitted_recurrence}.

\begin{figure*}[t]
    \centering
    \vspace{-0.3em}
    \includegraphics[trim={1.36cm 0 0 0cm},clip, width=\linewidth]{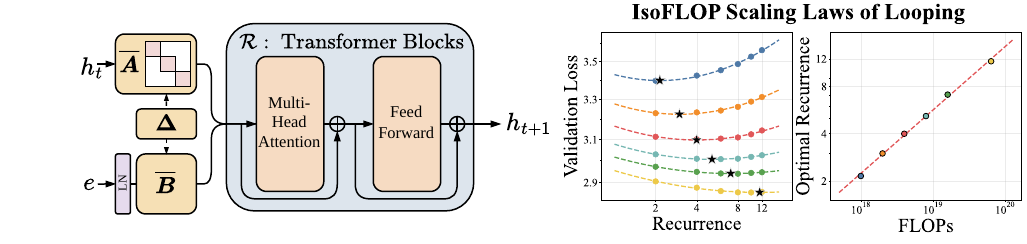}
    \vspace{-1.6em}
    \caption{\textbf{\sysname and the Scaling Laws of Looping.}
    (\emph{Left}) \sysname constrains the spectral norm of $\dA$ and normalizes the input injection, stabilizing the residual stream $h_t$ across loops. (\emph{Right}) We observe looping to be an orthogonal axis of scaling compute which follows a power law.}
    \label{fig:block-diagram}
    \vspace{-1em}
\end{figure*}

Unfortunately, prior research \citep{geiping_scaling_2025, mcleish_retrofitted_recurrence, LoopFormerElasticDepthLooped2025} and our work observe these models' training to be unstable, exhibiting residual state explosion and loss spikes.
Since these models loop the layers of complex non-linear architectures (e.g., transformer blocks \citep{vaswani2023attentionneed}), the source of instability in looped models can be difficult to understand analytically.
As a result, training requires sensitive hyperparameter selection and {residual normalization} (e.g., Post-Norm) to correct this instability \citep{geiping_scaling_2025}. 
Furthermore, even in convergent training runs, we observe loss spikes as looped models train on stochastic amounts of depth to induce stronger test-time scaling \citep{anilPathIndependentEquilibrium}.
In this paper, we study this instability and ask whether stabilizing these models can unlock looping as a predictable, orthogonal axis for scaling compute.

To analyze instability, we observe that prior looped architectures can be recast as a nonlinear time-variant dynamical system over the residual stream \citep{olsson2022incontextlearninginductionheads}, taking the form:
\begin{equation}
    \label{eq:alpha-af}
    h_{t+1} = \dA h_t + \dB e + \overline{\mathcal{R}}(h_t, e),
\end{equation}
where for an input $e$, the hidden state $h$ across the depth of an architecture is modulated by $\dA$, controlling the balance between prior and current residual states; $\dB$, conditioning the residual on the input $e$; and a non-linear operator $\overline{\mathcal{R}}$, which subsumes the original transformer modules (e.g., Attention, MLPs).
By linearizing this framework (e.g., removing $\overline{\mathcal{R}}$), we observe that \cref{eq:alpha-af} resolves to a linear time invariant (LTI) system from which classic control theory can be used to infer divergence conditions on the residual stream based on the spectral norm of $\dA$.
We observe that prior looped architectures can learn unstable parameterizations of $\dA$, which we empirically find to induce residual stream explosion (see \cref{fig:spectral-norm}).

To address these issues, we propose \emph{\textbf{\sysname}}, a novel looped transformer that corrects the parameter instability conditions of \cref{eq:alpha-af} and uses algorithmic fixes to reduce loss spikes during training. \emph{\textbf{\sysname}} explicitly uses discretization on a continuous representation $\A$ of \cref{eq:alpha-af} and parametrizes $\A$ as a negative diagonal matrix, constraining the spectral norm to prevent residual explosion in looped layers.
Additionally, \sysname introduces a normalization on $e$, which empirically prevents loss spikes in late stages of training. Finally, \sysname 
modifies the training algorithm (which aims to minimize the expected loss over variable depths) by enabling intra-batch per-sequence depth sampling to further reduce loss spikes.

We evaluate \sysname on end-to-end quality, training FLOP scaling, and test-time scaling:
\begin{itemize}[itemsep=0pt,topsep=0pt,leftmargin=*]
    \item\textbf{End-to-End Quality.} We compare \sysname against parameter- and data-matched RDMs \citep{geiping_scaling_2025} and Transformers. Against RDMs, \sysname reduces val. PPL by 6.3\%. When scaled up to 1.3B parameters and 100B tokens, \sysname outperforms parameter-matched Transformers by up to 2.99 and 1.18 points on Core and Core-Extended \citep{li2025datacomplmsearchgenerationtraining} benchmarks, respectively --- matching Transformers up to twice the size. 
    \item\textbf{Training FLOP Scaling.} To evaluate FLOP training scaling, we study scaling laws for looping in a parameter-matched isoFLOP setting (i.e., whether to scale FLOPs with increased data or looping).
    We find that looping introduces an orthogonal scaling axis, similar to parameters and data.
    Specifically, FLOP-optimal training increases looping and data following empirical power laws (see \cref{fig:block-diagram}~[\emph{right}]). 
    \item\textbf{Test-Time Scaling.} We study looping as a mechanism to scale test-time compute, observing that recurrence follows predictable exponential decay with an irreducible loss. We further combine both test-time and training power laws to create a single unifying scaling law for looping in \sysname models.
\end{itemize}

%% file: sections/background.tex
\section{Background}
\label{sec:background}

We first provide a brief background on looped models (\cref{sec:rdm-basics}), LTI systems (\cref{sec:lti-basics}), and modeling scaling laws (\cref{sec:scaling-laws-basics}).
Prior work has studied looped architectures along several design axes: loop placement (pre-, mid-, or post-looping)~\citep{saunshiReasoningLatentThoughts2025b}, halting mechanism (explicit
routers~\citep{baeMixtureofRecursionsLearningDynamic2025, zhuScalingLatentReasoning2025} vs.\
implicit stochastic depth~\citep{geiping_scaling_2025,
mcleish_retrofitted_recurrence}), topology (single
block~\citep{geiping_scaling_2025} or hierarchical~\citep{wangHierarchicalReasoningModel2025b,jolicoeur-martineauLessMoreRecursive2025}) and differentiation (explicit or implicit backpropagation \cite{bai2019deepequilibriummodels}). Our work focuses on implicit-halting middle-looped architectures using explicit differentiation; an extended review is in \cref{sec:lit-review}.

\subsection{Existing Middle-Looped Architectures}
\label{sec:rdm-basics}

In this paper, we focus on middle-looped architectures \citep{saunshiReasoningLatentThoughts2025b,geiping_scaling_2025}.
Middle-looped recurrent depth architecture contains three units: an initial \textit{prelude} unit \prelude, a middle \textit{recurrent} unit \recurrent, and a final \textit{coda} unit \coda. Formally, given an input $s \in V^n$, where $V$ is vocabulary and $n$ is sequence dimension, the outputs $p \in \mathbb{R}^{n \times |V|}$ can be computed by the following update rule:
$e = \mathcal{P}(s),~h_{t+1} = \mathcal{R}(h_t, e),~p = \mathcal{C}(h_T),$
where $h_0 \sim \mathcal{N}(0, \sigma^2 I_{d\times d})$ and $d$ the embedding dimension.
Intuitively, \prelude embeds inputs into the latent space, conditioning \recurrent as it recursively updates the hidden state $h_t \in \mathbb{R}^{n \times d}$ for $T$ iterations,
which \coda uses to generate $p$. Within \recurrent, prior work inject $e$ using addition $h_{t+1} = \mathcal{R}(h_t + e)$ \citep{yangLoopedTransformersAre2023} or concatenation with projection $h_{t+1} = \mathcal{R}(W[h_t; e])$ \citep{geiping_scaling_2025}, where $W \in \mathbb{R}^{d \times 2d}$. 

While looped models can be viewed as weight-sharing layers, modern variants allow for variable depth.
During training, depth $T$ is sampled per micro-batch \cite{bansalEndtoendAlgorithmSynthesis2022} from $\Lambda$ (e.g., Poisson with mean \meanrecurrence), exposing the model to variable depths for stronger test-time scaling \citep{anilPathIndependentEquilibrium}.
The training objective thus minimizes the expectation over the dataset and $\Lambda$.
Lastly, truncated backpropagation through depth, analogous to BPTT \citep{Hinton2013TrainingRN}, limits the backward pass to a constant \meanbackward \citep{geiping_scaling_2025}. 

\paragraph{Stability.}\citet{geiping_scaling_2025} found looped models unstable at scale and adopted a block pattern, combining Pre- and Post-Norm to normalize the residual: $\bar{x}^{(\ell)} = \text{LN}(\text{MHA}(\text{LN}(x^{(\ell-1)})) + x^{(\ell-1)}), \quad x^{(\ell)} = \text{LN}(\text{FFN}(\text{LN}(\bar{x}^{(\ell)})) + \bar{x}^{(\ell)})$
where $\mathrm{LN}(\cdot)$ denotes layer normalization, $\mathrm{MHA}(\cdot)$ multi-head attention, and $\mathrm{FFN}(\cdot)$ feed-forward networks. We later show that residual normalization is unnecessary when stability is properly controlled.

\subsection{Linear Time-Invariant Dynamical Systems}
\label{sec:lti-basics}
To study the instability of looped models, we will use an LTI dynamical system as a tractable linear surrogate for complex non-linear looped models.
In control theory, LTI systems are formalized through first-order differential equations
$\dot{h}(t)= \A h(t) + \B e(t),~y(t) = \C h(t)$
that describe the evolution of a hidden state $h(t) \in \mathbb{R}^{d_h}$ given an input signal $e(t) \in \mathbb{R}^{d_e}$,
where $\A \in \mathbb{R}^{d_h \times d_h}$ governs the dynamics of the system, $\B \in \mathbb{R}^{d_h \times d_e}$ controls how external inputs influence the state, and $\C \in \mathbb{R}^{d_e \times d_h}$ projects the hidden state to the output $y(t) \in \mathbb{R}^{d_e}$. The continuous system can be discretized to obtain 
$h_{t} = \dA h_{t-1} + \dB e_t,  y_t = \C h_t$
using a step size $\dt$; for instance, zero-order hold (ZOH) would yield $\dA = \exp(\Delta \A)$ and $\dB = (\Delta \A)^{-1}(\exp(\Delta \A) - I) \cdot \Delta \B$.

LTI systems fall into three regimes: \emph{stable} (bounded and convergent), \emph{marginally stable} (oscillatory), and \emph{unstable} (explosive and divergent). 
A fundamental property of LTI systems is that their \emph{stability} is determined by the eigenvalues of $\A$.
Continuous LTI systems require negative eigenvalues of $\A$; Discrete LTI systems requires $\rho(\dA) < 1$ \citep{1082819}, where $\rho$ computes the spectral norm, with unstable systems having $\rho(\dA)>1$. 

\subsection{Modeling Scaling Laws} 
\label{sec:scaling-laws-basics}
We follow \citet{hoffmann2022trainingcomputeoptimallargelanguage}, which modeled scaling law behaviors via parabolic and parametric fits for varying model sizes and training tokens with a fixed FLOP budget.
For parabolic fits, a quadratic is fit to several FLOP budgets to estimate the loss-optimal model size or number of training tokens. For parametric fits, a function form of $\widehat{\mathcal{L}}(N,D) = E + X \cdot N^{-x} + Y \cdot D^{-y}$ is fit using the Huber loss \citep{huber} between the predicted and empirical log loss values for varying parameters $N$ and tokens $D$, using L-BFGS \cite{lbfgs} to minimize.

%% file: sections/methodology.tex
\section{Understanding Instability in Looped Architectures}
\label{sec:dynamical-systems}

\begin{figure*}[t]
    \centering
    \includegraphics[width=\linewidth]{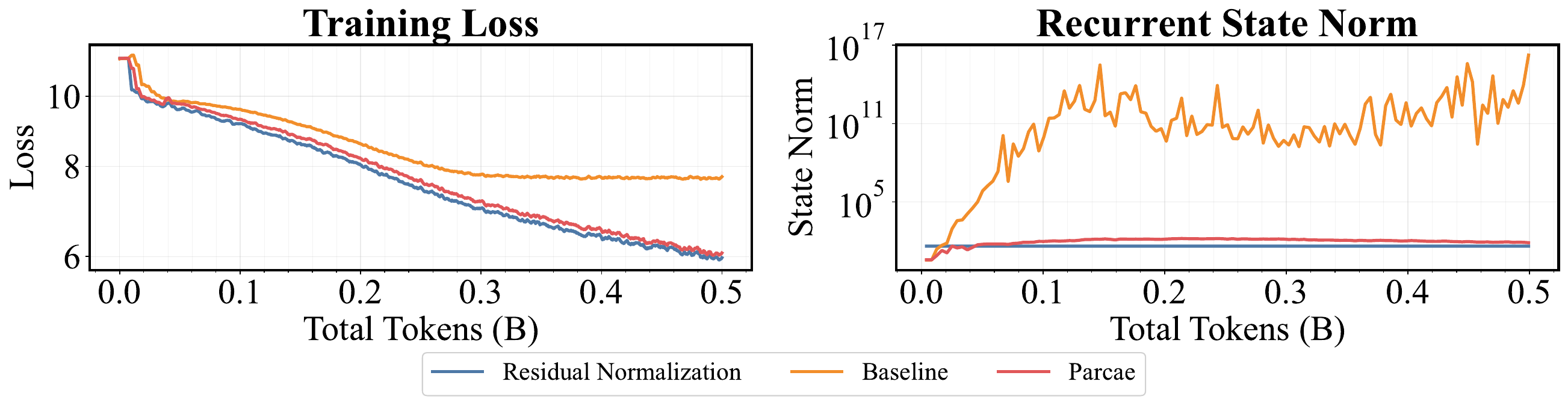}
    \caption{\textbf{Training Instability of Looped Architectures.} (\emph{left}) Pre-Norm looped models diverge, while residual norm. and \sysname converge. (\emph{right}) Instability stems from an exploding recurrent state norm $||h_T||_2$, the hidden embedding norm after $T$ recurrences.}
    \label{fig:instability}
    \vspace{-0.8em}
\end{figure*}

In this section, we study the instability of looped architectures. Using an LTI view over the residual, we find that instability stems from an unconstrained residual state explosion (\cref{fig:instability}; \cref{tab:hyperparameters}~[\emph{Baseline}]; \cref{sec:stability-ablations}).
While residual normalization helps mitigate this issue, it requires sensitive hyperparameter tuning (\cref{tab:hyperparameters} [\emph{Res. Norm}]), similar to fixed-depth transformers \citep{xu2019understandingimprovinglayernormalization, xiongLayerNormalizationTransformer2020}.
Using this LTI framework, we derive stability conditions for the eigenvalues of $\dA$. We find that prior work does not satisfy these conditions for $\dA$, which we empirically verify creates major state explosion (\cref{fig:spectral-norm}).

\paragraph{Dynamical System over Residual Stream.} Our key insight is to recast the forward pass as a dynamical system over the residual stream. Consider a transformer-based looped model as defined in \cref{sec:rdm-basics} for language modeling, where \prelude is an embedding layer that maps a sequence of tokens $s \in V^n$ into embedding space $e \in \R^{n \times d_h}$, \coda is a projection head that maps into probability space $g: d_h \to |V|$, and \recurrent is parameterized with $L$ transformer blocks. While several methods of input injection could condition \recurrent on $e$, building on prior work \citep{yang2024loopedtransformersbetterlearning, geiping_scaling_2025, mcleish_retrofitted_recurrence}, we focus on linear methods of injection (e.g., $\mathcal{R}(h_t, e) = \mathcal{R}(W_1 h_t + W_2 e)$, where $W_1 \in \R^{d_h  \times d_h}$ and $W_2 \in \R^{d_h \times d_e}$).\footnote{Both addition \citep{yang2024loopedtransformersbetterlearning} and concatenation \citep{geiping_scaling_2025} fall under this framework.}

Recall that \recurrent denotes the full recurrent update $h_{t+1} = \mathcal{R}(h_t,e)$, encompassing all transformer operations, including residual connections. 
The recurrent update can be exactly formulated as a non-linear time-variant dynamical system of the form $h_{t} = \dA h_{t-1} + \dB e + \overline{\mathcal{R}}(h_{t-1}, e), ~ y_t = \C h_t,$
where $\C \in R^{d_c \times d_h}$ decouples the \coda and \recurrent embedding dimension (i.e.~$p=\mathcal{C}(\C(h_T))$). 
This derivation is shown in \cref{sec:derivation-instabilty}. Though this formulation does not immediately elucidate instability,
linearizing of this system (i.e., dropping $\overline{\mathcal{R}}$) yields a discrete LTI system of the form:
\begin{equation}
h_{t+1} = \dA h_t + \dB e
\label{eq:lti}
\end{equation}

\begin{table}[!t]
    \centering
    \renewcommand{\arraystretch}{1.15}
    \small
    \begin{tabular*}{\textwidth}{@{\extracolsep{\fill}} l  c  c c c @{}}
    \toprule
         Method & $\dA$ & $\dB$ & $\rho(\dA)$ & LTI Stability \\
    \midrule
         Addition & $I$ & $I$ & $\rho(\dA) = 1$ &  \emph{marginally-stable} \\
         Concatenation & $\R^{d_h \times d_h}$ & $\R^{d_h \times d_e}$& $\rho(\dA) \in \R$  & \emph{unstable} \\
         \sysname (ours) & $\text{ZOH}(\texttt{Diag}(-\exp(\R^{d_h}))$ & $\text{Euler}(\R^{d_h \times d_e})$ &$\rho(\dA) < 1 $ & \emph{stable} \\
    \bottomrule
    \end{tabular*}
    \caption{\textbf{Comparison of Prior Update Rule Stability based on LTI Representation.}}
    \label{tab:stability-equation-comparison}
\end{table}

\begin{table}[!t]
    \centering
    \begin{minipage}[t]{0.4\textwidth}
        \vspace{0pt}
        \centering
        \small
        \begin{tabular}{lccc}
            \toprule
            \textbf{LR} & \textbf{Base} & \textbf{Res. Norm} & \textbf{Parcae} \\
            \midrule
            2e-4 & \cmark & \cmark & \cmark \\
            4e-4 & \xmark & \cmark & \cmark \\
            6e-4 & \xmark & \xmark & \cmark \\
            8e-4 & \xmark & \xmark & \cmark \\
            1e-3 & \xmark & \xmark & \cmark \\
            \bottomrule
        \end{tabular}
        \caption{\textbf{Hyperparameter Instability.} Convergence across learning rates for baseline RDMs, Res. Norm RDMs, and \sysname. \sysname is more robust to hyperparameter selection. Full logs are in \cref{sec:stability-ablations}.}
        \label{tab:hyperparameters}
    \end{minipage}
    \hfill
    \begin{minipage}[t]{0.56\textwidth}
        \vspace{-6pt}
        \centering
        \makeatletter\def\@captype{figure}\makeatother
        \includegraphics[width=\linewidth]{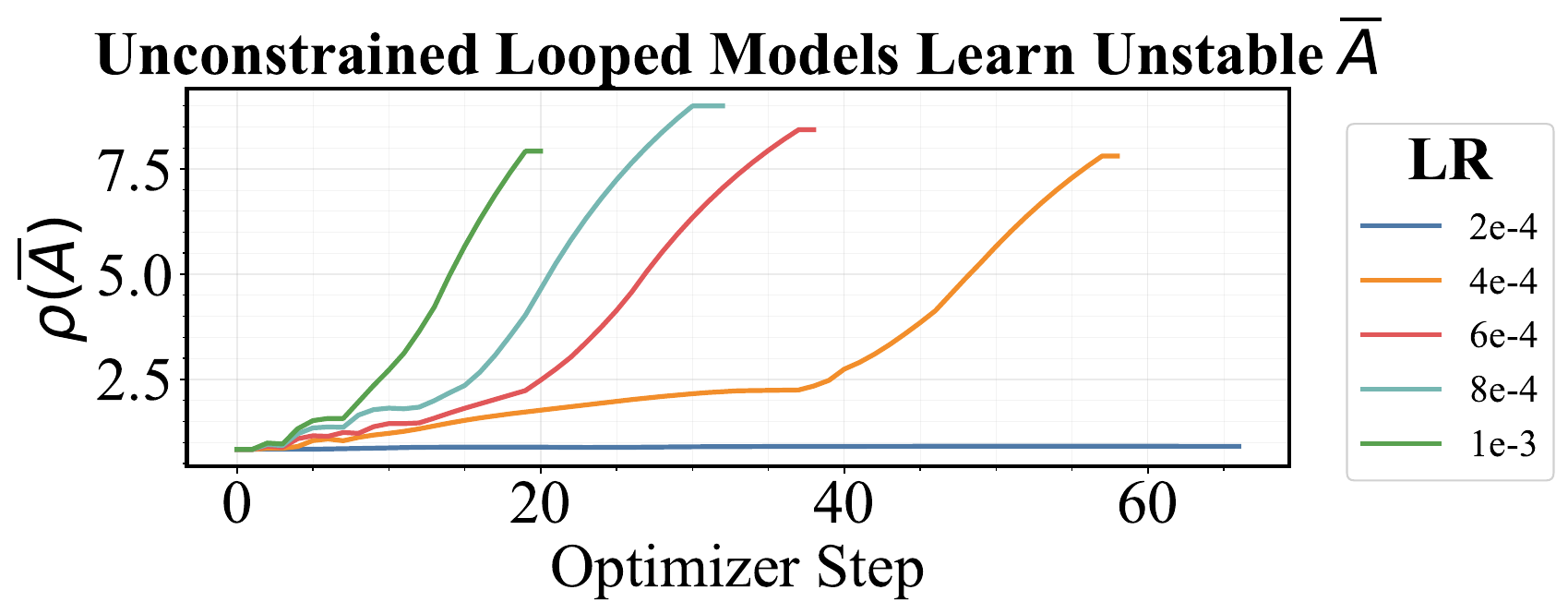}
        \vspace{-1.6em}
        \caption{\textbf{Spectral Radius of Unconstrained $\protect\dA$.} For a Pre-Norm RDM, we plot the $\rho(\dA)$ throughout training using different learning rates, observing divergent runs learn $\rho(\dA) > 1$. The state explosion, in \cref{fig:instability} is thus directly linked to $\dA$.}
        \label{fig:spectral-norm}
    \end{minipage}
    \vspace{-0.7em}
\end{table}

\paragraph{State Explosion from Unconstrained $\dA$ and $\dB$.}
Analyzing the stability of \cref{eq:lti} identifies $\rho(\dA)$ as a critical factor governing instability.
As shown in \cref{tab:stability-equation-comparison}, prior work \citep{geiping_scaling_2025, yang2024loopedtransformersbetterlearning} chooses parameterizations of $\dA$ such that $\rho(\dA)=1$ or $\rho(\dA)$ is unconstrained. Critically, these are \emph{marginally-stable} or \emph{unstable parameterizations}.

\cref{fig:spectral-norm} and \cref{tab:hyperparameters} confirm this empirically:
divergent runs learn a spectral radius of $\rho(\dA) \geq 1$, with convergent runs maintaining $\rho(\dA) < 1$, affirming that LTI stability constraints are necessary. 
Finally, at scale, we observe loss spikes late in training 
(e.g., after 170k steps), which we address by normalizing the input to $\dB$ (see \cref{sec:prelude-norm} for ablation).

%% file: sections/parcae.tex
\section{\sysname: A Stable Looped Architecture}
\label{sec:parcae}

Using our dynamical systems framework, we create \textbf{\emph{\sysname}}, a looped architecture that explicitly satisfies the stability constraints (\cref{sec:rfm-derivation}). 
Additionally, we propose a per-sequence depth sampling method to stabilize variance introduced by variable depth (\cref{sec:rfm-training}). 

\subsection{Block Design and Stable Parameterization of \sysname}
\label{sec:rfm-derivation}

We parameterize $\A$ and $\B$ in continuous form, and discretize using a learned $\dt \in \R^{d_h}$with ZOH and Euler schemes (i.e., $\dA = \exp(\dt \A)$ and $\dB = \dt \B$),\footnote{With abuse of notation, we let $\dt \A = \dt \odot \A$ (i.e., elementwise multiplication).} following prior sequence modeling work \citep{gu2024mambalineartimesequencemodeling,dao2024transformersssmsgeneralizedmodels}. 
To achieve our target stability conditions by constraining the eigenvalues of $\A$ to be negative, we parameterize $\A := \texttt{Diag}(-\exp(\texttt{log\_A}))$ as a negative diagonal matrix, where $\texttt{Diag}(-\exp(\cdot))$ of a vector enforces negativity and $\texttt{log\_A}\in \R^{d_h}$ is our learnable vector. 
While many formulations of $\A$ would work, ensuring negative eigenvalues in the diagonal case is simple and cheap.
$\B$ is left unconstrained; however, we introduce a normalization layer to the input $e$ to further stabilize training (see \cref{sec:prelude-norm} for ablation).
With this, our update rule, given an input sequence $s$, becomes
\begin{equation}
    e = \text{LN}(\mathcal{P}(s)), \qquad h_{t+1} = \dA h_t + \dB e + \overline{\mathcal{R}}(h_t, e), \qquad p = \mathcal{C}(\C h_T),
\end{equation}
where $h_0 \sim \mathcal{N}(0,~\sigma I_{d_h \times d_h})$ and $T$ is the number of loops.

We parameterize \prelude, $\overline{\mathcal{R}}$, and \coda using $L_{\mathcal{P}},L_{\mathcal{R}}$ and $L_{\mathcal{C}}$ transformer bloc:ks respectively. For exact block architecture, we match two different architectural setups: one for prior RDMs \citep{geiping_scaling_2025} and one for strong Transformer baselines \citep{nanochat}. \sysname's architecture matches RDMs, differing only in residual normalization and the dynamical systems parameters
(e.g., $\A, \B, \C, \dt$). Against Transformers, we follow a simplified \texttt{nanochat} \citep{nanochat} setup, where we match exact architecture, except we loop the middle third layers and include our dynamical systems parameters and a prelude norm. Exact model definitions and a forward pass can be found in \cref{sec:model-definitions} and \cref{sec:algorithm}, respectively.

\subsection{Stable Training Algorithms for \sysname}
\label{sec:rfm-training}

We further stabilize Parcae by adjusting the training objective. Specifically, looped models' training objective is
$\theta^\star \;=\; \arg\min_{\theta}\; \mathbb{E}_{(x,y)\sim \mathcal{D},\,T\sim \Lambda}\!\left[\;\ell\!\big(f_{\theta}(x;T),\, y\big)\;\right]$, implying that more depths should be sampled per global batch to more faithfully model the expectation over $\Lambda$.
Thus, we introduce a per-sequence depth sampling algorithm within a micro-batch, which we empirically observe to reduce loss spikes (ablation in \cref{sec:loss-spikes}).
Additionally, unlike prior work, we parameterize $\Lambda$ based on \meanrecurrence alone, as we find that truncating based on \meanbackward significantly hurts extrapolation to both lower and higher recurrences (ablation in \cref{sec:sampling-truncated-recurrence}).
Finally, we choose $\mu_{\text{bwd}} = \lceil \frac{\mu_{\text{rec}}}{2} \rceil$ throughout (see \cref{sec:scaling-of-truncated-backpropigation} for ablation). 
A detailed training algorithm is in \cref{sec:algorithm}.

%% file: sections/results.tex
\section{Results}
\label{sec:results}

We evaluate \sysname on end-to-end quality (\cref{sec:e2e}),
training FLOP scaling (\cref{sec:train-scaling}), and test-time scaling
(\cref{sec:inf-scaling}). We find that \sysname outperforms both parameter-
and data-matched RDMs and Transformers, optimal looping and data
follow predictable power laws, and test-time looping follows a saturating exponential decay.

\begin{table*}[!t]
    \centering
    \small
    \setlength{\tabcolsep}{4pt}
    \renewcommand{\arraystretch}{1.05}
    \begin{tabular*}{\textwidth}{@{\extracolsep{\fill}} llccc|ccccccc @{}}
        \toprule
         & \textbf{Model} & $\mathbf{T}$ & Val. & WikiText & Hellaswag & ARC-c & ARC-e & PIQA & BoolQ & SciQ & Avg. \\
         \midrule
         \multirow{2}{*}{\rotatebox{90}{100M}}
         & RDM& 16 & 14.23 & 63.27 & 27.16 & 17.66 & 42.38 & 59.14 & 51.35 & \textbf{72.50} & 45.03 \\
         & \sysname & 16 & \textbf{13.59} & \textbf{60.33} & \textbf{27.18} & \textbf{18.09} & \textbf{43.10} & \textbf{59.30} & \textbf{61.83} & 71.50 & \textbf{46.83} \\
        \midrule
         \multirow{2}{*}{\rotatebox{90}{350M}}
         & RDM& 8  & 10.76 & 41.31 & 28.55 & 20.90 & 47.26 & 61.75 & \textbf{61.53} & 76.70 & 49.45 \\
         & \sysname & 8  & \textbf{10.09} & \textbf{37.53} & \textbf{29.23} & \textbf{21.08} & \textbf{48.78} & \textbf{62.08} & 60.73 & \textbf{78.80} & \textbf{50.12} \\
         \bottomrule
    \end{tabular*}
    \caption{\textbf{Zero-Shot and Perplexity Results Trained on RDM Setup.} Comparison of \sysname and RDM \citep{geiping_scaling_2025} on 
    a variety of open source benchmarks and perplexity held-out validation set and Wikitext \citep{merity2016pointer}. \textbf{Best} results are \textbf{bolded}.}
    \label{tab:rdm-parcae}
\end{table*}

\begin{table}[!t]
\centering
\small
\setlength{\tabcolsep}{4.5pt}
\begin{tabular}{l ccc ccc ccc}
\toprule
& \multicolumn{3}{c}{Val Loss ($\downarrow$)} & \multicolumn{3}{c}{Core ($\uparrow$)} & \multicolumn{3}{c}{Core Ext ($\uparrow$)} \\
\cmidrule(lr){2-4} \cmidrule(lr){5-7} \cmidrule(lr){8-10}
Configuration & $T\!=\!1$ & $T\!=\!4$ & $T\!=\!8$ & $T\!=\!1$ & $T\!=\!4$ & $T\!=\!8$ & $T\!=\!1$ & $T\!=\!4$ & $T\!=\!8$ \\
\midrule
RDM          & \multicolumn{3}{c}{\textit{Divergent Training}} & \multicolumn{3}{c}{\textit{Divergent Training}} & \multicolumn{3}{c}{\textit{Divergent Training}}\\
+\,Constrained\ $\dA$       & 8.99 & 3.15 & 2.97 & $-2.0_{\pm0.1}$ & $11.0_{\pm0.1}$ & $13.2_{\pm0.2}$ & $0.5_{\pm0.1}$ & $7.8_{\pm0.0}$ & $9.1_{\pm0.5}$ \\
+\,Per-Seq.\ Sampling & 3.38 & 3.01 & 2.98 & $\mathbf{7.6_{\pm0.2}}$ & $13.4_{\pm0.2}$ & $14.0_{\pm0.2}$ & $\mathbf{5.9_{\pm0.4}}$ & $9.3_{\pm0.2}$ & $\mathbf{9.9_{\pm0.2}}$ \\
+\,Prelude Norm       & \textbf{3.28} & \textbf{2.97} & \textbf{2.95} & $7.5_{\pm0.3}$ & $\mathbf{13.5_{\pm0.0}}$ & $\mathbf{14.0_{\pm0.2}}$ & $5.8_{\pm0.3}$ & $\mathbf{9.4_{\pm0.1}}$ & $9.7_{\pm0.3}$ \\
\bottomrule
\end{tabular}
\caption{\textbf{Stability Results Trained on Transformer Setup.} To illustrate stability, we retrofit a baseline 140M Transformer into a RDM and then sequentially add our stability improvements.}
\label{tab:ablation}
\end{table}

\subsection{\sysname Improves End-to-End Quality}
\label{sec:e2e}

We compare \sysname against parameter- and data-matched RDMs and Transformers, finding that \sysname is more stable than prior looped models and that it outperforms both in quality. 

\paragraph{Setup.} 
For RDMs, we follow \citet{geiping_scaling_2025}, using the Huginn dataset and tokenizer for training. For transformers, we follow \citet{nanochat} and train on \texttt{FineWeb-Edu} \citep{penedo2024finewebdatasetsdecantingweb}.
For both RDM and Transformer setups, we perform hyperparameter sweeps for both RDMs and Transformers, and then use them for \sysname (i.e., we perform no hyperparameter sweeps for \sysname models). Extended model definitions, hyperparameter selection, and evaluation setup can be found in \cref{sec:model-definitions}, \cref{sec:hyperparameters}, and \cref{sec:evaluation-setup}, respectively. 

\textbf{Comparison against RDMs}. \cref{tab:rdm-parcae} shows that \sysname reduces perplexity by up to 6.2 \% and 9.1 \% on a held-out validation set and WikiText \citep{merity2016pointer} against prior RDMs \citep{geiping_scaling_2025}, while additionally performing up to 1.8 points better on the average of several downstream benchmarks. \cref{tab:ablation} ablates that each modification of \sysname contributes:
constraining $\dA$ enables convergence at high $T$ (e.g., $\mu_{\text{rec}}=T\!=\!8$),
per-sequence sampling stabilizes lower test-time depths, and the prelude norm
further improves quality across all $T$ (and late stage stability \cref{sec:prelude-norm}).

\begin{table*}[!t]
    \centering
    \small
    \setlength{\tabcolsep}{4pt}
    \renewcommand{\arraystretch}{1.05}
    \begin{tabular*}{\textwidth}{@{\extracolsep{\fill}} llc|cccc @{}}
        \toprule
         & \textbf{Model} & $\mathbf{T}$ & Val. PPL ($\downarrow$) & Lambada PPL ($\downarrow$) & Core ($\uparrow$) & Core-Extended ($\uparrow$) \\
        \midrule
        \multirow{2}{*}{\rotatebox{90}{140M}} & Transformer & -- & 21.48 & 127.39 & 13.00 ± 0.15 & 8.80 ± 0.21 \\
         & \sysname & 8  & \textbf{19.06} & \textbf{80.64} & \textbf{14.04 ± 0.20} & \textbf{9.67 ± 0.28} \\
        \midrule
        \multirow{2}{*}{\rotatebox{90}{370M}} & Transformer & -- & 15.79 & 40.77 & 17.46 ± 0.03 & 11.71 ± 0.22 \\
         & \sysname & 8  & \textbf{14.49} & \textbf{32.74} & \textbf{20.00 ± 0.06} & \textbf{12.75 ± 0.31} \\
        \midrule
        \multirow{2}{*}{\rotatebox{90}{770M}} & Transformer & -- & 13.08 & 22.37 & 22.42 ± 0.20 & 14.20 ± 0.63 \\
         & \sysname & 8  & \textbf{12.49} & \textbf{19.71} & \textbf{25.07 ± 0.33} & \textbf{15.19 ± 0.43} \\
        \midrule
        \multirow{2}{*}{\rotatebox{90}{1.3B}} & Transformer & -- & 11.95 & 17.26 & 25.45 ± 0.08 & 15.90 ± 0.23 \\
         & \sysname & 8  & \textbf{11.42} & \textbf{14.71} & \textbf{28.44 ± 0.28} & \textbf{17.08 ± 0.09} \\
        \bottomrule
    \end{tabular*}
    \caption{\textbf{Comparing \sysname to Fixed-Depth Transformers.} We pretrain Transformers and \sysname with a \texttt{nanochat} setup at several scales, evaluating on a held-out validation set, Lambada \citep{paperno2016lambada}, Core, and Core-Extended \citep{li2025datacomplmsearchgenerationtraining}. 
    \textbf{Best} results are \textbf{bolded.}}
    \label{tab:trans-parcae}
\end{table*}

\textbf{Comparison Against Transformers.} \cref{tab:trans-parcae} shows that \sysname reduces validation perplexity by 4.3--9.2\% and improves Core and Core-Extended Scores by up to 2.99 and 1.18 points, respectively. We find that our 770M \sysname model achieves quality comparable to the 1.3B Transformer on Core \citep{li2025datacomplmsearchgenerationtraining} with roughly half the parameters. 
Measured as a fraction of the quality gap to the next larger Transformer (e.g., for 140M Core-Extended: $\frac{9.67-8.80}{11.71-8.80} \cdot 100 \approx 29.9 \%$), \sysname achieves a \textbf{\emph{23.3-87.5\% and 29.9-58.2\%}} better parameter efficiency for Core and Core-Extended, respectively.

\begin{figure}[!t]
    \centering
    \vspace{-0.4em}
    \includegraphics[width=\linewidth]{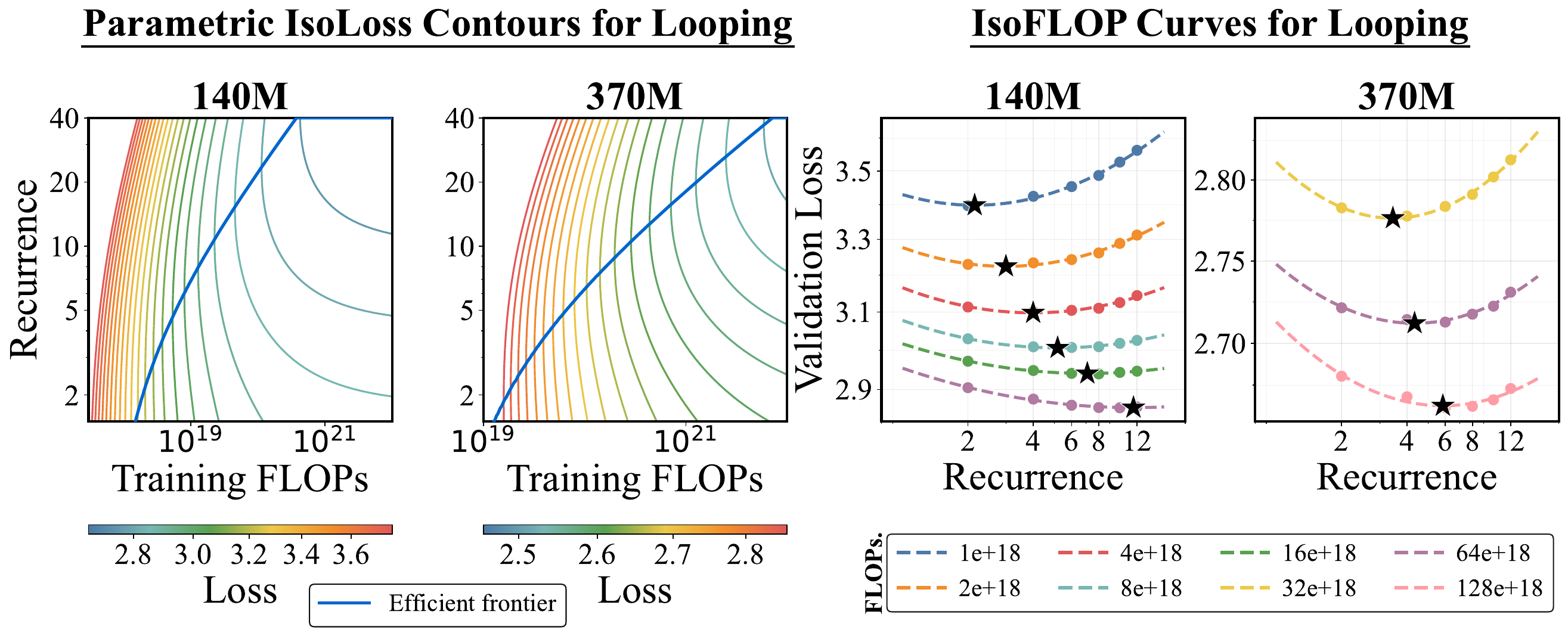}
    \vspace{-1.4em}
    \caption{\textbf{Looping Scales Training Compute Optimally.}~(\emph{Left}) Parametric isoLoss contours over \meanrecurrence and data. The efficient frontier (blue line) traces the lowest FLOP budget required to achieve each loss level, showing that optimal training requires increased looping. (\emph{Right}) Parabolic isoFLOP fits for 140M and 370M models reveal a clear optimum \meanrecurrence at each FLOP budget, indicating that looping is an orthogonal scaling axis to data.}
    \label{fig:train-scaling-laws}
    \vspace{-0.4em}
\end{figure}

\subsection{Looping as an Orthogonal Scaling Axis in Training}
\label{sec:train-scaling}

In this section, we explore the FLOP efficiency of looping under a fixed FLOP and parameter budgets. We find that looping introduces an orthogonal axis for scaling compute, where compute-optimal training increases \meanrecurrence and data in tandem following empirical power laws.

\paragraph{Setup.} We train 140M and 370M \sysname models under fixed FLOP and parameter budgets, varying training tokens and mean recursion \meanrecurrence using the \texttt{nanochat} setup. Additional training details and FLOP estimates can be found in \cref{sec:scaling-laws-setup} and \cref{sec:flop-estimate}, respectively.

\begin{figure}[!t]
    \centering
    \includegraphics[width=\linewidth]{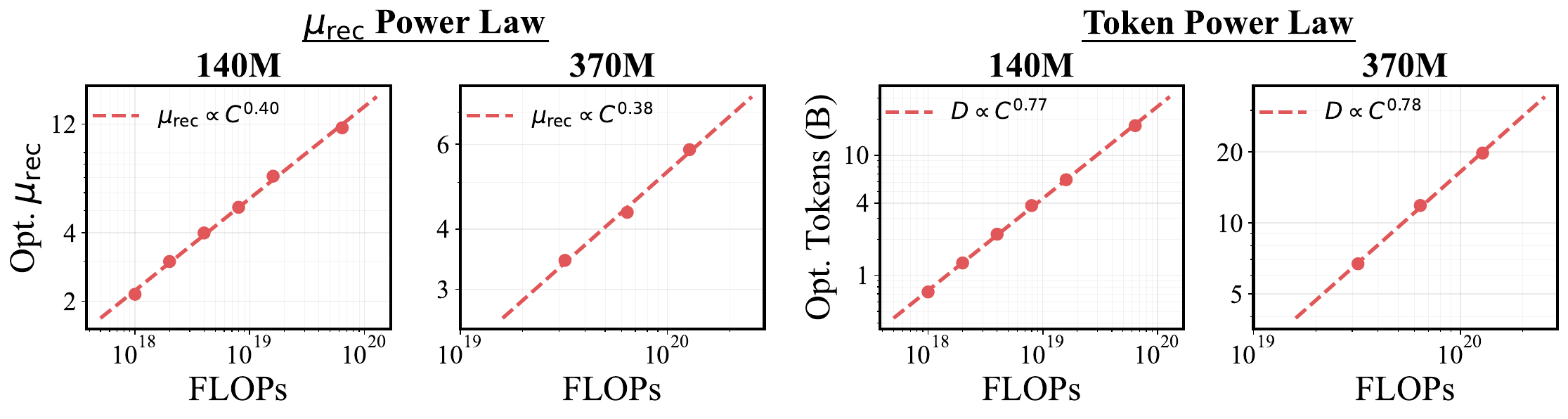}
    \vspace{-1.5em}
    \caption{\textbf{Optimal \meanrecurrence and Tokens Follows Predictable Power Laws.} We fit a parabola to each isoFLOP budget for both 140M and 370M \sysname models, using its minima to approximate the optimal \meanrecurrence and token budget at each scale. We observe that optimal recurrence (\emph{left plots}) and tokens (\emph{right plots}) follow a predictable power law with similar coefficients at both scales.}
    \label{fig:power-laws}
\end{figure}

\textbf{Modeling Scaling Laws of Looping}. At 140M and 370M scales, isoFLOP curves show that increasing \meanrecurrence while proportionally reducing tokens yields lower validation loss than training at low recurrence (\cref{fig:train-scaling-laws} [\emph{right}]). Using a parabolic fit, we extract the optimal \meanrecurrence and token budget at each FLOP level, finding that both follow predictable power laws (\cref{fig:power-laws}) with consistent exponents ($\gamma_{\mu} \approx 0.40$, $\gamma_D \approx 0.78$). 
We also fit a parametric function $\widehat{\mathcal{L}}(\mu_{\text{rec}}, D) = E + X\cdot \mathbf{N}(\mu_{\text{rec}})^{-x} + Y \cdot D^{-y}$ over the effective parameterization $\mathbf{N}(\mu_{\text{rec}})$ (i.e., parameters of unrolling the looped model) and tokens $D$ (\cref{fig:train-scaling-laws}, [\emph{left}]; details in \cref{sec:fit-par}), enabling predictable extrapolation of loss to unseen budgets. To verify, we predict the validation loss of held-out models in \cref{sec:e2e}, achieving 1.3\% and 0.8\% error at 140M and 370M, respectively.

\begin{table}[!t]
    \centering
    \begin{minipage}[t]{0.38\textwidth}
        \vspace{-6pt}
        \centering
        \makeatletter\def\@captype{figure}\makeatother
        \includegraphics[width=\linewidth]{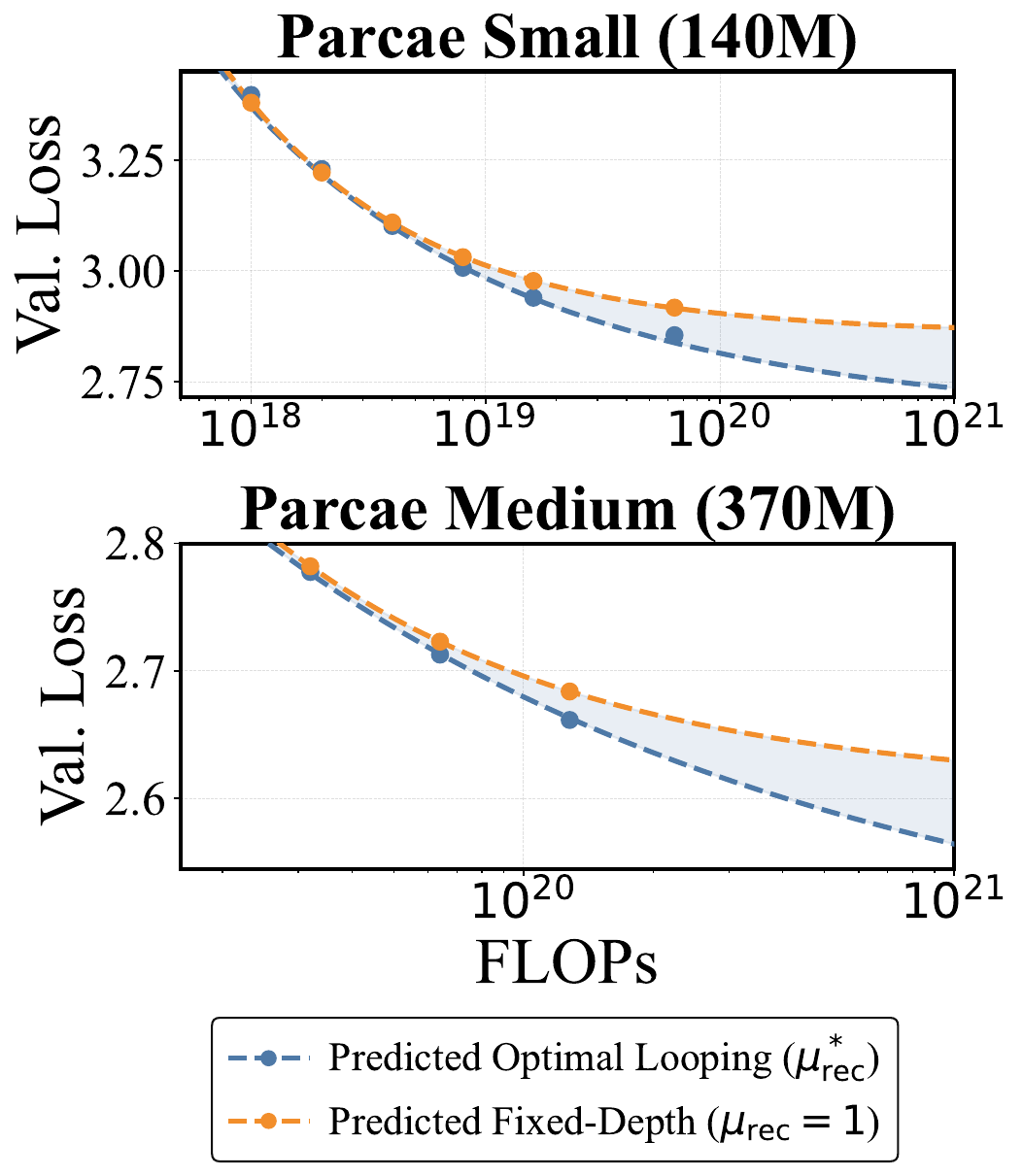}
        \vspace{-1.56em}
        \caption{\textbf{Pareto Frontier of Looping.} We observe that looping has a stricter IsoFLOP optimal loss frontier over fixed-depth, non-looped models. Dots are empirical points.}
        \label{fig:optimal-frontier}
    \end{minipage}
    \hfill
    \begin{minipage}[t]{0.60\textwidth}
        \vspace{0pt}
        \centering
        \begin{tabular}{@{}c cc | cc | cc@{}}
        \toprule
        & \textbf{FLOPs}& & \multicolumn{2}{c}{\textbf{Optimal $\mu_{\mathrm{rec}}^*$}} & \multicolumn{2}{c}{\textbf{Fixed-Depth}} \\
        \cmidrule(lr){4-5} \cmidrule(lr){6-7}
        & ($\times 10^{18})$ & $\mu_{\mathrm{rec}}^*$ & Core & Core Ext. & Core & Core Ext. \\
        \midrule
        \multirow{6}{*}{\rotatebox[origin=c]{90}{\textbf{140M}}}
        & $1$   & 2  & $7.6$ & $5.7$ & $\mathbf{7.9}$ & $\mathbf{6.1}$ \\
        & $2$   & 2  & $9.0$ & $6.2$ & $\mathbf{10.5}$ & $\mathbf{6.4}$ \\
        & $4$   & 4  & $\mathbf{11.2}$ & $\mathbf{8.4}$ & $10.7$ & $8.1$ \\
        & $8$   & 6  & $10.5$ & $\mathbf{7.8}$ & $\mathbf{11.8}$ & $7.7$ \\
        & $16$  & 8  & $\mathbf{14.6}$ & $\mathbf{9.8}$ & $13.0$ & $8.8$ \\
        & $64$  & 10 & $\mathbf{16.2}$ & $\mathbf{11.0}$ & $15.0$ & $9.5$ \\
        \midrule
        \multirow{3}{*}{\rotatebox[origin=c]{90}{\textbf{370M}}}
        & $32$  & 4  & $15.2$ & $10.1$ & $\mathbf{16.8}$ & $\mathbf{11.2}$ \\
        & $64$  & 6  & $\mathbf{18.1}$ & $11.6$ & $\mathbf{18.1}$ & $\mathbf{12.1}$ \\
        & $128$ & 6  & $\mathbf{20.1}$ & $\mathbf{13.0}$ & $18.1$ & $12.0$ \\
        \bottomrule
        \end{tabular}
        \caption{\textbf{Core Scores Comparison of Looping Optimal Frontier over Purely Scaling Data.} We evaluate the downstream quality of fixed-depth (\meanrecurrence=1) and looped \sysname models trained with fixed parameters and FLOP budgets. At both scales, using the optimal \meanrecurrence results in better Core and Core-Extended scores at extended FLOP budgets. Expanded results can be found in \cref{sec:fixed-comparison-expanded}.}
        \label{tab:qual-fix-loop}
    \end{minipage}
    \vspace{-1em}
\end{table}

\textbf{IsoFLOP comparison of Looping with Fixed-Depth} \cref{fig:optimal-frontier} shows fixed-depth \sysname models without looping at each FLOP budget. The optimal curve achieves a strictly lower loss, which translates to 1.2-2.0 points higher Core scores (\cref{tab:qual-fix-loop}).

\subsection{Test-Time Scaling Laws of \sysname}
\label{sec:inf-scaling}

We study looping as a mechanism for scaling test-time compute. We find the test-time compute follows a predictable saturating exponential decay, which can be unified with \cref{sec:train-scaling}, connecting both training and test-time scaling laws.

\paragraph{Setup.} We train 140M and 370M \sysname models under a fixed data budget with $\mu_{\text{rec}} \in \{2, 4, 6, 8, 10, 12\}$ following our \texttt{nanochat} setup, evaluating up to $T = 24$. We additionally evaluate models from \cref{sec:train-scaling} for the unified scaling laws. See \cref{sec:scaling-laws-setup} for details.

\begin{figure}[!t]
    \centering
    \includegraphics[width=\linewidth]{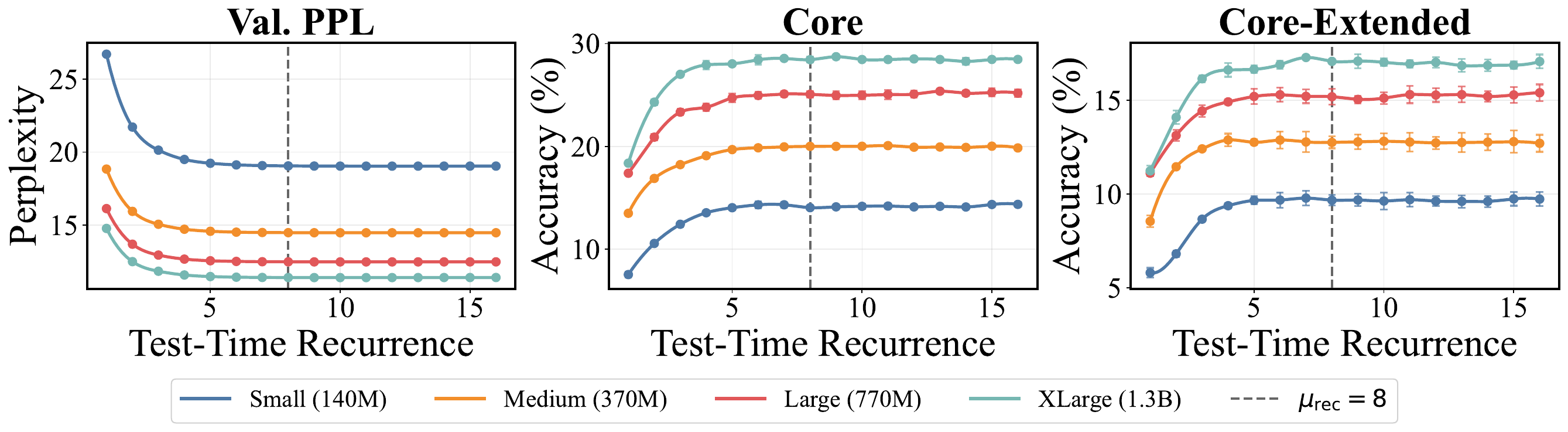}
    \vspace{-1.5em}
    \caption{\textbf{Test-Time Scaling of \sysname.} When evaluating \sysname models from \cref{tab:trans-parcae}, we observe test-time looping follows a predictable saturating trend, consistent across model sizes.}
    \label{fig:parcae-test-time-scaling}
\end{figure}

\textbf{Saturation of Test-Time Compute.} While prior works observed test-time generalization in small synthetic tasks \citep{yangLoopedTransformersAre2023, bansalEndtoendAlgorithmSynthesis2022}, we find quality to be bounded in large-scale language modeling. 
Evaluating models from \cref{sec:e2e} at $2\times$ \meanrecurrence across all four scales (\cref{fig:parcae-test-time-scaling}), we observe that gains plateau near \meanrecurrence, suggesting training depth determines the test-time scaling ceiling.

\begin{figure}[!t]
    \centering
    \includegraphics[width=\linewidth]{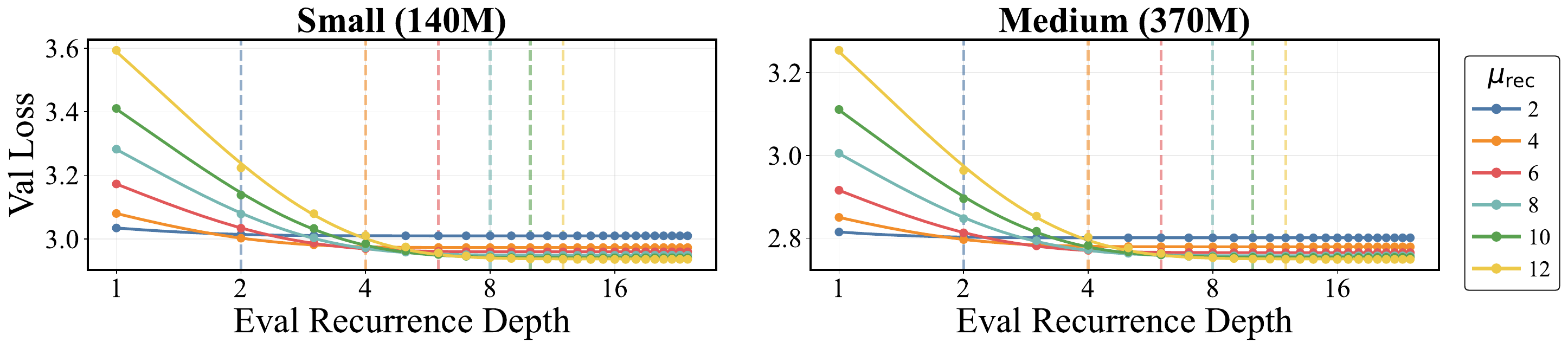}
    \vspace{-1.4em}
    \caption{\textbf{Scaling Test-Time Compute follows a Predictable Power Laws.}  We plot the validation loss with different \meanrecurrence as a function of test-time recurrence $T$, and find the fitted exponential decay (solid curve for each \meanrecurrence) tightly captures the test-time performance of looping.
    }
    \vspace{-1em}
    \label{fig:test-time-scaling-laws}
\end{figure}

\textbf{Modeling Scaling Laws of Test-Time Looping.} We find that the test-time scaling curves are well-described by a saturating exponential decay of the form: $\mathcal{L}(T) = \mathcal{L}_\infty + Ze^{-z\cdot T}$. This form tightly captures the saturation dynamics for each model (\cref{fig:test-time-scaling-laws}; see \cref{sec:test-par} for details), achieving an average Huber loss of $2.5 \times 10^{-7}$ and $1.8 \times 10^{-7}$ for 140M and 370M, respectively.

\textbf{Unifying Training and Test-Time Scaling Laws.} From the learned fits in \cref{fig:test-time-scaling-laws}, we observe that $\mathcal{L}_\infty$ matches the training law prediction at $T = \mu_\text{rec}$ (\cref{sec:train-scaling}), and that the per-curve decay rate scales inversely with training depth as $z / \mu_\text{rec}$ (see \cref{sec:test-par} for details).
These observations motivate a unified scaling law that connects training and test-time compute:
\begin{equation}
   \widehat{\mathcal{L}}_\text{unified}(T \mid \mu_\text{rec}, D) = \underbrace{E + X \cdot \mathbf{N}(\mu_\text{rec})^{-x} + Y \cdot D^{-y}}_{\text{Training Law Floor } \widehat{\mathcal{L}}_\text{train}(\mu_\text{rec}, D)} + \underbrace{Z \cdot \exp\!\left(-z \cdot T \cdot \mu_\text{rec}^{-1}\right)}_{\text{Test-Time Decay}}
   \label{eq:unified}
\end{equation}
where $\widehat{\mathcal{L}}_{\text{train}}(\mu_{\text{rec}}, {D})$ is the training law in \cref{sec:train-scaling}, and $(Z, z)$ are two fitted parameters governing the test-time scaling. 
The training law sets the irreducible floor, while the decay rate $-z \cdot T / \mu_\text{rec}$ captures how quickly additional recurrences approach it. 
On held-out 140M and 370M \sysname models (\cref{sec:e2e}), the unified fit predicts test-time loss within 0.85-1.31\% average error, dropping further to 0.1-0.17\% average error when the empirical loss at $T = \mu_{\text{rec}}$ is used. This confirms that \cref{eq:unified} captures saturation dynamics, with residual error attributable to the training law's $\sim1\%$ extrapolation gap (see \cref{sec:test-par} for extended details).

%% file: sections/conclusion.tex
\section{Discussion and Future Work}
\label{sec:lim-fut}

In this section, we briefly discuss limitations and future directions.

\paragraph{Looped Architectures.} While several design choices around looped architectures have been guided by small-scale empirical results, a deep investigation of loop-unit placement \cite{jacobs2026blockrecurrentdynamicsvisiontransformers}, composition (e.g., number of parameters in the recurrent unit and usage of different architectures), and extreme looping (e.g., increasing mean recurrence to deeper depths) at a larger scale is warranted. Within our dynamical systems framework, the use of different discretizations, full-rank parameterizations, and recurrent update rules warrants investigation to enable recurrence at larger depths.

\paragraph{Scaling.}
While we find Parcae to induce predictable, optimal scaling laws for layer looping, our observations are limited to small architectures. It remains to be seen if Parcae compares favorably when scaling these observations to large FLOP budgets and parameterizations. We are also interested in the interplay of parameters, data, and recurrence as orthogonal axes, and how they should be efficiently scaled together. Finally, one limitation of looping is that, as \meanrecurrence increases, the number of test-time steps required to achieve equivalent quality increases. An investigation of techniques that maintain quality with fewer inference time steps is an interesting future direction.

\section{Conclusion}
\label{sec:conclusion}

In this work, we study the stability of looped models through a dynamical systems framework and propose \textbf{\sysname}, a stable looped architecture that prevents residual explosion by constraining the spectral norm of the injection parameters.
\sysname outperforms data- and parameter-matched prior looped models and baseline Transformers, matching downstream quality of models up to twice its size.
We further establish scaling laws for looping: FLOP-optimal training increases looping and data in tandem following predictable power laws, while test-time looping follows a saturating exponential decay law, yielding a unified scaling law connecting training and inference compute.

%% file: appendix/notation.tex
\newpage
\section{Glossary}

We include a brief glossary of both notations and common metrics used to define and analyze looped architectures.

\subsection{Notation}
\label{sec:notation}

\begin{table}[h]
    \centering
    \begin{tabular}{lll}
    \toprule
         Notation &  Description \\
    \midrule
         $d$ & Embedding \textbf{dimension} of the model & \\
         $t$ & Discrete temporal \textbf{state} axis of \recurrent on $\mathbb{N}$& \\
         $b$ & Global batch size used during pretraining \\
    \midrule
         \prelude & Initial \textbf{prelude} block of a recurrent architecture & \\
         \recurrent & Middle \textbf{recurrent} block of a recurrent architecture & \\
         \coda & Final \textbf{coda} block of a recurrent architecture & \\
         $\A$ & The linear continuous state transition matrix & \\
         $\B$ & The linear continuous state injection matrix & \\
         $\C$ & The linear state output matrix & \\
         $\dt$ & Learnable discrete parameter for decay, discretizing our model & \\
    \midrule
         $s$ & Input sequence to a model & \\
         $e$ & Output embedding of the prelude block \prelude \\
         $h$ & Hidden embedding of the recurrent block \recurrent \\
         \meanrecurrence & Mean recurrent forward propagation steps during pre-training \\
         \meanbackward & Mean recurrent backward propagation steps during pre-training \\
         $n$ & Sampled number of recurrent steps with no gradient updates \\
         $k$ & Sampled number of recurrent steps with gradient updates \\
         $T$ & Sampled or fixed number of recurrent steps actually taken \\
         $\Lambda$ & Distribution that recurrences are sampled from during training \\
    \bottomrule
    \end{tabular}
    \caption{Glossary of notation and terminology. (\textit{Top}) Frequently used dimensions for tensors. (\textit{Middle}) Definition of Parcae blocks. (\textit{Bottom}) Tensors and distributions are used to express recurrent depth models.} 
    \label{tab:placeholder}
\end{table}

\subsection{Common Metrics}
\label{sec:metrics}
\begin{itemize}
    \item Recurrent Residual Metric: $||h_T - h_{T-1}||_2$, where $T \sim \Lambda$. This metric tells us how much we jump around at the final recurrence. Overly small jumps indicate that \recurrent isn't learning anything meaningful, while overly large jumps indicate \recurrent is suffering from state explosion or is unable to learn fixed-point dynamics.
    \item Recurrent State Norm: $||h_T||$, where $T \sim \Lambda$. In general, we don't want an overly large recurrent state norm as it creates numerical instabilities and leads to overly large gradients.
\end{itemize}

%% file: appendix/discussion.tex
\newpage
\section{Extended Literature Review}
\label{sec:lit-review}

Looping model depth has been well explored by prior work; with a large body of work studying looping within general language modeling \citep{dehghaniUniversalTransformers2019, zhuScalingLatentReasoning2025, geiping_scaling_2025, mcleish_retrofitted_recurrence, baeMixtureofRecursionsLearningDynamic2025} or small-scale algorithmic problems \citep{avi_learn_algorithm, yangLoopedTransformersAre2023, bansalEndtoendAlgorithmSynthesis2022, wangHierarchicalReasoningModel2025b, jolicoeur-martineauLessMoreRecursive2025}. Within looped architectures, the design of training paradigms can be relatively split between architectures with explicit halting mechanisms \citep{dehghaniUniversalTransformers2019, zhuScalingLatentReasoning2025, baeMixtureofRecursionsLearningDynamic2025, jolicoeur-martineauLessMoreRecursive2025, wangHierarchicalReasoningModel2025b} and those with implicit halting mechanisms \citep{geiping_scaling_2025, mcleish_retrofitted_recurrence, LoopFormerElasticDepthLooped2025, xuExpressivePowerLooped2025}. 
Looped architectures trained with an explicit halting mechanism use specialized architectures to predict when to early exit tokens, preventing additional computation updates on their recurrent stream \citep{wangHierarchicalReasoningModel2025b, jolicoeur-martineauLessMoreRecursive2025, baeMixtureofRecursionsLearningDynamic2025, dehghaniUniversalTransformers2019, elbayadDepthAdaptiveTransformer2020}. Specifically, \citet{wangHierarchicalReasoningModel2025b, jolicoeur-martineauLessMoreRecursive2025} formalize \emph{adaptive-computation-time}, a method that utilizes Q-learning as a means to determine convergence. Similarly, works such as \citet{baeMixtureofRecursionsLearningDynamic2025} define an architecture that uses light-weight routers to assign dynamic recursion depths, while \citet{zhuScalingLatentReasoning2025} uses a prediction head to dynamically define a probability of exiting after recurrent passes. A majority of these approaches draw on methods of layer skipping \citep{elhoushiLayerSkipEnablingEarly2024, raposoMixtureofDepthsDynamicallyAllocating2024}; however, these methods differ from using a shared parameterization for a recurrent block.

Alternatively, looped architectures with an implicit halting mechanism, such as \citet{geiping_scaling_2025, mcleish_retrofitted_recurrence,avi_learn_algorithm,bansalEndtoendAlgorithmSynthesis2022}, train models with stochastically sampled recurrent steps during pretraining, and then use the KL-divergence between two successive steps to decide when to exit from the recurrent unit early. Finally, \citet{LoopFormerElasticDepthLooped2025} ignores adaptive early exiting altogether, instead pretraining a recurrent unit on a static number of recurrences and enforcing a consistency loss on intermediate recurrences.
Our work focuses solely on implicit recurrent depth models \citep{geiping_scaling_2025, mcleish_retrofitted_recurrence}, which are derived from prior initial work \citep{avi_learn_algorithm, bansalEndtoendAlgorithmSynthesis2022}.

Beyond training paradigms, there are several differing architectural design choices for looped models \citep{geiping_scaling_2025,bansalEndtoendAlgorithmSynthesis2022, saunshiReasoningLatentThoughts2025b}. In simple looped architectures that only place a single recurrent unit, the placement of the looped unit is non-trivial, with certain works looping over all layers \citep{dehghaniUniversalTransformers2019,Csordas2024MoEUTMU,Bae2024RelaxedRT}. Alternatively, \citet{saunshiReasoningLatentThoughts2025b} find middle-looping recurrent units are the most effective in comparison to other formulations, such as pre-looping and post-looping, which loop the beginning and end of the model. The effectiveness of Middle-looping is consistent with the initial work in synthetic problems by \citet{bansalEndtoendAlgorithmSynthesis2022, avi_learn_algorithm} and with the architecture choices of \citet{geiping_scaling_2025,mcleish_retrofitted_recurrence} in large-scale language models before training. Within middle-looping architectures, the number of layers within each unit is mostly chosen ad hoc; however, when bootstrapping from a baseline model, \citet{koishekenov2025encodethinkdecodescaling} found that you optimize placement by algorithmically selecting layers within a model to loop.

While these prior formulations of looping focus on a single recurrent block, hierarchical \citep{wangHierarchicalReasoningModel2025b,jolicoeur-martineauLessMoreRecursive2025}, parallel \citep{wu2025parallellooptransformerefficient}, and multi-step \citep{jacobs2026blockrecurrentdynamicsvisiontransformers} formulations of layer looping exist. Furthermore, while not all under the same architectural paradigm, layer looping has been explored in multiple domains (e.g., language \citep{geiping_scaling_2025,mcleish_retrofitted_recurrence}, images \citep{jacobs2026blockrecurrentdynamicsvisiontransformers}, multi-modal systems \citep{alabdulmohsin2025recursiveinferencescalingwinning}, synthetic algorithmic problems \citep{avi_learn_algorithm,bansalEndtoendAlgorithmSynthesis2022,yangLoopedTransformersAre2023}), with the choice of looping style and model architecture design changing based on the specific modality. Where layer looping is introduced, how it is affected by individual modalities, and efficient, FLOP-optimal implementations of layer looping remain open questions.

Finally, layer looping is often deeply tied to deep equilibrium (DEQ) models \citep{bai2019deepequilibriummodels,bai2022neural}, due to the fixed-point nature often learned in recurrence. DEQs find the equilibrium points via root-finding to approximate an \emph{infinite depth} network. However, unlike looped architectures trained with truncated backpropagation, a key advantage of DEQ models is their use of implicit differentiation through \emph{infinite depth}, which keeps memory constant and independent of effective depth used to solve the fixed point using a rooting finding algorithm. While the use of implicit differentiation in DEQs enables more efficient training, we focus on work that does explicit backpropagation rollouts \citep{geiping_scaling_2025, mcleish_retrofitted_recurrence, bansalEndtoendAlgorithmSynthesis2022, yang2024loopedtransformersbetterlearning}.
Within looped architectures, \citet{geiping_scaling_2025,mcleish_retrofitted_recurrence} adopt the usage of path independence from equilibrium models \citep{anilPathIndependentEquilibrium} to warrant their choice of $h_0$ initialization.

%% file: appendix/derivation.tex
\section{Derivation of Instability Conditions of Prior Methods}
\label{sec:derivation-instabilty}

Recall from \cref{sec:rdm-basics}, that \recurrent denotes the full recurrent update $h_{t+1} = \mathcal{R}(h_t,e)$, encompassing all transformer operations, including residual connections. 
A common interpretation views the residual stream as a communication channel where $h_T$ is the sum of the relative outputs of all previous layers and the original embedding \cite{olsson2022incontextlearninginductionheads}. 
Applying this to looped models, let $\overline{\mathcal{R}}$ denote the \emph{relative contribution} of the nonlinear operations (i.e., $\overline{\mathcal{R}}(W_1 h_t + W_2 e) = \mathcal{R}(W_1 h_t + W_2 e) - (W_1 h_t + W_2 e)$). 
This gives the recurrent update rule
$h_{t+1} = W_1 h_t + W_2 e + \overline{\mathcal{R}}(h_t, e)$
where we write $\overline{\mathcal{R}}(h_t, e) = \overline{\mathcal{R}}(W_1 h_t + W_2 e)$ for brevity. 
Although $\overline{\mathcal{R}}$ is highly non-linear, the recurrent update can be exactly formulated as a \emph{non-linear time-variant dynamical system} of the form:
$h_{t} = \dA h_{t-1} + \dB e + \overline{\mathcal{R}}(h_{t-1}, e), ~ y_t = \C h_t,$
where $\dA = W_1$, $\dB = W_2$, and $\C \in R^{d_c \times d_h}$ decouples the \coda and \recurrent embedding dimension (i.e.~$p=\mathcal{C}(\C(h_T))$).

Using the \emph{relative contribution} representation of looped models above, we can recast
prior mediums of input injection discussed in \cref{sec:rdm-basics} in a form similar to our framework.
Specifically, for Pre-Norm looped models using addition as injection \cite{yangLoopedTransformersAre2023}, the dynamical systems update rule can thus be written in the form $h_{t+1} = Ih_t + Ie + \overline{\mathcal{R}}(Ih_t + Ie)$.
When linearized (i.e., dropping the nonlinear $\overline{\mathcal{R}}$ block), $\dA = I$, meaning that the model is a \emph{marginally-stable} system as all eigenvalues are 1. Alternatively, the update rule for Pre-Norm looped models using concatenation as injection \cite{geiping_scaling_2025} can be rewritten in the form $h_{t+1} = W[h_t;e] + \overline{\mathcal{R}}(W[h_t;e]) = W_1 h_t + W_2 e + \overline{\mathcal{R}}(W_1 h_t + W_2e)$. Here $\dA = W_1$ is unbounded and thus can create an explosion of the state if not carefully maintained during training.

%% file: appendix/flop-estimate.tex
\section{FLOP Estimate of Parcae}
\label{sec:flop-estimate}

In standard, fixed-depth architectures, a common means to approximate the number of FLOPs used in training is $C = 6ND$ from \citet{kaplan2020scalinglawsneurallanguage}, where $N$ is the number of parameters and $D$ is the number of tokens used in training. However, looped architectures differ from traditional models in that they exhibit the notion of \emph{effective parameters} $\hat N$ (e.g., for a model that is a single layer with $N$ parameters, if it is looped ten times, then it has an effective parameterization of $\hat N = 10 N$). Furthermore, as Parcae uses truncated backpropagation through depth, the effective parameters can thus be decoupled into two types: $\hat N_1$, which are effective parameters that \emph{\textbf{are not backpropagated}} through, and $\hat N_2$, which are effective parameters that \emph{\textbf{are backpropagated}} through. Thus, following \citet{kaplan2020scalinglawsneurallanguage}, we can formulate the effective FLOPs of Parcae as $C = (2\hat N_1 + 6 \hat N_2)D$, which further matches the setup of \citet{mcleish_retrofitted_recurrence}. Like \citet{mcleish_retrofitted_recurrence}, we exclude embedding parameters from $\hat N$, however, we do include unembedding parameters in $\hat N$ similar to \citet{nanochat}. Lastly, we additionally include an estimate for attention FLOPs following \citet{chowdhery2022palmscalinglanguagemodeling,nanochat}.

%% file: appendix/training-algorithm.tex
\section{Parcae Forward Pass and Training Algorithms}
\label{sec:algorithm}

A full forward pass of Parcae, combining our dynamical systems blocks $\A, \B, \C, \dt$ and looped models \prelude, \recurrent, \coda blocks can be found in \cref{alg:parcae}.

\begin{algorithm}[!ht]
\caption{Parcae Forward Pass}
\label{alg:parcae}
\begin{algorithmic}[1]
\Require Input sequence $s \in V^n$ and recurrent steps $T$.
\State $e \gets \text{LN}(\mathcal{P}(s))$ 
\State $h_0 \sim \mathcal{N}(0, \sigma^2 I_{n \times d})$ 
\State $\overline{\A}, \overline{\B} \gets \A, \B, \dt$ 
\For{$t = 1$ to $T$}
    \State $h_t \gets \overline{\A} h_{t-1} + \overline{\B} e + \overline{\mathcal{R}}(h_t, e)$ 
\EndFor
\State \textbf{return} $\mathcal{C}(\C h_T)$
\end{algorithmic}
\end{algorithm}

We display our algorithm to sample per-sequence depths during Parcae training while maintaining compute efficiency in Algorithm~\ref{alg:per-sequence-depth}. We do per-sequence depth sampling, but taking the max depth within a batch and performing no state updates at the \emph{beginning} of the recurrent computation. This allows for batched processing of different depths while maintaining efficient gradient flow.

\begin{algorithm}[!ht]
\caption{Efficient Per-Sequence Stochastic Depth Training}
\label{alg:per-sequence-depth}
\begin{algorithmic}[1]
\Require Batch of sequences $\{s_i\}_{i=1}^{B}$, means $\mu_{\text{rec}}, \mu_{\text{bwd}}$, and sampling distribution $\Lambda$
\State $\bm{e}^{(i)} \gets \mathcal{P}(s_i)$ for all $i$ \hfill  \textcolor{magenta}{$\triangleright$ \texttt{embed sequences}}
\State Sample $T^{(i)} \sim \Lambda(\mu_{\text{rec}})$ for each $i \in [B]$
\State $T_{\max} \gets \max_i T^{(i)}$, \quad $\tau^{(i)} \gets T_{\max} - T^{(i)}$
\State $\bm{h}_0^{(i)} \sim \mathcal{N}(0, \sigma \bm{I})$ for all $i$
\State $\overline{\bm{A}}, \overline{\bm{B}} \gets \textsc{Discretize}(\bm{A}, \bm{B}, \dt)$
\For{$t = 0, \ldots, T_{\max} - 1$}
    \State \textbf{for all} $i$ \textbf{where} $t < \tau^{(i)}$: \quad $\bm{h}_{t+1}^{(i)} \gets \bm{h}_t^{(i)}$ \hfill  \textcolor{magenta}{$\triangleright$ \texttt{no state update}}
    \State \textbf{for all} $i$ \textbf{where} $\tau^{(i)} \leq t < T_{\max} - \mu_{\text{bwd}}$: \hfill  \textcolor{magenta}{$\triangleright$ \texttt{without gradients}}
    \State \quad $\bm{h}_{t+1}^{(i)} \gets \overline{\bm{A}} \bm{h}_t^{(i)} + \overline{\bm{B}} \bm{e}^{(i)} + \mathcal{R}(\bm{h}_t^{(i)}, \bm{e}^{(i)})$
    \State \textbf{for all} $i$ \textbf{where} $t \geq T_{\max} - \mu_{\text{bwd}}$: \hfill  \textcolor{magenta}{$\triangleright$ \texttt{with gradients}}
    \State \quad $\bm{h}_{t+1}^{(i)} \gets \overline{\bm{A}} \bm{h}_t^{(i)} + \overline{\bm{B}} \bm{e}^{(i)} + \mathcal{R}(\bm{h}_t^{(i)}, \bm{e}^{(i)})$
\EndFor
\State \textbf{return} $\{\mathcal{C}(\bm{C} \bm{h}_{T_{\max}}^{(i)})\}_{i=1}^{B}$
\end{algorithmic}
\end{algorithm}

%% file: appendix/stability.tex
\section{Additional Stability Ablations}
\label{sec:stability-ablations}

\begin{figure}
    \centering
    \includegraphics[width=\linewidth]{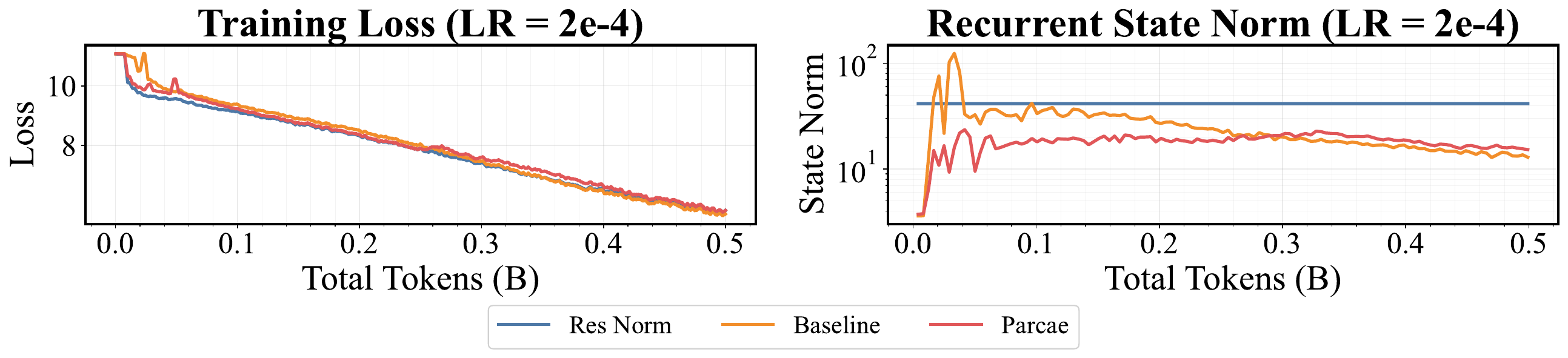}
    \includegraphics[width=\linewidth]{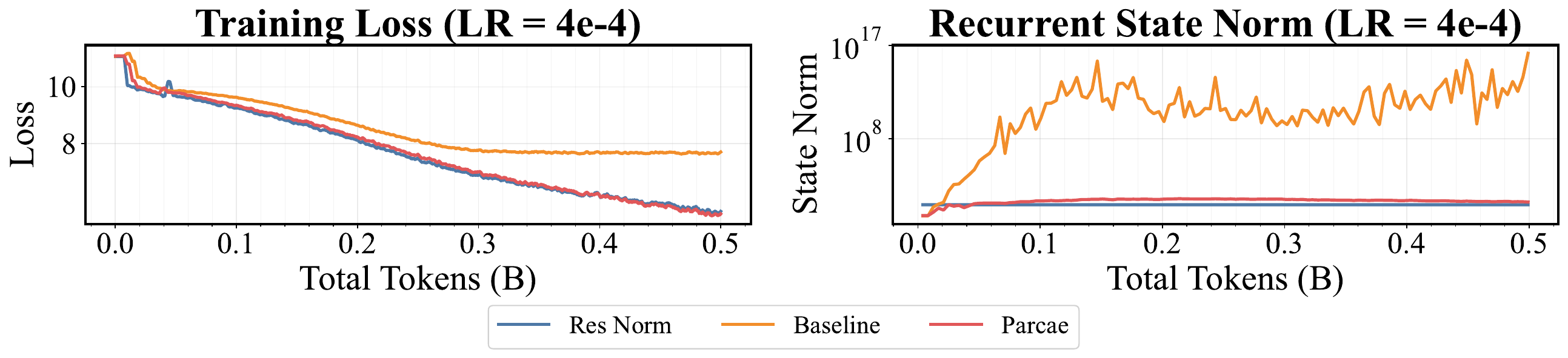}
    \includegraphics[width=\linewidth]{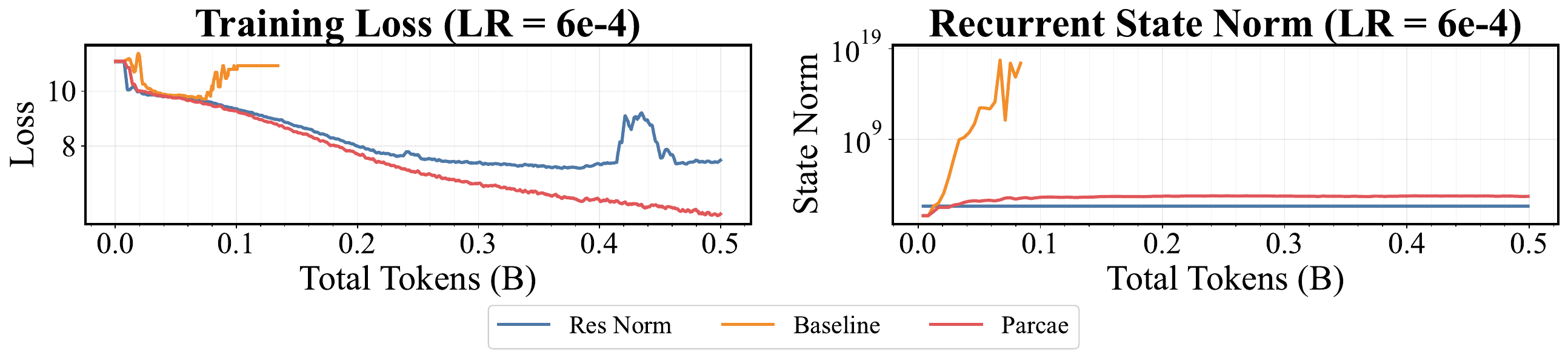}
    \includegraphics[width=\linewidth]{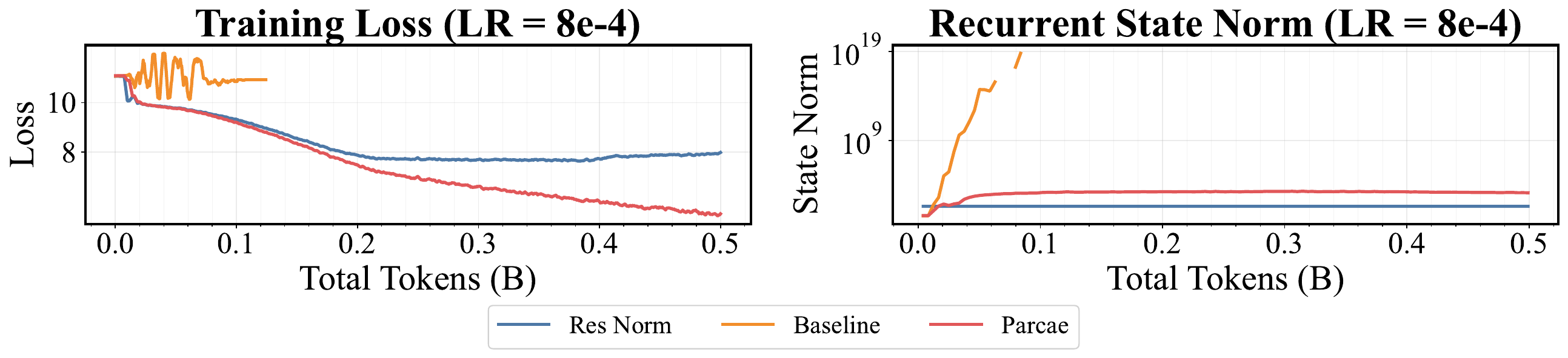}
    \includegraphics[width=\linewidth]{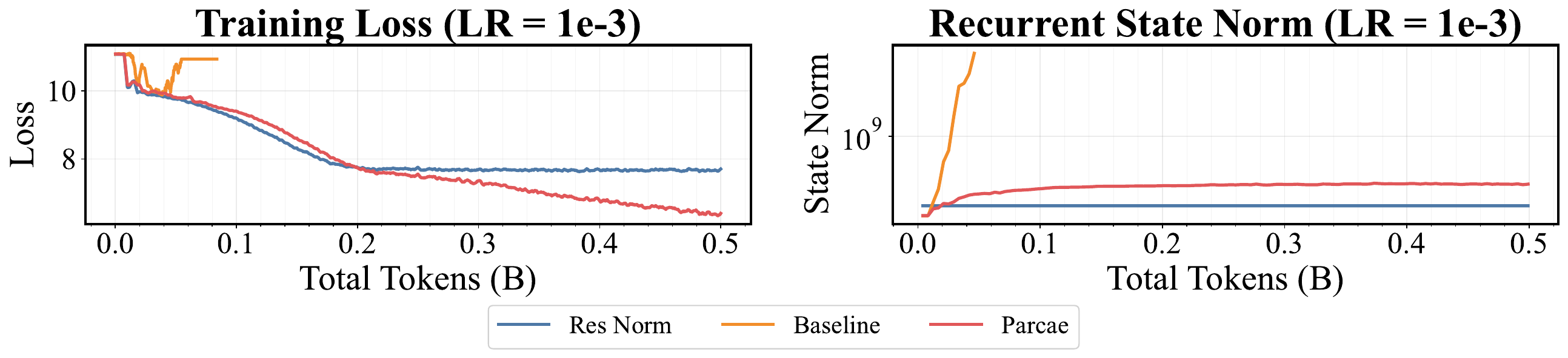}
    \caption{Training instability of recurrent depth models across different learning rates. We show both training losses and recurrent state norm to understand divergence and state explosion.}
    \label{fig:all-instability-curves}
\end{figure}

We include all training curves for our hyperparameter sweep experiments in \cref{sec:hyperparameters}. We conduct a learning rate sweep over $\{ 2e-4, 4e-4, 6e-4, 8e-4, 1e-3\}$ observing that Parcae exhibits stable training over both baseline Pre-Norm RDMs and residual normalized RDMs. The training curves and the accompanying recurrent state norm can be observed in \cref{fig:all-instability-curves}.

%% file: appendix/loss-spikes.tex
\newpage
\section{Per-sequence Sampling Reduces Loss Spikes}
\label{sec:loss-spikes}

\begin{figure}[!ht]
    \centering
    \includegraphics[width=\linewidth]{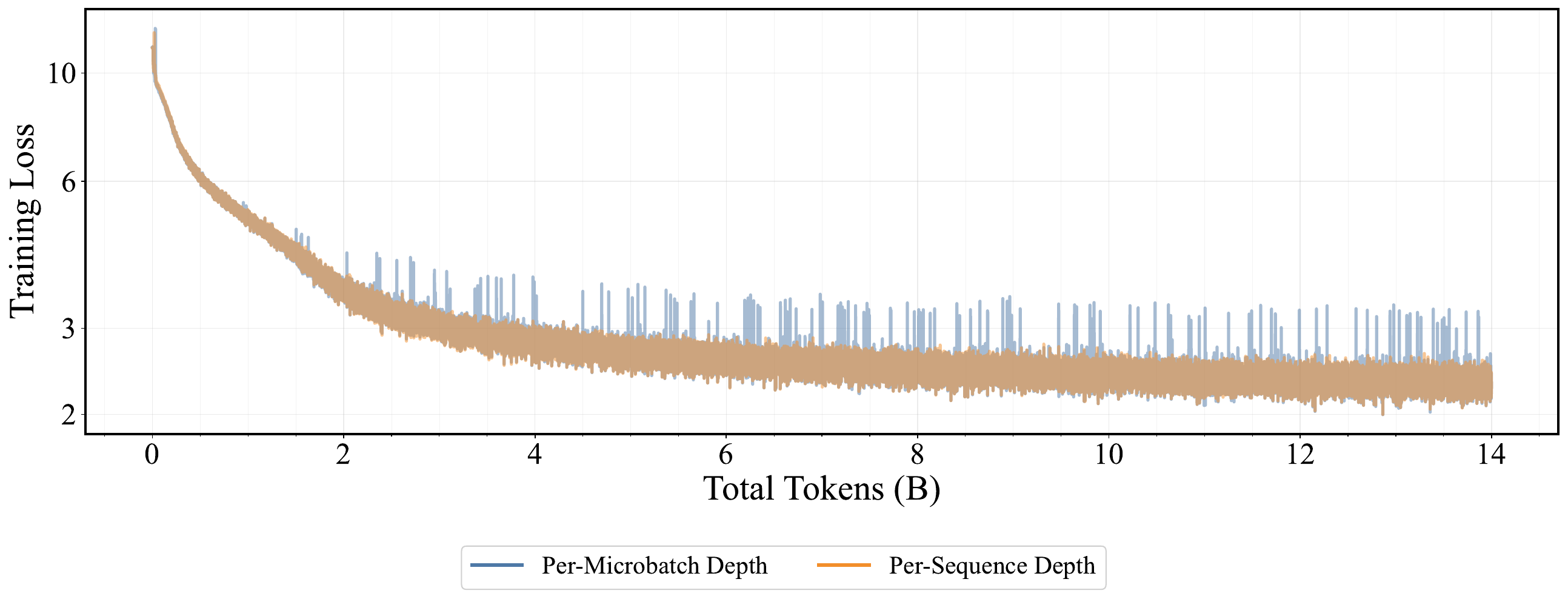}
    \vspace{-2.5em}
    \caption{Training curves showing per-sequence sampling effectively eliminates loss spikes in training over per-micro-batch sampling.}
    \label{fig:per-sample-training}
\end{figure}

\begin{figure}[!ht]
    \centering
    \vspace{-1.5em}
    \includegraphics[width=\linewidth]{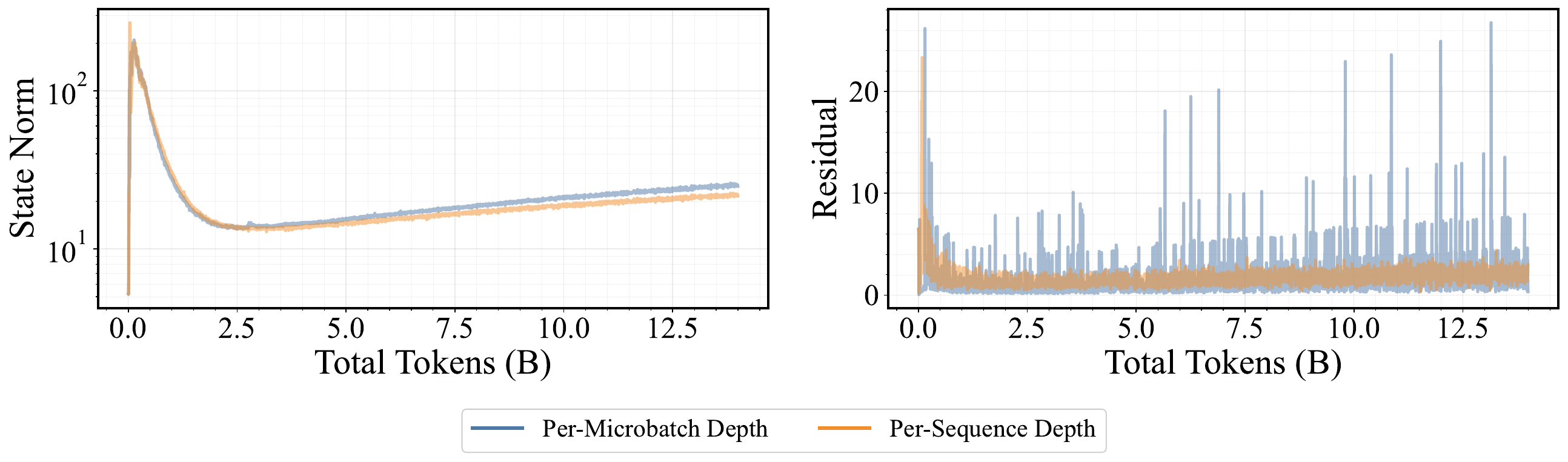}
    \vspace{-2em}
    \caption{Comparison of recurrent residual and state norm metrics (defined in \cref{sec:notation}), which show that per-sequence sampling enables stronger fixed point behavior in training.}
    \label{fig:residual-state-norm}
\end{figure}

When running our per-sequence sampling experiments, we observed that the training curves of per-sequence sampling helped eliminate loss spikes during training. Specifically, in \cref{fig:per-sample-training}, for our 350M parameter Parcae models, per-micro-batch has several loss spikes through training while per-sequence sampling does not. We can observe from \cref{fig:residual-state-norm}, that these training spikes stem directly from overly large recurrent residual jumps at the final recurrence, implying the model is not learning to converge to a steady-state fixed point solution. 
It can then be observed that per-sequence depth helps provide a better estimate for our training objective, enabling convergent fixed-point behavior and preventing loss-spikes during training. The direct benefit of this can be observed in \cref{tab:training-res}, where per-sequence sampling significantly improves the downstream quality of looped models, especially at low test-time recurrences. Finally, we note that per-sequence sampling adds a minimal amount of training overhead, increasing total wall clock time for pretraining by 1.8\%, which we believe can be further optimized away with a cleaner implementation.

\begin{table}[!ht]
        \vspace{0pt}
        \small
        \renewcommand{\arraystretch}{1.2} 
        \centering
        \begin{tabular}{@{}lccccc@{}}
        \toprule
        & \textbf{Method} & \textbf{T=1} & \textbf{T=4} & \textbf{T=8} & \textbf{T=16} \\
        \midrule
        \multirow{2}{*}[0pt]{\rotatebox{90}{\small{100M}}} & Per-Batch & 300.32 & 36.75 & 16.65 &  13.81 \\
                                                    & Per-Sequence & \textbf{70.47} & \textbf{17.15} & \textbf{14.08} &  \textbf{13.59} \\
        \midrule
        \multirow{2}{*}[0pt]{\rotatebox{90}{\small{350M}}} & Per-Batch & 167.61 &  12.80 &  10.40 &  10.24  \\
                                                    & Per-Sequence & \textbf{17.92} & \textbf{10.49} &  \textbf{10.09} &  \textbf{10.11} \\
        \bottomrule
        \end{tabular}
        \caption{\textbf{Per-Microbatch vs. Per-Sequence Comparison}. We compare perplexity of Parcae models trained with per-microbatch sampling \citep{geiping_scaling_2025} and per-sequence sampling, using different recurrences ($T$) on a held-out validation set. \textbf{Bolded} results indicate \textbf{best} at each scale.}
        \label{tab:training-res}
\end{table}

%% file: appendix/truncated.tex
\clearpage
\section{Sampling of Truncated Recurrence}
\label{sec:sampling-truncated-recurrence}

\begin{figure}[!h]
\centering
\begin{minipage}[t]{0.48\textwidth}
\begin{algorithm}[H]
\caption{\textsc{\citet{geiping_scaling_2025}}}
\label{alg:poisson-fill}
\begin{algorithmic}[1]
\State \textbf{Input:} $\mu_{\text{rec}}$, $\mu_{\text{bwd}}, \Lambda$, $e$
\State $n \sim \Lambda(\mu_{\text{rec}} - \mu_{\text{bwd}})$
\State $k \gets \mu_{\text{bwd}}$
\State $T = n + k$
\State $h_0 \gets \mathcal{N}(0, \sigma^2 I)$
\For{$t = 1$ \textbf{to} $T$}
    \If{$t \leq n$}
        \State $h_t \gets \mathcal{R}(h_{t-1}, e)$ \textbf{w/o grad}
    \Else
        \State $h_t \gets \mathcal{R}(h_{t-1}, e)$ \textbf{w/ grad}
    \EndIf
\EndFor
\State \Return $x_T$
\end{algorithmic}
\end{algorithm}
\end{minipage}
\hfill
\begin{minipage}[t]{0.48\textwidth}
\begin{algorithm}[H]
\caption{\textsc{Correction (Ours)}}
\label{alg:poisson-trunc-full}
\begin{algorithmic}[1]
\State \textbf{Input:} $\mu_{\text{rec}}$, $\mu_{\text{bwd}}$, $\Lambda$, $e$
\State $T \sim \Lambda(\mu_{\text{rec}})$
\State $n \gets \max(T - \mu_{\text{bwd}}, 0)$
\State $k \gets \min(T, \mu_{\text{bwd}})$
\State $h_0 \gets \mathcal{N}(0, \sigma^2 I)$
\For{$t = 1$ \textbf{to} $T$}
    \If{$t \leq n$}
        \State $h_t \gets \mathcal{R}(h_{t-1}, e)$ \textbf{w/o grad}
    \Else
        \State $h_t \gets \mathcal{R}(h_{t-1}, e)$ \textbf{w/ grad}
    \EndIf
\EndFor
\State \Return $x_T$
\end{algorithmic}
\end{algorithm}
\end{minipage}
\caption{Comparison of our sampling method with \cite{geiping_scaling_2025}. It can be observed that the actual distribution of forward recurrence for \cite{geiping_scaling_2025} is a shifted Poisson distribution. The implications of this sampling strategy can be better visualized in \cref{fig:method-distribution-mismatch}.}
\label{fig:sampling-algos}
\end{figure}

\begin{figure}[h]
    \centering
    \includegraphics[width=\linewidth]{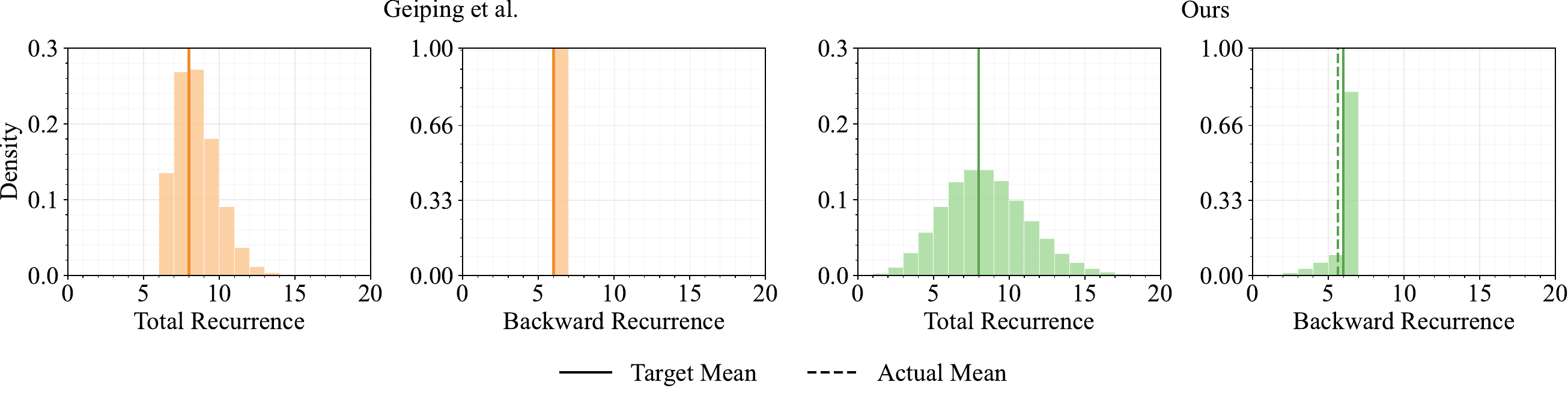}
    \caption{A distributional mismatch can be observed from the recurrent sampling method of \cite{geiping_scaling_2025}. Specifically, if our desired pre-training distribution for \meanrecurrence is a Poisson distribution, the distribution total recurrence $T$ of \cite{geiping_scaling_2025} is truncated based on \meanbackward. However, our sampling method decouples the effects of \meanbackward on $\Lambda$, allowing the recurrent distribution to be faithfully sampled from.}
    \label{fig:method-distribution-mismatch}
\end{figure}

In our very initial experiments, we observed that we could make a small change to the sampling algorithm of \cite{geiping_scaling_2025}, which stems from \cite{avi_learn_algorithm}, to enhance the training of Parcae\footnote{We make the same change to RDMs in the main body, observing that they perform better with it and to make comparison fair.}. When given an arbitrary distribution to sample from $\Lambda$ and two hyperparameters \meanrecurrence (the desired mean steps of the recurrent blocks in pre-training) and \meanbackward (the desired mean back-propagation steps in pre-training), we observe that previous work by \cite{geiping_scaling_2025} had a distributional mismatch. Previously, the sampling method of \cite{geiping_scaling_2025} exactly followed \cref{alg:poisson-fill} with a poisson log-normal distribution with the following distribution
\begin{align}
    \tau \sim \mathcal{N}(\log(\mu_{\text{rec}} - \mu_{\text{bwd}}) - \frac{1}{2}\sigma^2, \sigma) \qquad n \sim \mathcal{P}(e^\tau) + 1 \qquad k \gets \mu_{\text{bwd}}
\end{align}
where $\sigma = \frac{1}{2}$. To maintain a fixed computation memory budget, \cite{geiping_scaling_2025} sets $k$ to \meanbackward; however, this minor change significantly impacts the underlying recurrent distribution, truncating and compressing the distribution of recurrence actually observed during pre-training. We propose making a minor algorithmic fix to the sampling method, which can be observed in \cref{alg:poisson-trunc-full}. While minor, observe in \cref{fig:method-distribution-mismatch} the impact of improving generalization to other recurrences. 

\begin{figure}[t]
    \centering
    \includegraphics[width=\textwidth]{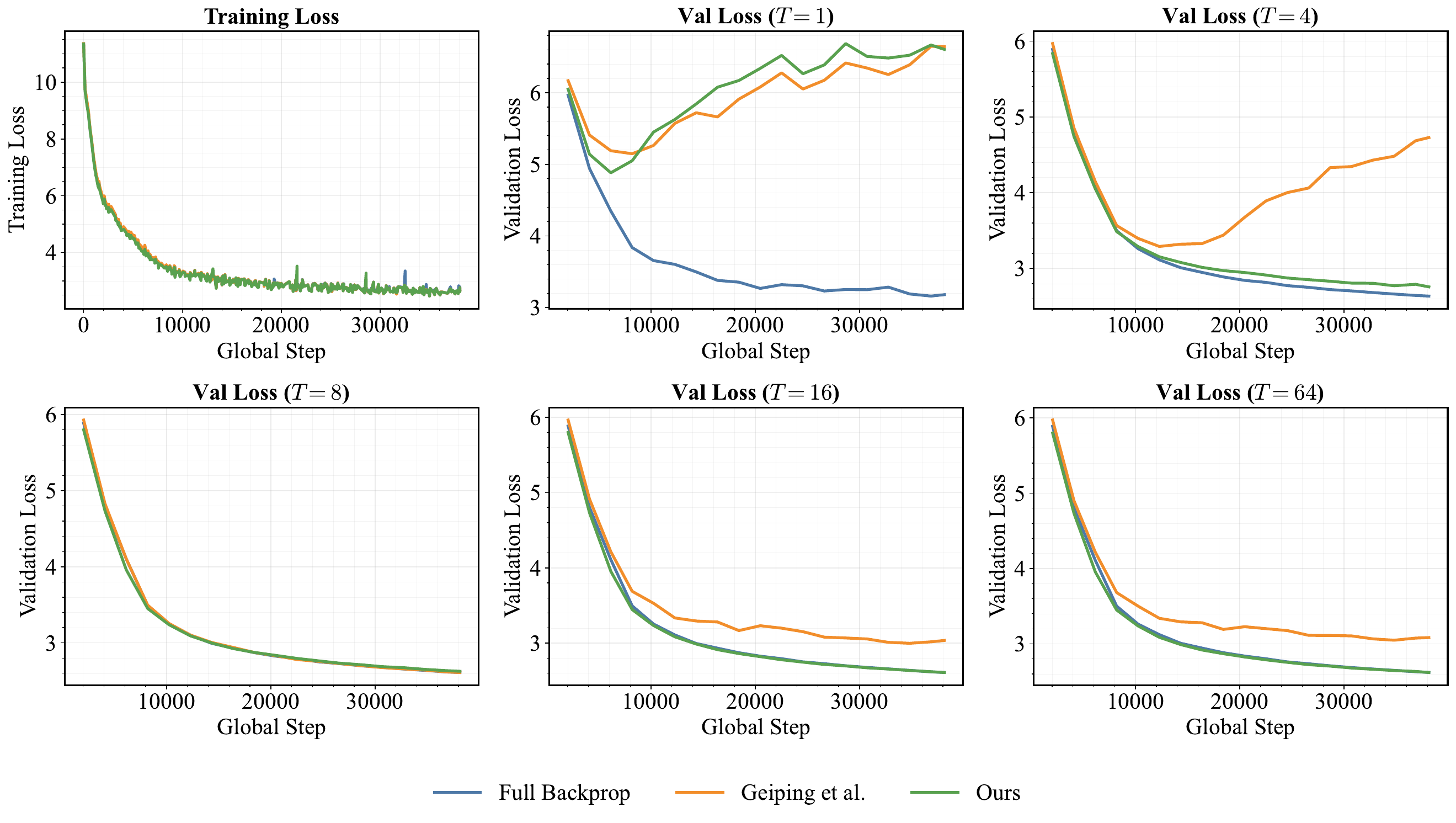}
    \caption{Training and validation curves of three 100 million parameter Parcae models pretrained on 10 billion tokens, comparing different truncated back-propagation methods (baseline is a model with no back-propagation truncation). Each model has identical architecture and hyperparameters, with \meanrecurrence and \meanbackward both being set to eight, all using $\Lambda \sim \text{Poisson}$. It can be observed that even though each model has similar training loss and validation loss when using $T=8$, our implementation more faithfully follows the validation loss of full back-propagation. Specifically for $T=4$, our implementation significantly improves validation loss compared to \cite{geiping_scaling_2025} sampling method.}
    \label{fig:sampling-mismatch-results}
    \vspace{-1em}
\end{figure}

To verify our change, we pretrain several small Parcae models on 10 billion tokens to ablate on our design choice. Specifically, we set $\mu_{\text{rec}} = \mu_{\text{bwd}} = 8$ and use $\Lambda \sim \text{Poisson}$, and use fixed architecture, hyperparameters, and data stream. We train three models: a baseline Parcae model that performs full backpropagation through recurrences, a Parcae model following \cref{alg:poisson-fill} by \cite{geiping_scaling_2025}, and a Parcae model following \cref{alg:poisson-trunc-full}. The results of this ablation can be found in \cref{fig:sampling-mismatch-results}.

From \cref{fig:sampling-mismatch-results}, observe that training trajectories and validation loss at $T=\mu_{\text{rec}}=8$ are almost identical for each run; however, our method significantly improves performance for the validation loss of $T \in [4,16,64]$. Simply put, the constricting effect \cite{geiping_scaling_2025} observed in \cref{fig:method-distribution-mismatch} reduces the effective range of recurrence seen in pretraining, hurting the validation loss of using more or fewer recurrence at test-time. 

%% file: appendix/choosing.tex
\newpage
\section[Selecting mu rec and mu bwd]{Selecting $\mu_\text{rec}$ and $\mu_{\text{bwd}}$}
\label{sec:scaling-of-truncated-backpropigation}

\begin{figure}[h]
    \centering
    \includegraphics[width=\linewidth]{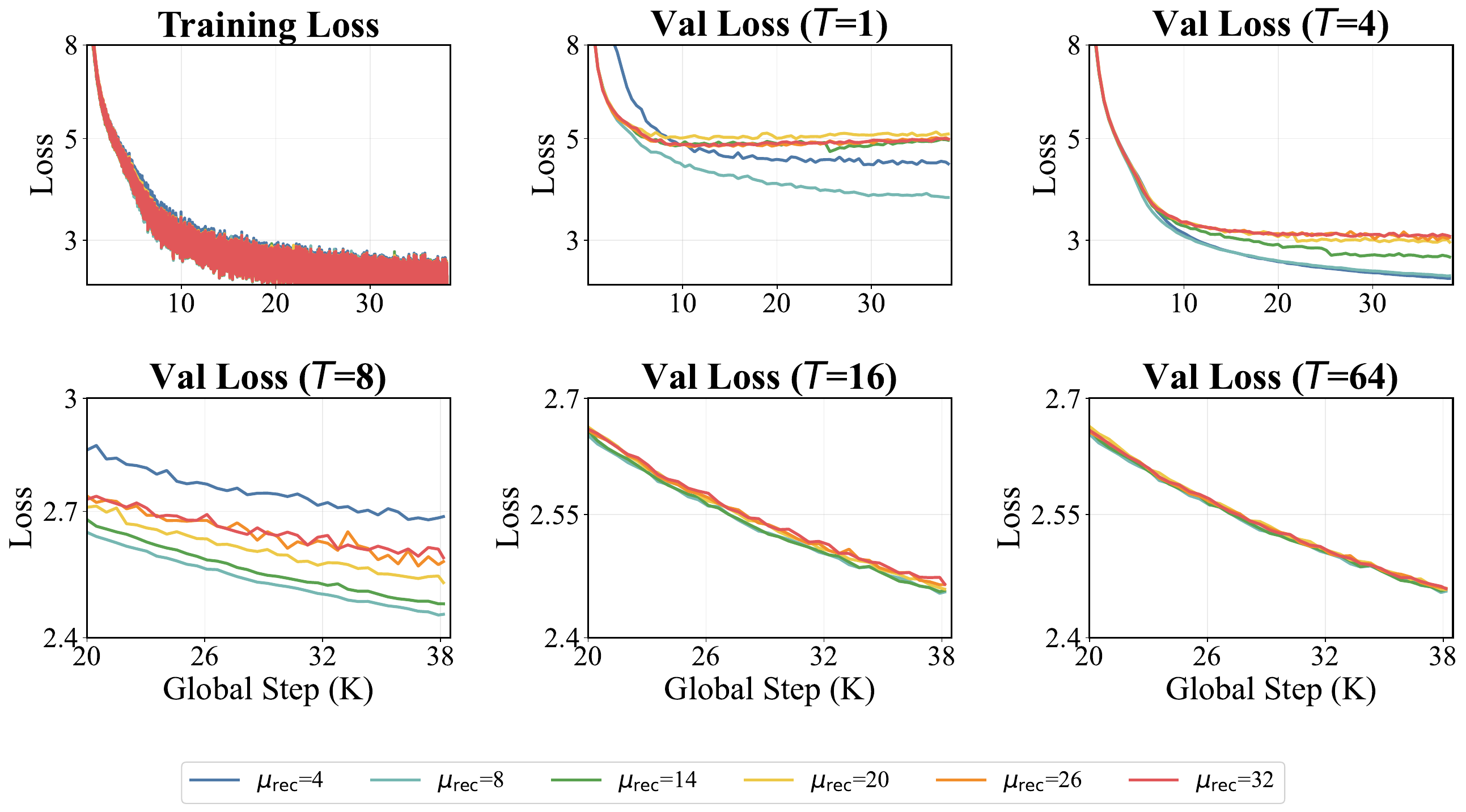}
    \caption{Validation curves of six different recurrent depth models, pretrained on 10 billion tokens, with a fixed architecture and hyperparameters. Each model is pretrained with a fixed \meanbackward of 8 and varying \meanrecurrence in $[4,8,14,20,26,32]$. The key observation is that scaling up \meanrecurrence while keeping \meanbackward fixed results in models that perform worse than if just pretrained on \meanrecurrence of eight.}
    \label{fig:mean-recurrence-scaling}
\end{figure}

\begin{table}[h]
    \centering
    \begin{tabular}{ccccccc}
    \toprule
         & $\mu_{\text{rec}}=4$ & $\mu_{\text{rec}}=8$ & $\mu_{\text{rec}}=14$ & $\mu_{\text{rec}}=20$ & $\mu_{\text{rec}}=26$ & $\mu_{\text{rec}}=32$\\
         \midrule
         Val Loss & 2.477 & \textbf{2.453} & 2.456 & 2.457 & 2.458 & 2.458 \\
         Val Perplexity & 11.906 & \textbf{11.624} & 11.665 & 11.671 & 11.692 & 11.687 \\
         \bottomrule
    \end{tabular}
    \caption{Validation loss and perplexity for looped models trained with different \meanrecurrence and a fixed $\mu_{\text{bwd}}=4$. We use $T=\mu_{\text{rec}}$. Surprisingly, $\mu_{\text{rec}}=8$ performs the best.}
    \label{tab:mean-forward-results}
\end{table}

A natural question is what choice of \meanrecurrence and \meanbackward is appropriate for pretraining looped models. To answer this question, we conduct an experiment where we scale up \meanrecurrence, while keeping \meanbackward fixed. In our very initial experiments, we pretrained several small recurrent depth models \citep{geiping_scaling_2025} on 10 billion tokens, with a fixed $\mu_{\text{bwd}}=4$ and with $\mu_{\text{rec}} \in [4,8,14,20,26,32]$\footnote{Note that these experiments were run before the distribution mismatch fix discussed in \cref{sec:sampling-truncated-recurrence}. As the mismatch becomes more drastic as \meanrecurrence gets closer to \meanbackward, we expect the model pretrained with $\mu_{\text{rec}}=4$ to be performing sub-optimally.}. The results for each of these models on a held-out set of validation data can be observed in \cref{fig:mean-recurrence-scaling}. We additionally include \cref{tab:mean-forward-results}, which gives the validation loss of each model with $\mu_{\text{rec}} \in [4,8,14,20,26,32]$, where the recurrence that we use for each model at test-time is $T=\mu_{\text{rec}}$. 

\begin{figure}[h]
    \centering
    \includegraphics[width=\linewidth]{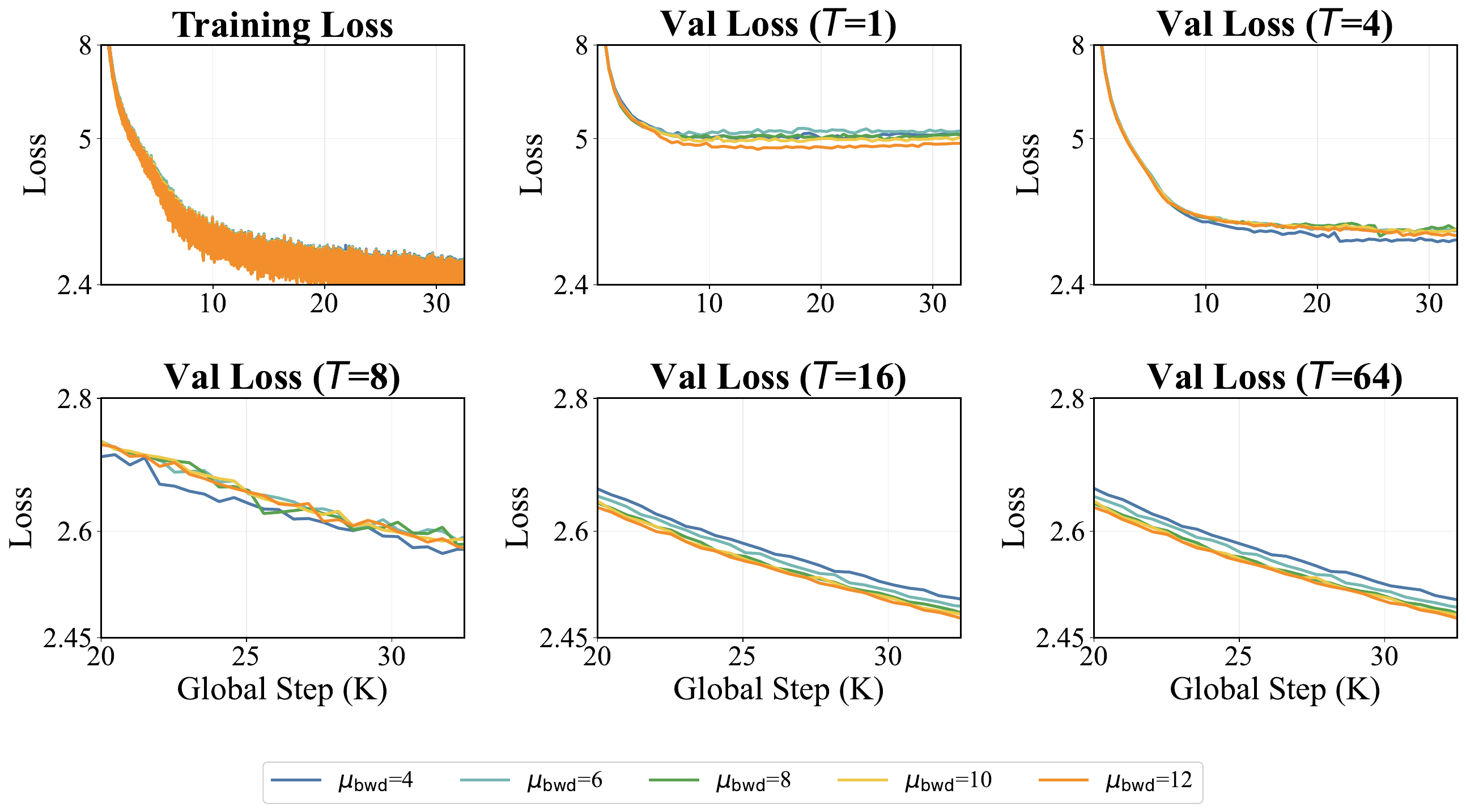}
    \caption{Validation and training curves of looped models, pretrained on 8.5 billion tokens. Each model is trained with a fixed $\mu_{\text{rec}}=20$ and $\mu_{bwd} \in [4,6,8,10,12]$. Observe that scaling up \meanbackward improves validation performance at higher and lower recurrences monotonically for $T=1,16,64$.}
    \label{fig:mean-backward-recurrence-scaling}
\end{figure}

The fascinating observation of \cref{fig:mean-recurrence-scaling} is that, contrary to our initial beliefs, models trained with additional \meanrecurrence beyond 8 perform worse at both lower and higher $r$ used at test-time, though more FLOPs were spent during pretraining. While it is a natural expectation that models trained with lower \meanrecurrence perform better than models with larger \meanrecurrence at low $T$, the fact that a \meanrecurrence of eight performs the best at higher $T$ (i.e., $T=16$ and $T=64$) is surprising. To determine if this is an inherent limitation of the capacity of looped models or an artifact of \meanbackward, we ran an additional experiment where we fixed $\mu_{rec}=20$ and instead varied $\mu_{\text{bwd}} \in [4,6,8,10,12]$, pretraining on 8.5 billion tokens for each model. We keep hyperparameters fixed. The results for each of these models on a held-out set of validation data can be visualized in \cref{fig:mean-backward-recurrence-scaling} and \cref{tab:mean-backward-results}.

\begin{table}[h]
    \centering
    \begin{tabular}{cccccc}
    \toprule
         & $\mu_{\text{bwd}}=4$ & $\mu_{\text{bwd}}=6$ & $\mu_{\text{bwd}}=8$ & $\mu_{\text{bwd}}=10$ & $\mu_{\text{bwd}}=12$ \\
         \midrule
         Val Loss & 2.500 & 2.490 & 2.480 & 2.479 & \textbf{2.474} \\
         Val Perplexity & 12.09 & 12.06 & 11.94 & 11.93 & \textbf{11.86} \\
         \bottomrule
    \end{tabular}
    \caption{Validation loss and perplexity of looped models trained with variable \meanbackward, but fixed \meanrecurrence.}
    \label{tab:mean-backward-results}
\end{table}

While lower \meanbackward (i.e., $\mu_{\text{bwd}}=4,6,8$) appears to perform better with lower validation recurrences than higher \meanbackward, the validation loss using $T=16,64$ improves as \meanbackward increases. This implies that the capabilities of looped models utilizing deeper recurrences are heavily coupled with \meanbackward. However, it can be observed that increasing \meanbackward from ten to twelve has minimal impact on validation performance, at the cost of higher pretraining FLOPs. Using this insight, for our main training runs, we choose to use 
\begin{equation}
\mu_{\text{bwd}} = \lceil \frac{\mu_{\text{rec}}}{2} \rceil
\label{equation:mean-backward}
\end{equation}
We leave the exploration of FLOP optimal choices of \meanrecurrence and \meanbackward to future work.

%% file: appendix/prelude-norm.tex
\clearpage
\section{Ablation of Prelude Normalization}
\label{sec:prelude-norm}

\begin{figure}[!h]
    \centering
    \vspace{-0.5em}
    \includegraphics[width=\linewidth]{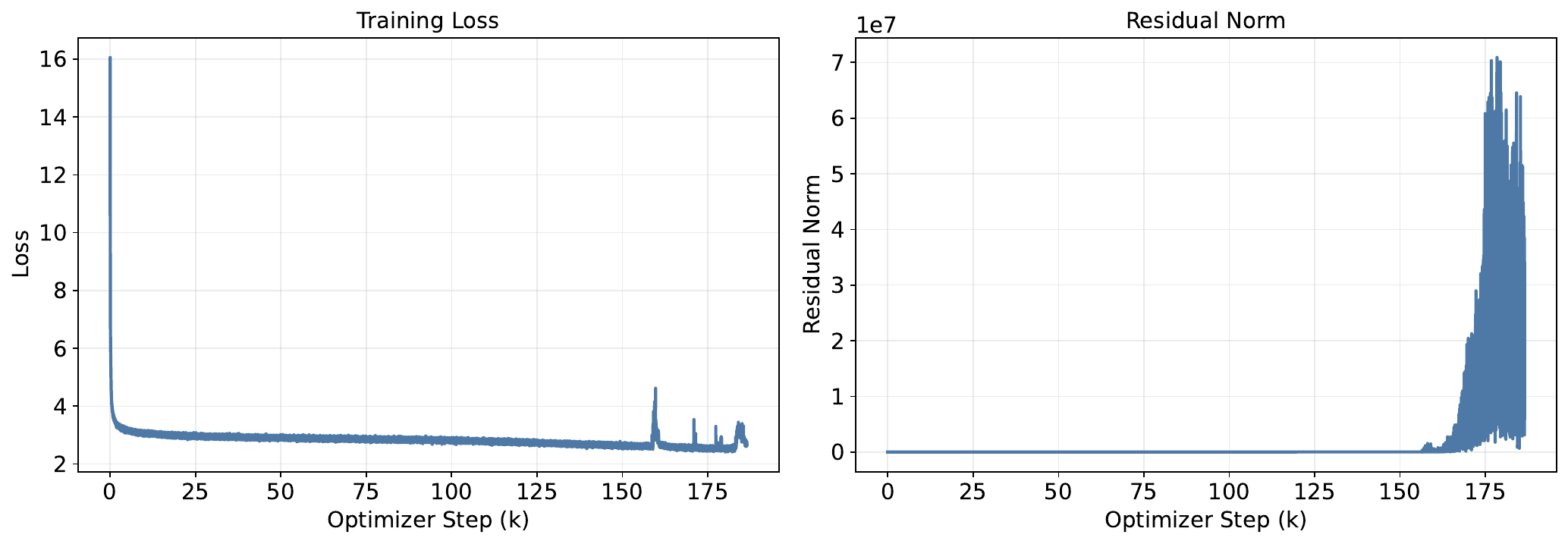}
    \vspace{-2em}
    \caption{\textbf{Late Stage Instability of 1.3B Parcae models.} We observe loss spikes and state explosion at the final stages of our large-scale run.}
    \label{fig:prelude-instability}
\end{figure}

In our initial set of experiments, we found that Parcae was able to train stably on the 140M, 370M, and 770M model configurations. Unfortunately, at the 1.3B scale, training appeared stable for the first 150k optimizer steps, afterwards exhibited state explosion and loss spikes, an observation which can be made in \cref{fig:prelude-instability}. To diagnose and fix these issues, we performed a deep exploration of the weight checkpoints before and during loss spikes, investigating both dynamical systems parameters (e.g., $\A, \B, \C, \dt$) and non-linear parameters $\overline{\mathcal{R}}$.

\begin{figure}[!h]
    \centering
    \vspace{-0.5em}
    \includegraphics[width=\linewidth]{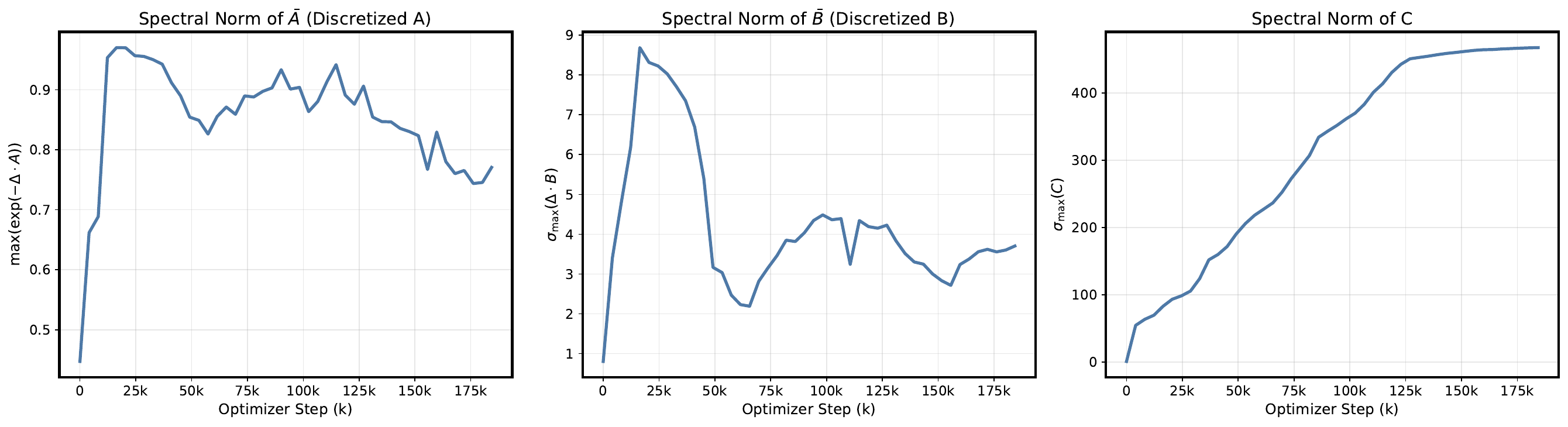}
    \vspace{-2em}
    \caption{\textbf{Spectral Norms of $\dA, \dB, \C$ throughout training 1.3B Parcae.} We find that the spectral norm of $\dA$ and $\dB$ remain stable throughout training, while the spectral norm of $\C$ grows.}
    \label{fig:abc_spectral}
\end{figure}

\begin{figure}[!h]
    \centering
    \includegraphics[width=\linewidth]{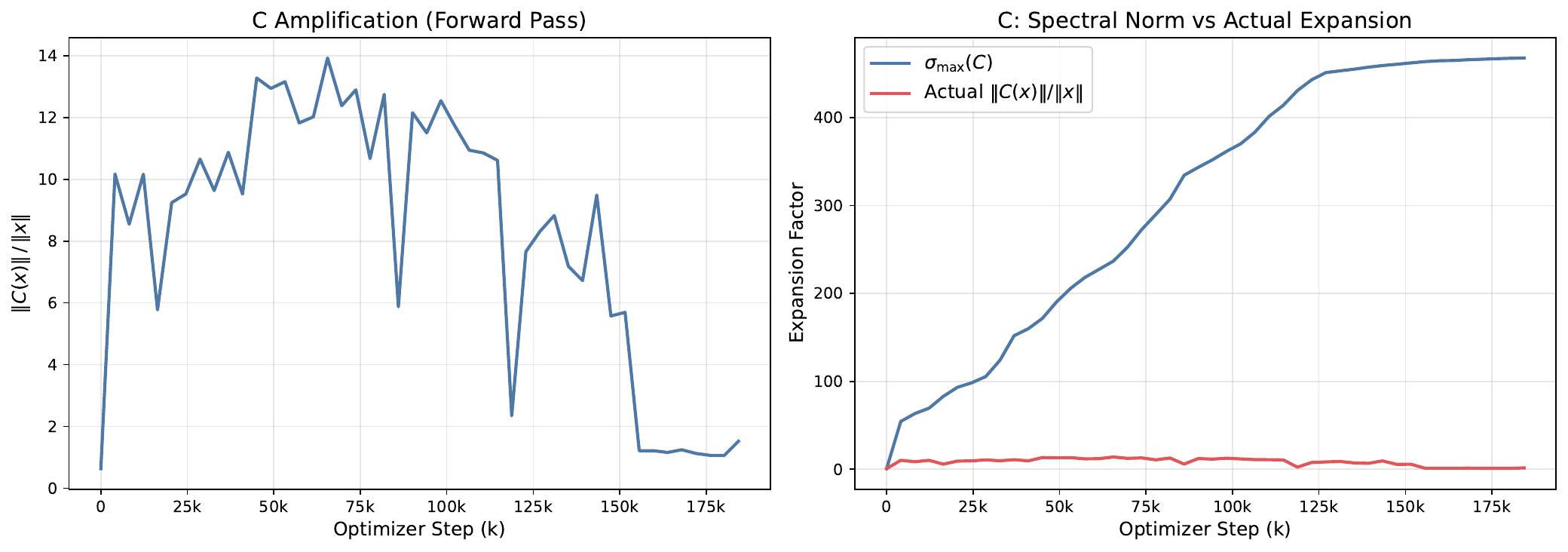}
    \caption{\textbf{Comparison of $\C$ Amplification with Spectral Norm.} We observe that the actual expansion ratio of $\C$ is small and decreasing slowly throughout training.}
    \label{fig:c_expansion}
\end{figure}

We begin by exploring the spectral norm of $\dA$, $\dB$, $\C$ to see if our dynamical systems block was creating instability, results of which can be found in \cref{fig:abc_spectral}. While we observe that the spectral norm remains relatively low for $\dA$ and $\dB$, we observed that the spectral norm of $\C$ grew significantly throughout training. While this could be concerning, we find that when passing real activations to $\C$, using a subset of the validation set, the empirical expanse ratio $\frac{||C(x)||}{||x||}$ (i.e., how much the norm of the residual $x$ grew after performing $\C(x)$) remained relatively low, as seen in \cref{fig:c_expansion}.

\begin{figure}[!h]
    \centering
    \includegraphics[width=\linewidth]{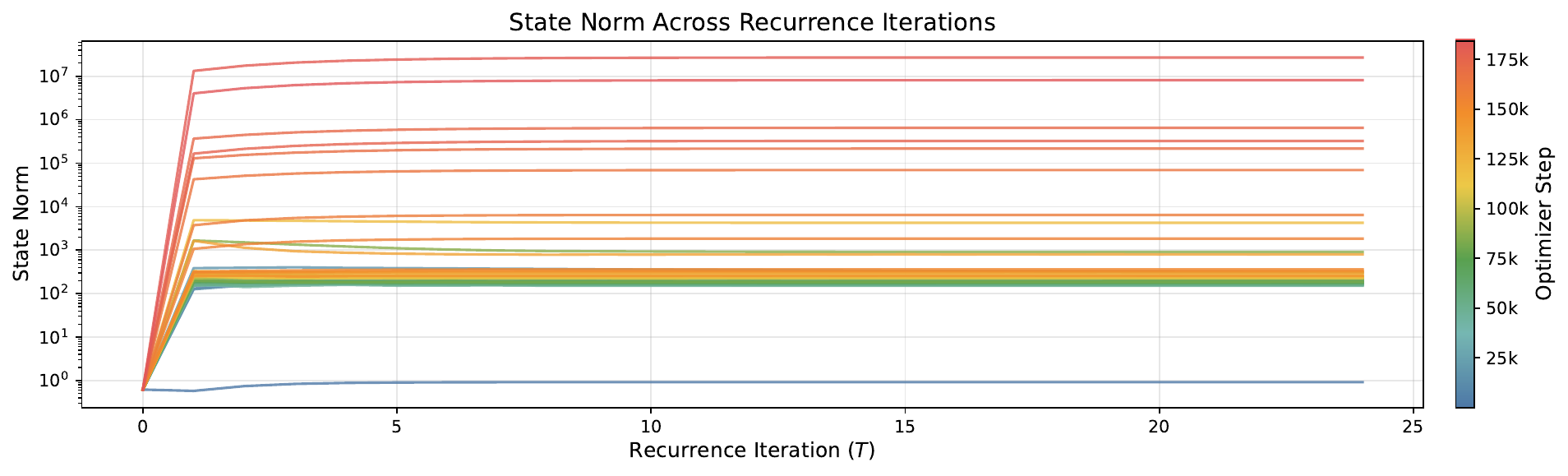}
    \caption{\textbf{Empirical Average of Recurrent State Norm over $T$ iterations.} For each checkpoint we have for our failed 1.3B Parcae model run, we evaluate the recurrent norm through $T=24$ recurrences at test time, on a held out validation set of fineweb-edu \cite{penedo2024finewebdatasetsdecantingweb}. We find that after an initial explosion on the first recurrence, the state remains relatively stable.}
    \label{fig:state_norm_diagnose}
\end{figure}

\begin{figure}[!t]
    \centering
    \includegraphics[width=\linewidth]{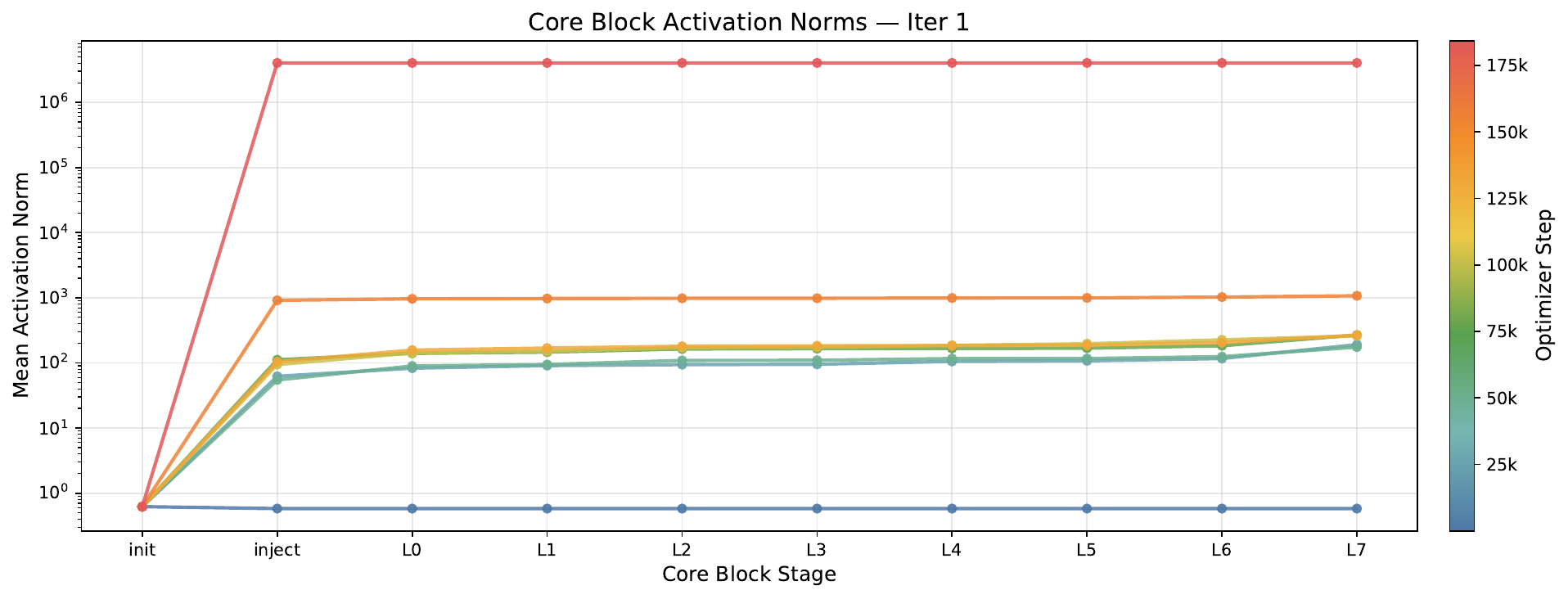}
    \caption{\textbf{Recurrent State Norm Progression After Each Transformer Block for $T=1$.} For each checkpoint we have for our failed 1.3B Parcae model run, we evaluate the recurrent norm after injection and each non-linear transformer block for only $T=1$. We find that the non-linear parts of Parcae have little effect on explosion, which instead mainly stems from the initial injection of prelude output $e$.}
    \label{fig:injection-explosion}
\end{figure}

\begin{figure}[!t]
    \centering
    \includegraphics[width=\linewidth]{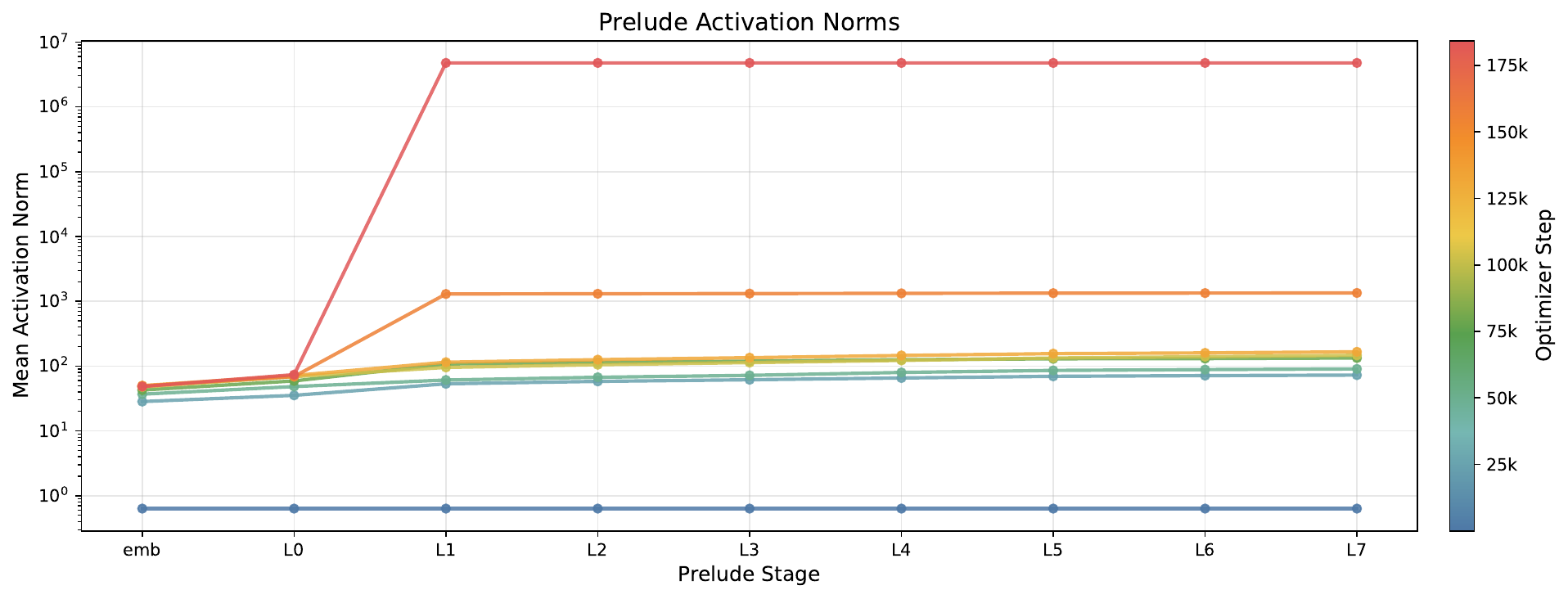}
    \caption{\textbf{State Norm Progression Throughout each Transformer Layer in the Prelude Block.} For each checkpoint we have for our failed 1.3B Parcae model run, we evaluate residual norm after each transformer block in the prelude \prelude. We find that a single layer creates an explosion of the residual norm and leads to divergence.}
    \label{fig:prelude-explosion}
\end{figure}

These results indicate that the dynamical systems units are likely not causing an explosion, and thus, we turn our exploration of the dynamics of the entire recurrent unit. Specifically, we track the recurrent state norm at test-time after $T=24$ recurrences, results of which can be found in \cref{fig:state_norm_diagnose}. We found that on the first recurrence, the recurrent state norm jumped drastically, and then remained relatively stable throughout increased recurrences. To determine what caused the initial spike, we perform a fine-grained analysis of the first recurrence (i.e., $T=1$), tracking the recurrent state norm after injection and through each transformer block, the results of which can be found in \cref{fig:injection-explosion}. The major takeaway from \cref{fig:injection-explosion} is that the non-linear parts of Parcae do not appear to cause the explosion in state and that the initial explosion steps from the input injections of $e$, the output of the prelude block \prelude. We confirm that this is the case, and visualization of which can be seen in \cref{fig:prelude-explosion}.

Given this, we propose a simple fix of adding a normalization layer on the output of the prelude block \prelude (i.e., for an input $x$ then $e \gets \text{LN}(\mathcal{P}(x))$, where $\text{LN}(\cdot)$ is some form of normalization). We note that this does two things: (1) normalizes the input to the recurrent unit, which we observe to further stabilize the recurrent dynamics of looping, and (2) stabilizes the gradient flow to the \prelude.\footnote{We do not directly prove or show this; however, it can be inferred by prior work on how normalization stabilizes forward and backward passes of transformers \cite{xu2019understandingimprovinglayernormalization, xiongLayerNormalizationTransformer2020}.} This simple fix enables our stable training run for the 1.3B Parcae reported in \cref{sec:results}.

\begin{figure}[!h]
    \centering
    \includegraphics[width=\linewidth]{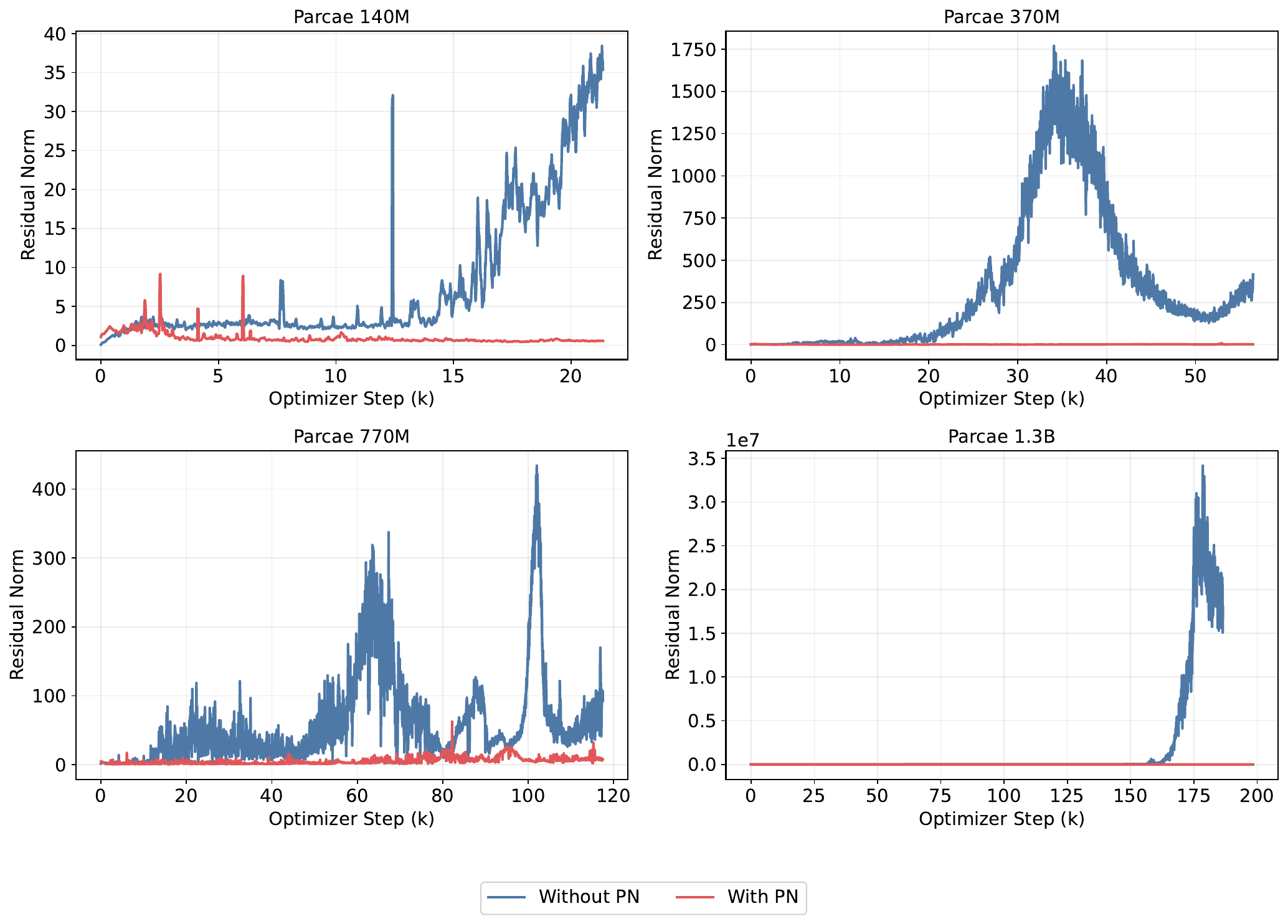}
    \vspace{-1em}
    \caption{\textbf{Prelude Norm Stabilizes Recurrent Norm.} We find that prelude norm helps stabilize recurrent state norm in Parcae models following the setup in \cref{sec:e2e} for Transformers.}
    \label{fig:prelude-norm-better}
\end{figure}

\begin{figure}[!h]
    \centering
    \vspace{-1em}
    \includegraphics[width=\linewidth]{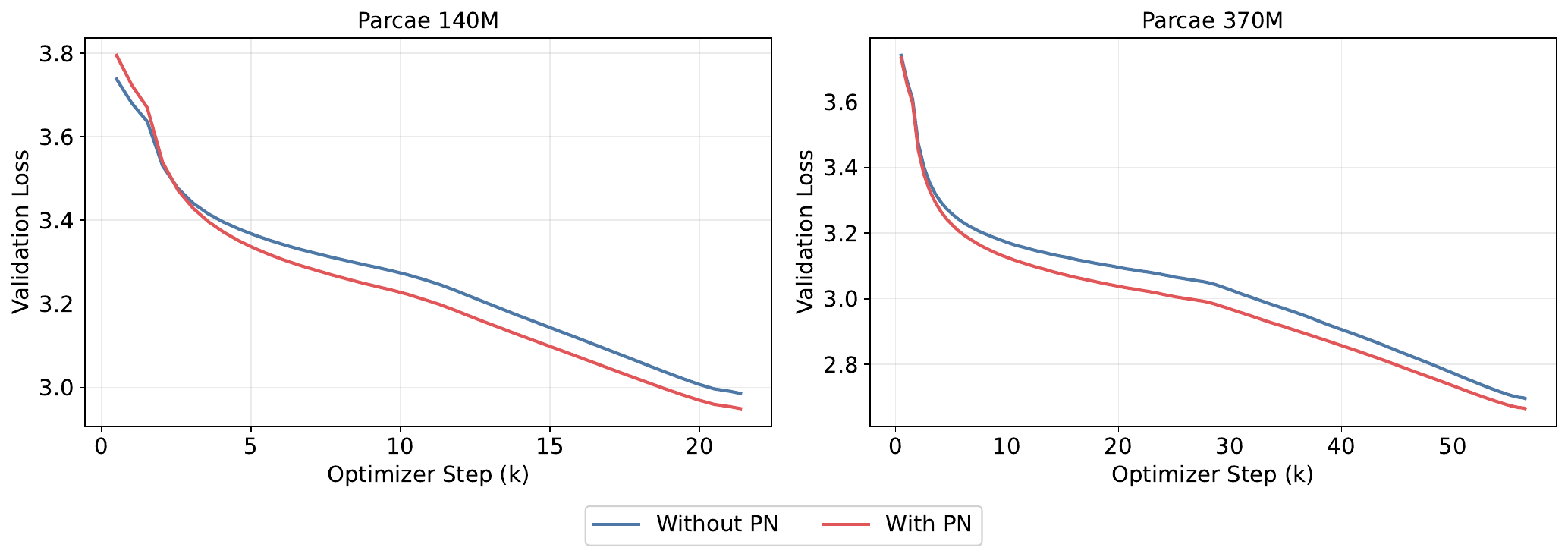}
    \vspace{-1em}
    \caption{\textbf{Prelude Norm Improves Quality.} We find that in our 140M and 370M Parcae models trained in the same setup as \cref{sec:e2e} for Transformers, normalizing the prelude output leads to better convergence.}
    \label{fig:prelude-quality}
\end{figure}

Empirically, we find that using a prelude norm directly stabilizes the recurrent norm further, preventing the recurrent norm from growing too large (see \cref{fig:prelude-norm-better}). Additionally, we find that using a prelude norm leads to better convergence in both our 140M and 370M Parcae models (see \cref{fig:prelude-quality}), with only a negligible improvement for our 770M and 1.3B Parcae models.

%% file: appendix/train-parametric.tex
\clearpage
\section{Fitting a Parametric Function for Looping}
\label{sec:fit-par}

\begin{figure}[!h]
    \centering
    \includegraphics[width=\linewidth]{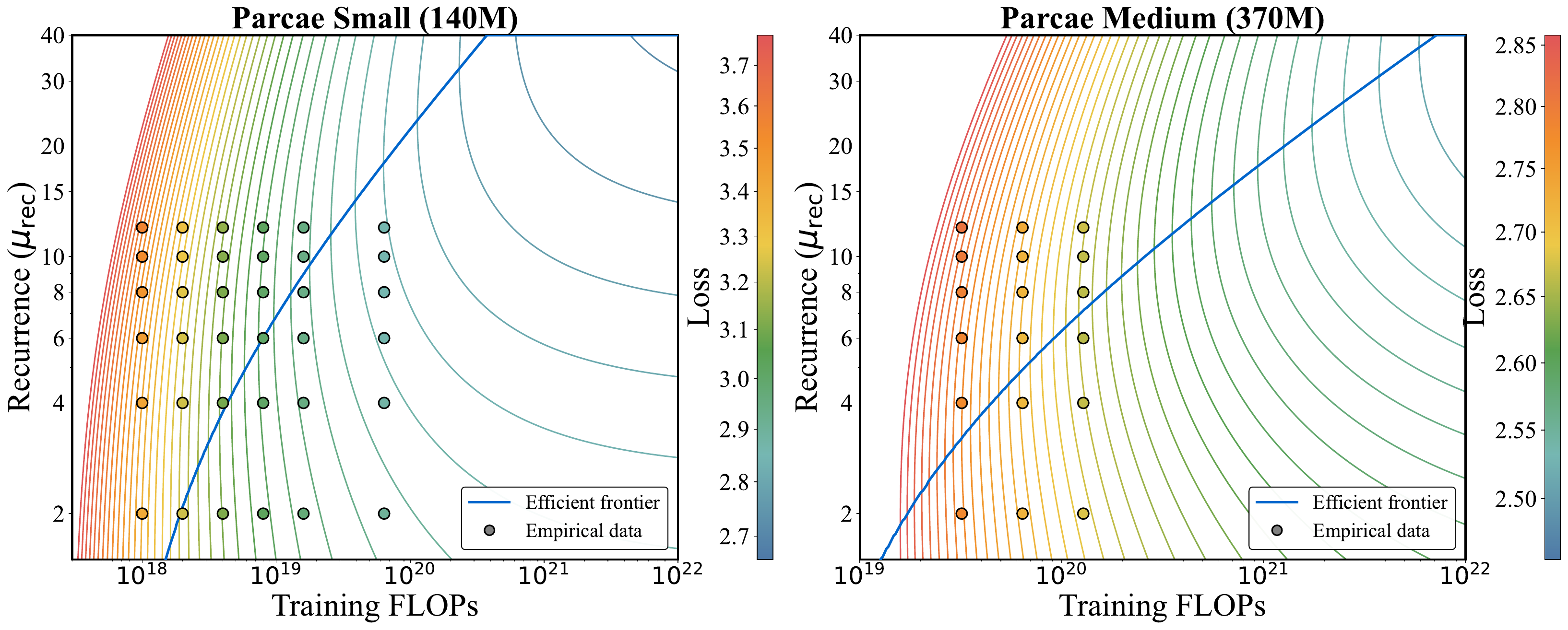}
    \caption{\textbf{Parametric Fit of Looping.} Visualization of our parametric function $\widehat{\mathcal{L}}_{\text{train}}(\mu_{\text{rec}}, D)$, which displays the IsoLoss contours for both 140M Parcae (\emph{left}) and 370M Parcae (\emph{right}) models. }
    \label{fig:parameteric-fit-app}
\end{figure}

We follow \citet{hoffmann2022trainingcomputeoptimallargelanguage} setup for fitting a parametric loss function. Specifically, using the models trained with several IsoFLOP budgets in \cref{sec:train-scaling}, we fit a parametric function of the form
\begin{equation}
    \widehat{\mathcal{L}}_{\text{train}}(\mu_{\text{rec}}, \mathcal{D}) = E + A \cdot \mathbf{N}(\mu_{\text{rec}})^{-a} + B \cdot \mathcal{D}^{-b}
\end{equation}
where $\mathbf{N}(\mu_{\text{rec}})$ is the \emph{effective parameter count} of the model if you were to unroll all loops into real parameters, $\mathcal{D}$ is the number of tokens that were used in training, and $A, B, a, b$ are learned parameters. We specifically use Huber loss \citep{huber} on the log loss between the prediction of the parametric fit and the validation loss of the models, using L-BFGS \citep{lbfgs} to minimize. We choose the parametric function of this form as it exactly follows \cite{hoffmann2022trainingcomputeoptimallargelanguage}, but with parameters $\mathbf{N}$ now being a function of \meanrecurrence. Finally, we take the best result from 500 random restarts of L-BFGS, each with up to 10,000 iterations, selecting the initialization that achieves the lowest Huber loss. The results of fitting the parametric function can be visualized in \cref{fig:parameteric-fit-app}, and the learned values can be observed in \cref{tab:parametric-fit-app}.

\begin{table}[!h]
    \centering
    \begin{tabular*}{\textwidth}{@{\extracolsep{\fill}} l|ccccccc @{}}
        \toprule
        \textbf{Model} & $\boldsymbol{E}$ & $\boldsymbol{A}$ & $\boldsymbol{a}$ & $\boldsymbol{B}$ & $\boldsymbol{b}$ & \textbf{Huber} ($\times 10^{-4}$) \\
        \midrule
        Small (140M)  & 2.662 & 522733.307 & 0.771 & 25420.102 & 0.525 & 0.44 \\
        Medium (370M) & 2.439 & 832134.346 & 0.775 & 6386.865 & 0.448 & 0.01 \\
        \bottomrule
    \end{tabular*}
    \caption{\textbf{Optimal Scaling Coefficients for Parametric Fits.}}
    \label{tab:parametric-fit-app}
\end{table}

%% file: appendix/test-parametric.tex
\section{Fitting Parametric Functions to Test-Time Looping}
\label{sec:test-par}

In this section, we provide a more detailed analysis of the test-time scaling laws discussed in \cref{sec:inf-scaling}. Following the setup discussed in \cref{sec:scaling-laws-setup}, we train several Parcae models on varying \meanrecurrence, fixing data and parameter count, and evaluate each at test-time recurrences up to $T=24$.

\paragraph{Choice of Functional Form.}
We aim to find a parametric function that captures the saturating relationship between test-time recurrence $T$ and validation loss. We consider four candidate functional forms, each with an irreducible loss floor $\mathcal{L}_\infty$ (except the pure power law):
\begin{enumerate}[label=(\alph*), nosep]
  \item $\mathcal{L}(T) = \mathcal{L}_\infty + Z \cdot e^{-zT}$ \hfill (exponential decay)
  \item $\mathcal{L}(T) = \mathcal{L}_\infty + Z \cdot (1+T)^{-z}$ \hfill (shifted power law)
  \item $\mathcal{L}(T) = \mathcal{L}_\infty + Z \cdot T^{-z}$ \hfill (power law)
  \item $\mathcal{L}(T) = Z \cdot T^{-z}$ \hfill (power law, no floor)
\end{enumerate}
Each form has 3 free parameters ($\mathcal{L}_\infty, Z, z$), except (d), which has 2. We fit each form independently to every test-time curve using least-squares on log-loss, and report the average Huber loss ($\delta = 10^{-3}$) across all curves. To evaluate extrapolation, we additionally fit each form on $T \leq \mu_{\text{rec}}$ and evaluate on held-out $T > \mu_{\text{rec}}$.

\begin{table}[h]
\centering
\begin{tabular}{l cccc}
\toprule
& $\mathcal{L}_\infty {+} Z e^{-zT}$ & $\mathcal{L}_\infty {+} Z(1{+}T)^{-z}$ & $\mathcal{L}_\infty {+} Z T^{-z}$ & $Z T^{-z}$ \\
\midrule
\multicolumn{5}{l}{\textit{In-Distribution}} \\
\quad 140M & \textbf{2.52} & 5.42 & 11.11 & 112.89 \\
\quad 370M & \textbf{1.88} & 5.26 & 10.77 & 104.95 \\
\midrule
\multicolumn{5}{l}{\textit{Extrapolation ($T > \mu_{\text{rec}}$)}} \\
\quad 140M & \textbf{3.18} & 21.41 & 43.99 & 397.90 \\
\quad 370M & \textbf{2.29} & 18.51 & 38.68 & 369.83 \\
\bottomrule
\end{tabular}
\caption{Functional form comparison for test-time scaling. We report average Huber loss ($\times 10^{-7}$) across all per-curve fits, both in-distribution (all $T$) and in extrapolation (fit $T \leq \mu_{\text{rec}}$, evaluate $T > \mu_{\text{rec}}$). Lower is better.}
\label{tab:functional-form-ablation}
\end{table}

As shown in \cref{tab:functional-form-ablation}, the exponential decay form achieves the lowest Huber loss both in-distribution ($2.3\times$ better than the shifted power law) and under extrapolation ($7.1\times$ better), consistently across both model sizes. Notably, omitting the irreducible floor $\mathcal{L}_\infty$ (form (d)) increases error by over $40\times$, confirming that test-time scaling saturates to a finite loss determined by training (this is also obvious from looking at \cref{fig:test-time-scaling-laws}). 

While purely speculative, there is a nice connection between the exponential form and Parcae's dynamical systems framework. In classical control theory literature, a stable discrete-time linear system with a spectral radius below unity converges exponentially in the state norm. The observed exponential decay in loss is thus consistent with the dynamical system formulation that Parcae uses.

\paragraph{Recovery of the training law at $T = \mu_\text{rec}$.}
We additionally observe that the fitted irreducible loss $\mathcal{L}_\infty$ closely matches the empirical loss at $T = \mu_\text{rec}$ (\cref{tab:recovery}), motivating the use of the training scaling law $\hat{\mathcal{L}}_{\mathrm{train}}(\mu_\text{rec}, D)$ as the irreducible floor in a unified law.

\begin{table}[h]
\centering
\begin{tabular}{l cc}
\toprule
Model & Mean \% Err & Max \% Err \\
\midrule
140M & 0.16\% & 0.59\% \\
370M & 0.05\% & 0.22\% \\
\bottomrule
\end{tabular}
\caption{Mean and max absolute percent error between $\mathcal{L}_\infty$ and $\mathcal{L}(T{=}\mu_\text{rec})$ across all isoFLOP configurations.}
\label{tab:recovery}
\end{table}

\paragraph{Conditioning on Training Recurrence.}
To model test-time scaling across models trained at different \meanrecurrence, the decay rate must depend on the training depth. We compare three forms for the unified test-time law, all using the training scaling law $\hat{\mathcal{L}}_{\mathrm{train}}(\mu_{\text{rec}}, D)$ from \cref{sec:train-scaling} as the irreducible floor:
\begin{enumerate}[label=(\alph*), nosep]
  \item $\hat{\mathcal{L}}_{\mathrm{train}} + Z \cdot \exp\!\bigl(-z \cdot \mu_\text{rec}^{-\gamma} \cdot T\bigr)$ \hfill (learned $\gamma$, 3 params)
  \item $\hat{\mathcal{L}}_{\mathrm{train}} + Z \cdot \exp\!\bigl(-z / \mu_\text{rec}\cdot T\bigr)$ \hfill ($\gamma = 1$, 2 params)
  \item $\hat{\mathcal{L}}_{\mathrm{train}} + Z \cdot \exp\!\bigl(-z \cdot T\bigr)$ \hfill (no conditioning, 2 params)
\end{enumerate}
\begin{table}[h]
\centering
\begin{tabular}{l ccc}
\toprule
& $Z e^{-z \mu^{-\gamma} T}$ & $Z e^{-z T / \mu}$ ($\gamma{=}1$) & $Z e^{-z T}$ (no $\mu$) \\
\midrule
\multicolumn{4}{l}{\textit{Train (isoFLOP)}} \\
\quad 140M & 0.001116 & 0.001177 & 0.003253 \\
\quad 370M & 0.000229 & 0.000283 & 0.001438 \\
\midrule
\multicolumn{4}{l}{\textit{Test (held-out, $\mu_\text{rec}{=}8$)}} \\
\quad 140M & \textbf{0.000207} & 0.000212 & 0.000266 \\
\quad 370M & 0.000133 & \textbf{0.000131} & 0.000189 \\
\bottomrule
\end{tabular}
\caption{Ablation of \meanrecurrence conditioning in the unified test-time law. We report total Huber loss on the isoFLOP training set and on held-out Table~5 models ($\mu_\text{rec} = 8$, fixed data budget). Lower is better.}
\label{tab:mu-conditioning-ablation}
\end{table}

As shown in \cref{tab:mu-conditioning-ablation}, removing $\mu_\text{rec}$ conditioning entirely increases training error by $3.5\times$ and held-out error by ${\sim}33\%$, confirming that the decay rate must depend on training depth (also obvious from looking at \cref{fig:test-time-scaling-laws}). The learned $\gamma$ offers a modest improvement (${\sim}8\%$) over $\gamma = 1$ on the training set, with fitted values of $\gamma = 1.19$ (140M) and $\gamma = 1.17$ (370M) consistent across scales; on held-out models, the two are indistinguishable. We therefore adopt $\gamma = 1$ for simplicity, yielding the unified law:
\begin{equation}
  \hat{\mathcal{L}}_{\mathrm{unified}}(T \mid \mu_\text{rec}, D) = \underbrace{E + X \cdot N(\mu_\text{rec})^{-x} + Y \cdot D^{-y}}_{\text{Training Law Floor } \hat{\mathcal{L}}_{\mathrm{train}}(\mu_\text{rec}, D)} + \underbrace{Z \cdot \exp\!\left(-\frac{z \cdot T}{\mu_\text{rec}}\right)}_{\text{Test-Time Decay}}
\end{equation}
where the test-time term depends on the ratio $T / \mu_\text{rec}$, i.e., the fraction of training depth used at inference.

\begin{figure}[!h]
  \centering
  \includegraphics[width=\linewidth]{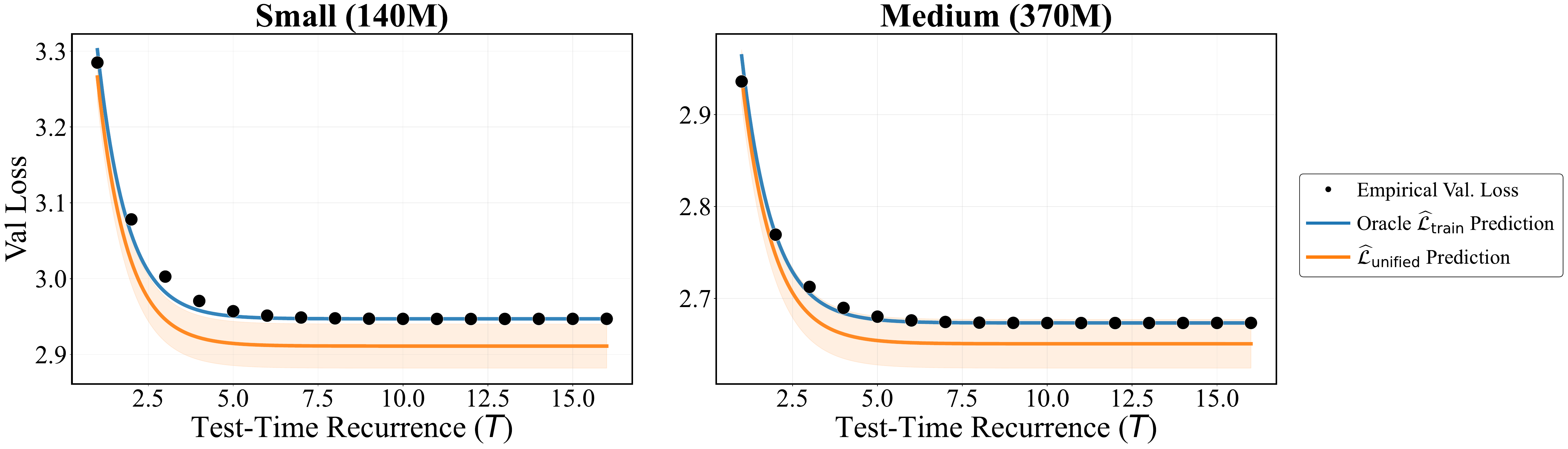}
  \vspace{-1em}
  \caption{\textbf{Out-of-Distribution Prediction of Unified Parametric Fit.}
  We visualize the prediction of our unified parametric fit (orange) and an oracle fit using the empirical loss at $T = \mu_\text{rec}$ for $\widehat{\mathcal{L}}_\text{train}$ (blue) against empirical validation loss with increasing $T$ for models trained in \cref{sec:e2e}.}
  \label{fig:unified-pred}
\end{figure}

\paragraph{Testing the Unified Parametric Fit.}
To evaluate generalization, we use the unified law fitted on isoFLOP data to predict the test-time scaling curves of held-out 140M and 370M Parcae models from \cref{sec:e2e}, which were trained on fixed data budgets and are a completely out-of-distribution setting. As shown in \cref{fig:unified-pred}, the unified fit (orange) predicts validation loss within 0.85--1.31\% average error. When the training law floor is replaced with the empirical loss at $T = \mu_\text{rec}$ (oracle, blue), error drops to 0.10--0.17\%, confirming that the test-time decay is faithfully captured and the residual error is attributable to the training law's ${\sim}1\%$ extrapolation gap.

%% file: appendix/eval-setup.tex
\newpage
\section{Extended Evaluation Details and Setup}
\label{sec:evaluation-setup}

\begin{table}[!h]
\centering
\setlength{\tabcolsep}{4pt}
\renewcommand{\arraystretch}{0.92}
\small
\begin{tabular}{@{}l l c c c@{}}
\toprule
\textbf{Category} & \textbf{Task} & \textbf{Type} & \textbf{Shots} & \textbf{Core} \\
\midrule
\multirow{8}{*}{\rotatebox[origin=c]{90}{\textit{Understanding}}}
  & HellaSwag \cite{zellers2019hellaswag} (0-shot)              & MC & 0  & \cmark \\
  & HellaSwag \cite{zellers2019hellaswag} (10-shot)             & MC & 10 & \cmark \\
  & Lambada \cite{paperno2016lambada}                & LM & 0  & \cmark \\
  & Winograd WSC \cite{wsc:2015}                   & S  & 0  & \cmark \\
  & WinoGrande \cite{sakaguchi2021winogrande}                      & S  & 0  & \cmark \\
  & BIG-Bench Language ID \cite{srivastava2023imitationgamequantifyingextrapolating}           & MC & 10 & \cmark \\
  & BIG-Bench Conlang Translation \cite{srivastava2023imitationgamequantifyingextrapolating}  & LM & 0  &        \\
  & BIG-Bench Conceptual Comb. \cite{srivastava2023imitationgamequantifyingextrapolating}     & MC & 10 &        \\
\midrule
\multirow{7}{*}{\rotatebox[origin=c]{90}{\textit{World Knowl.}}}
  & Jeopardy \cite{kaggle200000Jeopardy}                        & LM & 10 & \cmark \\
  & BIG-Bench QA WikiData \cite{srivastava2023imitationgamequantifyingextrapolating}           & LM & 10 & \cmark \\
  & ARC-Easy \cite{clark2018think}                        & MC & 10 & \cmark \\
  & ARC-Challenge \cite{clark2018think}                   & MC & 10 & \cmark \\
  & MMLU (0-shot) \cite{hendrycks2021measuringmassivemultitasklanguage}                  & MC & 0  &        \\
  & MMLU (5-shot) \cite{hendrycks2021measuringmassivemultitasklanguage}                   & MC & 5  &        \\
  & BIG-Bench Misconceptions \cite{srivastava2023imitationgamequantifyingextrapolating}        & MC & 10 &        \\
\midrule
\multirow{8}{*}{\rotatebox[origin=c]{90}{\textit{Commonsense}}}
  & COPA \cite{gordon-etal-2012-semeval}                           & MC & 0  & \cmark \\
  & CommonsenseQA \cite{talmor2019commonsenseqaquestionansweringchallenge}                   & MC & 10 & \cmark \\
  & PIQA \cite{bisk2020piqa}                            & MC & 10 & \cmark \\
  & OpenBookQA \cite{mihaylov2018suitarmorconductelectricity}                      & MC & 0  & \cmark \\
  & SIQA \cite{sap2019socialiqacommonsensereasoningsocial}                            & MC & 10 &        \\
  & BIG-Bench Novel Concepts \cite{srivastava2023imitationgamequantifyingextrapolating}        & MC & 10 &        \\
  & BIG-Bench Strange Stories \cite{srivastava2023imitationgamequantifyingextrapolating}       & MC & 10 &        \\
  & BIG-Bench Strategy QA \cite{srivastava2023imitationgamequantifyingextrapolating}           & MC & 10 &        \\
\midrule
\multirow{11}{*}{\rotatebox[origin=c]{90}{\textit{Symbolic / Math}}}
  & BIG-Bench Dyck Languages \cite{srivastava2023imitationgamequantifyingextrapolating}        & LM & 10 & \cmark \\
  & AGI Eval LSAT AR \cite{zhong2021arlsat}                & MC & 3  & \cmark \\
  & BIG-Bench CS Algorithms \cite{srivastava2023imitationgamequantifyingextrapolating}         & LM & 10 & \cmark \\
  & BIG-Bench Operators \cite{srivastava2023imitationgamequantifyingextrapolating}             & LM & 10 & \cmark \\
  & BIG-Bench Repeat Copy Logic \cite{srivastava2023imitationgamequantifyingextrapolating}     & LM & 10 & \cmark \\
  & BIG-Bench Elementary Math QA \cite{srivastava2023imitationgamequantifyingextrapolating}    & MC & 10 &        \\
  & BIG-Bench Logical Deduction \cite{srivastava2023imitationgamequantifyingextrapolating}     & MC & 10 &        \\
  & Simple Arithmetic (no spaces) \cite{llmfoundry}  & LM & 10 &        \\
  & Simple Arithmetic (w/ spaces) \cite{llmfoundry}  & LM & 10 &        \\
  & MathQA \cite{amini2019mathqa}                          & MC & 10 &        \\
  & LogiQA \cite{liu2020logiqa}                          & MC & 10 &        \\
\midrule
\multirow{8}{*}{\rotatebox[origin=c]{90}{\textit{Reading Comp.}}}
  & SQuAD \cite{rajpurkar2016squad100000questionsmachine}                           & LM & 10 & \cmark \\
  & CoQA \cite{reddy2019coqaconversationalquestionanswering}                            & LM & 0  & \cmark \\
  & BoolQ \cite{clark2019boolq}                           & MC & 10 & \cmark \\
  & PubMedQA (labeled) \cite{jin2019pubmedqa}              & LM & 10 &        \\
  & AGI Eval LSAT RC \cite{zhong2023agieval}                & MC & 3  &        \\
  & AGI Eval LSAT LR \cite{wang2022lsat}                & MC & 3  &        \\
  & AGI Eval SAT English \cite{zhong2023agieval}            & MC & 3  &        \\
  & BIG-Bench Understanding Fables \cite{srivastava2023imitationgamequantifyingextrapolating}  & MC & 10 &        \\
\midrule
\multirow{4}{*}{\rotatebox[origin=c]{90}{\textit{Safety}}}
  & Winogender MC (Female) \cite{rudinger2018genderbiascoreferenceresolution}         & MC & 10 &        \\
  & Winogender MC (Male) \cite{rudinger2018genderbiascoreferenceresolution}            & MC & 10 &        \\
  & Enterprise PII Classification   & MC & 10 &        \\
  & BBQ \cite{parrish2022bbqhandbuiltbiasbenchmark}                             & MC & 3  &        \\
\bottomrule
\end{tabular}
\caption{\textbf{Full list of downstream evaluation} Tasks marked with \cmark{} are included in the \textbf{Core} \cite{li2025datacomplmsearchgenerationtraining}; all tasks are included in \textbf{Core-Extended} \cite{li2025datacomplmsearchgenerationtraining}. Type indicates the scoring method: \textbf{MC} (multiple choice, lowest mean NLL), \textbf{S} (schema-based NLL), or \textbf{LM} (exact greedy match).}
\label{tab:eval-tasks}
\end{table}

We include a complete list of benchmarks used for evaluation in \cref{tab:eval-tasks}. For our results in \cref{sec:results} where we are comparing against baseline transformers, we run each benchmark with three different seeds, as this changes both the initial recurrent state and the in-context few-shot examples. 

%% file: appendix/fixed-expanded-results.tex
\section{Expanded Results For Fixed-Depth and Looping IsoFLOP Comparison}
\label{sec:fixed-comparison-expanded}

We included an expanded form of \cref{tab:qual-fix-loop} to ensure reproducibility, which additionally includes error bars in \cref{tab:qual-fix-loop-expanded}.

\begin{table}[!h]
\centering
\begin{tabular}{@{}c c c cc cc@{}}
\toprule
& \textbf{FLOPs} & & \multicolumn{2}{c}{\textbf{Optimal $\mu_{\mathrm{rec}}^*$}} & \multicolumn{2}{c}{\textbf{Fixed-Depth} ($\mu_{\mathrm{rec}}=1$)} \\
\cmidrule(lr){4-5} \cmidrule(lr){6-7}
& ($\times 10^{18}$) & $\mu_{\mathrm{rec}}^*$ & Core & Core Ext. & Core & Core Ext. \\
\midrule
\multirow{6}{*}{\rotatebox[origin=c]{90}{\textbf{140M}}}
& $1$   & 2  & $7.6 \pm 0.3$  & $5.7 \pm 0.5$  & $\mathbf{7.9 \pm 0.2}$  & $\mathbf{6.1 \pm 0.1}$ \\
& $2$   & 2  & $9.0 \pm 0.2$  & $6.2 \pm 0.1$  & $\mathbf{10.5 \pm 0.1}$ & $\mathbf{6.4 \pm 0.2}$ \\
& $4$   & 4  & $\mathbf{11.2 \pm 0.0}$ & $\mathbf{8.4 \pm 0.2}$ & $10.7 \pm 0.1$ & $8.1 \pm 0.3$ \\
& $8$   & 6  & $10.5 \pm 0.1$ & $\mathbf{7.8 \pm 0.2}$ & $\mathbf{11.8 \pm 0.2}$ & $7.7 \pm 0.2$ \\
& $16$  & 8  & $\mathbf{14.6 \pm 0.1}$ & $\mathbf{9.8 \pm 0.4}$ & $13.0 \pm 0.2$ & $8.8 \pm 0.4$ \\
& $64$  & 10 & $\mathbf{16.2 \pm 0.2}$ & $\mathbf{11.0 \pm 0.1}$ & $15.0 \pm 0.2$ & $9.5 \pm 0.4$ \\
\midrule
\multirow{3}{*}{\rotatebox[origin=c]{90}{\textbf{370M}}}
& $32$  & 4  & $15.2 \pm 0.1$ & $10.1 \pm 0.2$ & $\mathbf{16.8 \pm 0.1}$ & $\mathbf{11.2 \pm 0.4}$ \\
& $64$  & 6  & $\mathbf{18.1 \pm 0.2}$ & $11.6 \pm 0.2$ & $\mathbf{18.1 \pm 0.1}$ & $\mathbf{12.1 \pm 0.2}$ \\
& $128$ & 6  & $\mathbf{20.1 \pm 0.1}$ & $\mathbf{13.0 \pm 0.1}$ & $18.1 \pm 0.1$ & $12.0 \pm 0.1$ \\
\bottomrule
\end{tabular}
\caption{\textbf{Expanded Core Scores Comparison of Looping Optimal Frontier over Purely Scaling Data.} Including variance bars now.}
\label{tab:qual-fix-loop-expanded}
\end{table}

%% file: appendix/scaling-laws-setup.tex
\section{Expanded Setup For Training and Test-Time Scaling Laws}
\label{sec:scaling-laws-setup}

For our scaling laws experiments, we train models under two setups: (1) an isoFLOP training setup where we train models with variable amounts of \meanrecurrence, but with fixed FLOP and parameter budgets, and (2) where we vary \meanrecurrence, but keep data and parameters constant. Additionally, for our unified scaling laws experiments, we reuse the models trained in setup (1) and then evaluate them with varying amounts of test-time recurrences. All of the experiments use the same exact experimental setup for Transformers described in \cref{sec:hyperparameters} and \cref{sec:model-definitions} (i.e., using a \texttt{nanochat} \citep{nanochat}). We will discuss each experiment in detail below.

\paragraph{(1) Setup for IsoFLOP Experiments.} For each parameter count (140M and 370M), we fix the total training FLOP budget and vary $\mu_{\text{rec}} \in \{2, 4, 6, 8, 10, 12\}$, adjusting the number of training tokens to maintain the FLOP budget (i.e., increasing $\mu_{\text{rec}}$ reduces the token budget proportionally). For 140M models, we use FLOP budgets of $\{1, 2, 4, 8, 16, 64\} \times 10^{18}$; for 370M models, $\{32, 64, 128\} \times 10^{18}$.
This yields 36 and 18 trained models for 140M and 370M, respectively. Each model is evaluated on a held-out validation set at $T = \mu_{\text{rec}}$. 
We use these validation losses to fit the parametric training scaling law $\widehat{\mathcal{L}}_{\text{train}}(\mu_{\text{rec}}, \mathcal{D})$ and to extract optimal $\mu_{\text{rec}}^*$ at each FLOP budget via parabolic fits. 
Additionally, we train fixed-depth ($\mu_{\text{rec}} = 1$) Parcae models at each FLOP budget to serve as baselines for the looping frontier comparison. Expanded details of the predicted frontiers calculation can be found in \cref{sec:fixed-comparison-expanded}.

\paragraph{(2) Setup for Test-Time Saturation and Power Laws.} To study how test-time recurrence scales quality, we train 140M and 370M Parcae models under a fixed data budget of 11.2B tokens with $\mu_{\text{rec}} \in \{2, 4, 6, 8, 10, 12\}$. Each model is then evaluated on a held-out validation set at test-time recurrences $T \in \{1, 2, 3, \ldots, 24\}$, yielding a saturation curve per $\mu_{\text{rec}}$. We fit an independent exponential decay law $\mathcal{L}(T) = \mathcal{L}_\infty + Z \cdot \exp(-z \cdot T)$ to each curve following the procedure in Section~\ref{sec:test-par}. We additionally evaluate the Parcae models from \cref{sec:e2e} (140M--1.3B, trained at $\mu{\text{rec}}=8$) at test-time recurrences $T \in \{1, \ldots, 16\}$ to verify that the saturation behavior is consistent across model sizes.

\paragraph{(3) Setup for Unified Scaling Law.} To fit the unified scaling law (\cref{eq:unified}), we reuse the isoFLOP models from setup (1) and evaluate each at test-time recurrences $T \in \{1, 2, 4, 6, 8, 10, 12, 16, 20, 24\}$, yielding approximately 540 data points per model size. We fit all 8 parameters of \cref{eq:unified} jointly on this data using Huber loss on the log loss with L-BFGS over 1,000 random restarts. To validate, we evaluate the unified fit on held-out 140M and 370M Parcae models from \cref{sec:e2e}, which were trained on fixed data budgets outside the isoFLOP sweep, at test-time recurrences $T \in \{1, \ldots, 16\}$.

%% file: appendix/definitions.tex
\section{Model Definitions}
\label{sec:model-definitions}

As we perform experiments in two setups, one following prior work in recurrent depth models \citep{geiping_scaling_2025} and one following a strong baseline transformer \citep{nanochat}, we separate the model definitions into \cref{sec:rdm-parcae-model-def} and \cref{sec:trans-parcae-model-def}, respectively.

\subsection{Model Definitions for RDM and Parcae Comparison}
\label{sec:rdm-parcae-model-def}

In this section, we will discuss the model configuration used for models in \cref{sec:e2e} for RDMs \citep{geiping_scaling_2025}. For all \prelude, \recurrent, and \coda modules, we follow \citet{geiping_scaling_2025}, and use standard, causal self-attention and gated SwiGLU MLP \citep{shazeer2020gluvariantsimprovetransformer}. For attention, we use RoPE \citep{su2023roformerenhancedtransformerrotary} with $\theta=50000$ and for normalization we use RMSNorm \citep{zhang2019rootmeansquarelayer}. We use Pre-Norm transformer blocks for all modules within Parcae, and follow \citet{takase2025spikemorestabilizingpretraining}, initializing weights using $\mathcal{N}(0,\frac{2}{5d})$, where $d$ is the model dimension.

\begin{table}[h]
    \centering
    \begin{tabular}{ccccc}
    \toprule
        & Parcae-100M & Parcae-350M & RDM-100M & RDM-350M \\
    \midrule
        \textbf{Parameters} 
            & 114,242,560 & 378,558,464 & 114,242,560  & 382,765,056 \\ 
        Layers in \prelude  
            & 1 & 1 & 1 & 1 \\ 
        Layers in \coda 
            & 1 & 1 & 1 & 1 \\ 
        Layers in \recurrent 
            & 1 & 2 & 1 & 2 \\
        $d_{\text{model}}$ 
            & 1{,}024 & 2{,}048 & 1{,}024 & 2{,}048 \\
        $d_{\text{intermediate}}$ 
            & 3{,}520 & 7{,}040 & 3{,}520 & 7{,}040 \\
    \midrule
        Attention 
            & \multicolumn{4}{c}{Causal Self-Attention \citep{vaswani2023attentionneed}} \\
        MLP 
            & \multicolumn{4}{c}{SwiGLU \citep{elfwing2017sigmoidweightedlinearunitsneural,shazeer2020gluvariantsimprovetransformer}} \\
        Pos. Embed. 
            & \multicolumn{4}{c}{RoPE \citep{su2023roformerenhancedtransformerrotary}} \\
        Vocab Size 
            & \multicolumn{4}{c}{65{,}536} \\
        Norm 
            & \multicolumn{4}{c}{RMS-Norm \citep{zhang2019rootmeansquarelayer}} \\
        Init 
            & \multicolumn{4}{c}{Scaled \citep{takase2025spikemorestabilizingpretraining}} \\
        Tied Embeddings 
            & \multicolumn{4}{c}{Yes} \\
      State Init.
          & \multicolumn{4}{c}{\texttt{like-init} \citep{geiping_scaling_2025}} \\
      \meanrecurrence
          & 16 & 8 & 16 & 8 \\
      Backprop Depth
          &  8 & 4 & 8 & 4\\
      Sampling
          & \multicolumn{4}{c}{Poisson Distribution} \\
    \bottomrule
    \end{tabular}
    \caption{Model definitions of both Parcae and baseline residual-norm RDMs \citep{geiping_scaling_2025}.}
    \label{tab:parcae-definitions}
\end{table}

\subsection{Model Definitions for Transformer and Parcae Comparison}
\label{sec:trans-parcae-model-def}

In this section, we will discuss the model definitions used for our experiments in \cref{sec:e2e} for Transformers. Our architecture is derived from \citet{nanochat}, while being slightly adapted to fit with GPT2 \citep{radford2019language} style parameter classes. Model definitions of both Parcae and baseline Transformers can be found in \cref{tab:transformer-parcae-definitions}, while the difference in parameter count can be found in \cref{tab:transform-parcae-parameter-count}.\footnote{We note that Parcae does technically introduce additional parameters over baseline Transformers; however, they are negligible in comparison to total parameter counts.}

\begin{table}[!h]
  \centering
  \small
  \begin{tabular}{lccccc}
  \toprule
      & & Small (140M) & Medium (370M) & Large (770M) & XLarge (1.3B) \\
  \midrule
      \multirow{9}{*}{\rotatebox[origin=c]{90}{\textit{Architecture}}}
      & Layers (Transformer)
          & 6 & 12 & 18 & 24 \\
      & Layers in \prelude (Parcae)
          & 2 & 4 & 6 & 8 \\
      & Layers in \recurrent (Parcae)
          & 2 & 4 & 6 & 8 \\
      & Layers in \coda (Parcae)
          & 2 & 4 & 6 & 8 \\
      & $d_{\text{model}}$
          & 768 & 1{,}024 & 1{,}280 & 1{,}536 \\
      & $d_{\text{intermediate}}$
          & 3{,}072 & 4{,}096 & 5{,}120 & 6{,}144 \\
      & Attention Heads
          & 6 & 8 & 10 & 12 \\
      & Head Dimension
          & 128 & 128 & 128 & 128 \\
  \midrule
      \multirow{10}{*}{\rotatebox[origin=c]{90}{\textit{Shared Details}}}
      & Attention
          & \multicolumn{4}{c}{Causal Self-Attention \citep{vaswani2023attentionneed} w/ QK-Norm \citep{henry2020querykeynormalizationtransformers} } \\
      & MLP
          & \multicolumn{4}{c}{$\text{ReLU}^2$ \citep{zhang2024relu2winsdiscoveringefficient}} \\
      & Value Embeddings
          & \multicolumn{4}{c}{Gated, alternating layers \citep{tian2023resformerscalingvitsmultiresolution}} \\
      & Pos.\ Embed.
          & \multicolumn{4}{c}{RoPE ($\theta{=}50{,}000$) \citep{su2023roformerenhancedtransformerrotary}} \\
      & Vocab Size
          & \multicolumn{4}{c}{32{,}768} \\
      & Norm
          & \multicolumn{4}{c}{RMS-Norm (Pre-Norm) \citep{zhang2019rootmeansquarelayer}} \\
      & Context Length
          & \multicolumn{4}{c}{2{,}048} \\
      & Bias
          & \multicolumn{4}{c}{None} \\
      & Init.
          & \multicolumn{4}{c}{Scaled-zero \citep{takase2025spikemorestabilizingpretraining, nanochat}} \\
      & Tied Embeddings
          & \multicolumn{4}{c}{Yes} \\
  \midrule
      \multirow{5}{*}{\rotatebox[origin=c]{90}{\textit{Parcae}}}
      & Injection
          & \multicolumn{4}{c}{Diagonal} \\
      & State Init.
          & \multicolumn{4}{c}{\texttt{like-init} \citep{geiping_scaling_2025}} \\
      & \meanrecurrence
          & \multicolumn{4}{c}{8} \\
      & Backprop Depth
          & \multicolumn{4}{c}{4} \\
      & Sampling
          & \multicolumn{4}{c}{Poisson (truncated, per-sequence)} \\
  \bottomrule
  \end{tabular}
  \caption{Model definitions of both Parcae and baseline Transformers.}
  \label{tab:transformer-parcae-definitions}
\end{table}

\begin{table}[!h]
  \centering
  \small
  \begin{tabular}{lcccc}
  \toprule
      & Small (140M) & Medium (370M) & Large (770M) & XLarge (1.3B) \\
  \midrule
      Transformer Parameters & 143,141,184 & 385,903,104 & 773,375,040 & 1,333,868,544 \\
      Parcae Parameters      & 144,323,136 & 388,003,328 & 776,655,680 & 1,338,591,744 \\
      \midrule
      Additional Parameters  & 1{,}181{,}952 & 2{,}100{,}224 & 3{,}280{,}640 & 4{,}723{,}200 \\
      Additional (\%)        & 0.83\% & 0.54\% & 0.42\% & 0.35\% \\
  \bottomrule
  \end{tabular}
  \caption{Comparison of Parcae and Transformer parameter count.}
  \label{tab:transform-parcae-parameter-count}
\end{table}

%% file: appendix/hyperparameters.tex
\newpage

\section{Hyperparameters and Training Details}
\label{sec:hyperparameters}

Again, as we perform experiments in two setups, one following prior work in recurrent depth models \citep{geiping_scaling_2025} and one following a strong baseline transformer \citep{nanochat}, we separate the hyperparameter configurations into \cref{sec:rdm-parcae-hyp} and \cref{sec:trans-parcae-hyp}, respectively.

\subsection{Hyperparameters for Parcae and RDM Comparison}
\label{sec:rdm-parcae-hyp}
In this section, we will discuss the hyperparameter configuration used in \cref{sec:e2e} for RDMs \citep{geiping_scaling_2025}. We train with a warm-up and cool-down (4096 steps following \citep{geiping_scaling_2025}) and a constant learning rate ($\eta = 4 \times 10^{-3}$ for 100M models and $\eta = 2 \times 10^{-3}$ for 350M models) \citep{pmlr-v202-geiping23a, Zhai_2022_CVPR}. As our optimizer, we use Adam with decoupled weight regularization ($\beta_1 = 0.9, \beta_2 0.95$) \citep{kingma2017adammethodstochasticoptimization, loshchilov2019decoupledweightdecayregularization}, using update clipping \citep{wortsman2023stable} and removing the $\epsilon$ constant \citep{everett2024scalingexponentsparameterizationsoptimizers}. Gradients above 1 are clipped. 

For learning rates, we swept our selection of learning rates for RDMs \citep{geiping_scaling_2025}, over the search space $[2e-4, 4e-4, 6e-4, 8e-4, 1e-3]$, approximately using 10 to 1 token to parameter ratio. We then select the best learning rate for each scale (e.g., 4e-4 for 100M and 2e-4 for 350M). We perform no learning rate sweep for Parcae, using the best learning rate for RDMs \citep{geiping_scaling_2025}. We do this so that our comparison between Parcae and prior methods is fair, as we observed significant divergence in training for RDMs based on learning rate (see \cref{sec:stability-ablations}).
We stipulate that Parcae models would likely perform better with stronger hyperparameter tuning.

\subsection{Hyperparameters for Parcae and Transformer Comparison}
\label{sec:trans-parcae-hyp}

In this section, we will discuss the hyperparameter configuration used in \cref{sec:e2e} for Transformers. We use a simplified version of \texttt{nanochat} \citep{nanochat}, with the main difference being a simplified learning rate selection. Specifically, in \texttt{nanochat} \citep{nanochat}, different parameter groups have different learning rates (e.g., MLP, value-embeddings, and projection head have different learning rates), which we simplify into just two parameter groups, one for AdamW \citep{kingma2017adammethodstochasticoptimization,loshchilov2019decoupledweightdecayregularization} and one for Muon \citep{jordan2024muon}. A breakdown of which parameters are placed with each of these groups follows \texttt{nanochat} \citep{nanochat}, and can be found in \cref{tab:parameter-groups}.

\begin{table}[!h]
  \centering
  \small
  \begin{tabular}{ll}
  \toprule
      \textbf{Optimizer} & \textbf{Parameters} \\
  \midrule
      \multirow{5}{*}{AdamW \citep{kingma2017adammethodstochasticoptimization}}
      & Token embeddings (\texttt{wte}) \\
      & LM head (\texttt{lm\_head}) \\
      & Normalization layers (\texttt{RMSNorm}) \\
      & Value embedding gates (\texttt{ve\_gate}) \\
      & All 1D parameters \\
  \midrule
      \multirow{2}{*}{AdamW \citep{kingma2017adammethodstochasticoptimization} (Parcae only)}
      & Injection parameters ($\A$, $\dt$, $\B$) \\
      & Readout projection ($\C$) \\
  \midrule
      \multirow{2}{*}{Muon \citep{jordan2024muon}}
      & Attention projections ($W_Q$, $W_K$, $W_V$, $W_O$) \\
      & MLP weights ($W_{\text{fc}}$, $W_{\text{proj}}$) \\
  \bottomrule
  \end{tabular}
  \caption{Optimizer parameter group assignment for Parcae and baseline Transformers.}
  \label{tab:parameter-groups}
\end{table}

As we simplify the learning rate setup used in \texttt{nanochat} \citep{nanochat}, we perform a rigorous hyperparameter sweep of baseline Transformers to create the strongest baseline. Specifically, for small and medium models, we form a sweep over $\{3e-4, 5e-4, 6e-4, 8e-4, 1e-3, 1.5e-3, 2e-3, 3e-3, 4e-3, 8e-3, 1e-2, 1.5e-2, 2e-2 \}$ for AdamW learning rates and a sweep over $\{3e-4, 5e-4, 1e-3, 2e-3, 4e-3, 8e-3, 1e-2, 1.5e-2, 2e-2\}$ for Muon learning rates using 1:20 param to token ratios for the search, where we find that for both models $8e-3$ works best for both sizes and optimizers. For large and xlarge transformer models, we perform a constrained sweep of learning rate in $\{2e-3, 3e-3, 4e-3, 6e-3, 8e-3\}$ for AdamW \citep{kingma2017adammethodstochasticoptimization}, while keeping the Muon learning rate fixed at $8e{-3}$, using a 1:7 parameter to token ratio, where we find that a learning rate of $6e-3$ performs the best. We perform \emph{no learning rate sweeps for Parcae}, to ensure that we are giving the fairest comparison. We expect that there likely exists a more optimal learning rate for Parcae, which could further improve performance.

Following \texttt{nanochat} \citep{nanochat}, we use a fixed learning rate, with no warmup and 50\% cooldown. For Muon \citep{jordan2024muon}, we use five iterations of polar express orthogonalization \citep{amsel2025polarexpressoptimalmatrix}, factored variance reductions \citep{si2025adamuonadaptivemuonoptimizer}, and cautious weight decay \citep{chen2026cautiousweightdecay}. We train with BF16 mixed precision. For our data pipeline, we use a BOS-aligned dataloader with BestFit-Crop packing \citep{ding2024fewertruncationsimprovelanguage} and training on FineWeb-edu \citep{penedo2024finewebdatasetsdecantingweb}. We clip gradients above 1.
A table of hyperparameter details can be found in \cref{tab:transformer-parcae-hyperparameters}.

\begin{table}[h]
  \centering
  \small
  \begin{tabular}{lcccc}
  \toprule
      & Small (140M) & Medium (370M) & Large (770M) & XLarge (1.3B) \\
  \midrule
      Training Tokens
          & 11.2B & 29.6B & 61.6B & 104B \\
      Batch Size (sequences)
          & 256 & 256 & 256 & 256 \\
      Sequence Length
          & 2{,}048 & 2{,}048 & 2{,}048 & 2{,}048 \\
      Precision
          & \multicolumn{4}{c}{\texttt{bf16-mixed}} \\
  \midrule
      AdamW LR
          & $8 \times 10^{-3}$ & $8 \times 10^{-3}$ & $6 \times 10^{-3}$ & $6 \times 10^{-3}$ \\
      AdamW $(\beta_1, \beta_2)$
          & \multicolumn{4}{c}{$(0.8, 0.95)$} \\
      AdamW Weight Decay
          & \multicolumn{4}{c}{$0.0$} \\
      AdamW $\epsilon$
          & \multicolumn{4}{c}{$10^{-10}$} \\
  \midrule
      Muon LR
          & \multicolumn{4}{c}{$8 \times 10^{-3}$} \\
      Muon Momentum
          & \multicolumn{4}{c}{$0.95$} \\
      Muon Weight Decay
          & \multicolumn{4}{c}{$0.2$ (linear decay to 0)} \\
      Muon Orthogonalization Steps
          & \multicolumn{4}{c}{5} \\
  \midrule
      LR Schedule
          & \multicolumn{4}{c}{Fixed (0\% warmup, 50\% cooldown)} \\
      Gradient Clipping
          & \multicolumn{4}{c}{$1.0$} \\
  \bottomrule
  \end{tabular}
  \caption{Hyperparameter used from training Parcae and Transformer models in \cref{sec:e2e} for Transformers.}
  \label{tab:transformer-parcae-hyperparameters}
\end{table}

Lastly, following \texttt{nanochat} \citep{nanochat}, we train our own tokenizer, which we use for all models. Details of the tokenizer training and setup can be found in \cref{sec:tokenizer}.

%% file: appendix/tokenizer.tex
\section{Tokenizer Training}
\label{sec:tokenizer}

We train a custom BPE tokenizer with a vocabulary size of 32,768 using the HuggingFace \texttt{tokenizers} library. We follow a GPT-4 style configuration \citep{openai2024gpt4technicalreport}: byte-level BPE with byte fallback, no text normalization, and a GPT-4 style pre-tokenization split pattern. The tokenizer is trained on 2 billion characters from the FineWeb-Edu training set \citep{penedo2024finewebdatasetsdecantingweb}, with individual documents capped at 10,000 characters. We define three special tokens: \texttt{<|bos|>}, \texttt{<|eos|>}, and \texttt{<|pad|>}. A small comparison of our tokenizer used in our experiments with others can be found in \cref{tab:tokenizer-compression}.

\begin{table}[h]
  \centering
  \small
  \begin{tabular}{lccc}
  \toprule
      \textbf{Tokenizer} & \textbf{Vocab Size} & \multicolumn{2}{c}{\textbf{Bytes/Token} $\uparrow$} \\
      \cmidrule(lr){3-4}
      & & Train & Val \\
  \midrule
      GPT-2 (\texttt{gpt2})       & 50{,}257  & 4.67 & 4.63 \\
      GPT-4 (\texttt{cl100k})     & 100{,}277 & 4.81 & 4.76 \\
      Ours                        & 32{,}768  & 4.72 & 4.65 \\
  \bottomrule
  \end{tabular}
  \caption{Compression ratio (bytes per token) on FineWeb-Edu for tokenizer used in training.}
  \label{tab:tokenizer-compression}
\end{table}

%% file: main.bib
@inproceedings{geiping_scaling_2025,
title={Scaling up Test-Time Compute with Latent Reasoning: A Recurrent Depth Approach},
author={Jonas Geiping and Sean Michael McLeish and Neel Jain and John Kirchenbauer and Siddharth Singh and Brian R. Bartoldson and Bhavya Kailkhura and Abhinav Bhatele and Tom Goldstein},
booktitle={The Thirty-ninth Annual Conference on Neural Information Processing Systems},
year={2025},
url={https://openreview.net/forum?id=S3GhJooWIC}
}

@inproceedings{
avi_learn_algorithm,
title={Can You Learn an Algorithm?  Generalizing from Easy to Hard Problems with Recurrent Networks},
author={Avi Schwarzschild and Eitan Borgnia and Arjun Gupta and Furong Huang and Uzi Vishkin and Micah Goldblum and Tom Goldstein},
booktitle={Advances in Neural Information Processing Systems},
editor={A. Beygelzimer and Y. Dauphin and P. Liang and J. Wortman Vaughan},
year={2021},
url={https://openreview.net/forum?id=Tsp2PL7-GQ}
}

@inproceedings{
bansalEndtoendAlgorithmSynthesis2022,
title={End-to-end Algorithm Synthesis with Recurrent Networks: Extrapolation without Overthinking},
author={Arpit Bansal and Avi Schwarzschild and Eitan Borgnia and Zeyad Emam and Furong Huang and Micah Goldblum and Tom Goldstein},
booktitle={Advances in Neural Information Processing Systems},
editor={Alice H. Oh and Alekh Agarwal and Danielle Belgrave and Kyunghyun Cho},
year={2022},
url={https://openreview.net/forum?id=PPjSKy40XUB}
}

@article{
mcleish_retrofitted_recurrence,
    title={Teaching Pretrained Language Models to Think Deeper with Retrofitted Recurrence}, 
    author={Sean McLeish and Ang Li and John Kirchenbauer and Dayal Singh Kalra and Brian R. Bartoldson and Bhavya Kailkhura and Avi Schwarzschild and Jonas Geiping and Tom Goldstein and Micah Goldblum},
    journal={arXiv preprint arXiv:2511.07384},
    year={2025}
}

@inproceedings{
yangLoopedTransformersAre2023,
title={Looped Transformers are Better at Learning Learning Algorithms},
author={Liu Yang and Kangwook Lee and Robert D Nowak and Dimitris Papailiopoulos},
booktitle={The Twelfth International Conference on Learning Representations},
year={2024},
url={https://openreview.net/forum?id=HHbRxoDTxE}
}

@inproceedings{
    dehghaniUniversalTransformers2019,
    title={Universal Transformers},
    author={Mostafa Dehghani and Stephan Gouws and Oriol Vinyals and Jakob Uszkoreit and Lukasz Kaiser},
    booktitle={International Conference on Learning Representations},
    year={2019},
    url={https://openreview.net/forum?id=HyzdRiR9Y7},
}

@misc{
zhuScalingLatentReasoning2025,
      title={Scaling Latent Reasoning via Looped Language Models}, 
      author={Rui-Jie Zhu and Zixuan Wang and Kai Hua and Tianyu Zhang and Ziniu Li and Haoran Que and Boyi Wei and Zixin Wen and Fan Yin and He Xing and Lu Li and Jiajun Shi and Kaijing Ma and Shanda Li and Taylor Kergan and Andrew Smith and Xingwei Qu and Mude Hui and Bohong Wu and Qiyang Min and Hongzhi Huang and Xun Zhou and Wei Ye and Jiaheng Liu and Jian Yang and Yunfeng Shi and Chenghua Lin and Enduo Zhao and Tianle Cai and Ge Zhang and Wenhao Huang and Yoshua Bengio and Jason Eshraghian},
      year={2025},
      eprint={2510.25741},
      archivePrefix={arXiv},
      primaryClass={cs.CL},
      url={https://arxiv.org/abs/2510.25741}, 
}

@misc{
wangHierarchicalReasoningModel2025b,
      title={Hierarchical Reasoning Model}, 
      author={Guan Wang and Jin Li and Yuhao Sun and Xing Chen and Changling Liu and Yue Wu and Meng Lu and Sen Song and Yasin Abbasi Yadkori},
      year={2025},
      eprint={2506.21734},
      archivePrefix={arXiv},
      primaryClass={cs.AI},
      url={https://arxiv.org/abs/2506.21734}, 
}

@misc{
jolicoeur-martineauLessMoreRecursive2025,
      title={Less is More: Recursive Reasoning with Tiny Networks}, 
      author={Alexia Jolicoeur-Martineau},
      year={2025},
      eprint={2510.04871},
      archivePrefix={arXiv},
      primaryClass={cs.LG},
      url={https://arxiv.org/abs/2510.04871}, 
}

@misc{baeMixtureofRecursionsLearningDynamic2025,
      title={Mixture-of-Recursions: Learning Dynamic Recursive Depths for Adaptive Token-Level Computation}, 
      author={Sangmin Bae and Yujin Kim and Reza Bayat and Sungnyun Kim and Jiyoun Ha and Tal Schuster and Adam Fisch and Hrayr Harutyunyan and Ziwei Ji and Aaron Courville and Se-Young Yun},
      year={2025},
      eprint={2507.10524},
      archivePrefix={arXiv},
      primaryClass={cs.CL},
      url={https://arxiv.org/abs/2507.10524}, 
}

@inproceedings{
LoopFormerElasticDepthLooped2025,
title={LoopFormer: Elastic-Depth Looped Transformers for Latent Reasoning via Shortcut Modulation},
author={Ahmadreza Jeddi and Marco Ciccone and Babak Taati},
booktitle={The Fourteenth International Conference on Learning Representations},
year={2026},
url={https://openreview.net/forum?id=RzYXb5YWBs}
}

@misc{xuExpressivePowerLooped2025,
  title = {On {{Expressive Power}} of {{Looped Transformers}}: {{Theoretical Analysis}} and {{Enhancement}} via {{Timestep Encoding}}},
  shorttitle = {On {{Expressive Power}} of {{Looped Transformers}}},
  author = {Xu, Kevin and Sato, Issei},
  year = 2025,
  month = jun,
  number = {arXiv:2410.01405},
  eprint = {2410.01405},
  primaryclass = {cs},
  publisher = {arXiv},
  doi = {10.48550/arXiv.2410.01405},
  urldate = {2025-12-22},
  archiveprefix = {arXiv},
  langid = {english}
}

@inproceedings{
elbayadDepthAdaptiveTransformer2020,
title={Depth-Adaptive Transformer},
author={Maha Elbayad and Jiatao Gu and Edouard Grave and Michael Auli},
booktitle={International Conference on Learning Representations},
year={2020},
url={https://openreview.net/forum?id=SJg7KhVKPH}
}

@misc{
raposoMixtureofDepthsDynamicallyAllocating2024,
      title={Mixture-of-Depths: Dynamically allocating compute in transformer-based language models}, 
      author={David Raposo and Sam Ritter and Blake Richards and Timothy Lillicrap and Peter Conway Humphreys and Adam Santoro},
      year={2024},
      eprint={2404.02258},
      archivePrefix={arXiv},
      primaryClass={cs.LG},
      url={https://arxiv.org/abs/2404.02258}, 
}

@inproceedings{
elhoushiLayerSkipEnablingEarly2024,
   title={LayerSkip: Enabling Early Exit Inference and Self-Speculative Decoding},
   url={http://dx.doi.org/10.18653/v1/2024.acl-long.681},
   DOI={10.18653/v1/2024.acl-long.681},
   booktitle={Proceedings of the 62nd Annual Meeting of the Association for Computational Linguistics (Volume 1: Long Papers)},
   publisher={Association for Computational Linguistics},
   author={Elhoushi, Mostafa and Shrivastava, Akshat and Liskovich, Diana and Hosmer, Basil and Wasti, Bram and Lai, Liangzhen and Mahmoud, Anas and Acun, Bilge and Agarwal, Saurabh and Roman, Ahmed and Aly, Ahmed and Chen, Beidi and Wu, Carole-Jean},
   year={2024},
   pages={12622–12642} }

@misc{
saunshiReasoningLatentThoughts2025b,
      title={Reasoning with Latent Thoughts: On the Power of Looped Transformers}, 
      author={Nikunj Saunshi and Nishanth Dikkala and Zhiyuan Li and Sanjiv Kumar and Sashank J. Reddi},
      year={2025},
      eprint={2502.17416},
      archivePrefix={arXiv},
      primaryClass={cs.CL},
      url={https://arxiv.org/abs/2502.17416}, 
}

@misc{
anilPathIndependentEquilibrium,
      title={Path Independent Equilibrium Models Can Better Exploit Test-Time Computation}, 
      author={Cem Anil and Ashwini Pokle and Kaiqu Liang and Johannes Treutlein and Yuhuai Wu and Shaojie Bai and Zico Kolter and Roger Grosse},
      year={2022},
      eprint={2211.09961},
      archivePrefix={arXiv},
      primaryClass={cs.LG},
      url={https://arxiv.org/abs/2211.09961}, 
}

@misc{bai2019deepequilibriummodels,
      title={Deep Equilibrium Models}, 
      author={Shaojie Bai and J. Zico Kolter and Vladlen Koltun},
      year={2019},
      eprint={1909.01377},
      archivePrefix={arXiv},
      primaryClass={cs.LG},
      url={https://arxiv.org/abs/1909.01377}, 
}

@misc{hoffmann2022trainingcomputeoptimallargelanguage,
      title={Training Compute-Optimal Large Language Models}, 
      author={Jordan Hoffmann and Sebastian Borgeaud and Arthur Mensch and Elena Buchatskaya and Trevor Cai and Eliza Rutherford and Diego de Las Casas and Lisa Anne Hendricks and Johannes Welbl and Aidan Clark and Tom Hennigan and Eric Noland and Katie Millican and George van den Driessche and Bogdan Damoc and Aurelia Guy and Simon Osindero and Karen Simonyan and Erich Elsen and Jack W. Rae and Oriol Vinyals and Laurent Sifre},
      year={2022},
      eprint={2203.15556},
      archivePrefix={arXiv},
      primaryClass={cs.CL},
      url={https://arxiv.org/abs/2203.15556}, 
}

@misc{yang2024loopedtransformersbetterlearning,
      title={Looped Transformers are Better at Learning Learning Algorithms}, 
      author={Liu Yang and Kangwook Lee and Robert Nowak and Dimitris Papailiopoulos},
      year={2024},
      eprint={2311.12424},
      archivePrefix={arXiv},
      primaryClass={cs.LG},
      url={https://arxiv.org/abs/2311.12424}, 
}

@misc{
xiongLayerNormalizationTransformer2020,
title={On Layer Normalization in the Transformer Architecture},
author={Ruibin Xiong and Yunchang Yang and Di He and Kai Zheng and Shuxin Zheng and Huishuai Zhang and Yanyan Lan and Liwei Wang and Tie-Yan Liu},
year={2020},
url={https://openreview.net/forum?id=B1x8anVFPr}
}

@misc{vaswani2023attentionneed,
      title={Attention Is All You Need}, 
      author={Ashish Vaswani and Noam Shazeer and Niki Parmar and Jakob Uszkoreit and Llion Jones and Aidan N. Gomez and Lukasz Kaiser and Illia Polosukhin},
      year={2023},
      eprint={1706.03762},
      archivePrefix={arXiv},
      primaryClass={cs.CL},
      url={https://arxiv.org/abs/1706.03762}, 
}

@misc{gu2024mambalineartimesequencemodeling,
      title={Mamba: Linear-Time Sequence Modeling with Selective State Spaces}, 
      author={Albert Gu and Tri Dao},
      year={2024},
      eprint={2312.00752},
      archivePrefix={arXiv},
      primaryClass={cs.LG},
      url={https://arxiv.org/abs/2312.00752}, 
}

@misc{dao2024transformersssmsgeneralizedmodels,
      title={Transformers are SSMs: Generalized Models and Efficient Algorithms Through Structured State Space Duality}, 
      author={Tri Dao and Albert Gu},
      year={2024},
      eprint={2405.21060},
      archivePrefix={arXiv},
      primaryClass={cs.LG},
      url={https://arxiv.org/abs/2405.21060}, 
}

@misc{shazeer2020gluvariantsimprovetransformer,
      title={GLU Variants Improve Transformer}, 
      author={Noam Shazeer},
      year={2020},
      eprint={2002.05202},
      archivePrefix={arXiv},
      primaryClass={cs.LG},
      url={https://arxiv.org/abs/2002.05202}, 
}

@misc{su2023roformerenhancedtransformerrotary,
      title={RoFormer: Enhanced Transformer with Rotary Position Embedding}, 
      author={Jianlin Su and Yu Lu and Shengfeng Pan and Ahmed Murtadha and Bo Wen and Yunfeng Liu},
      year={2023},
      eprint={2104.09864},
      archivePrefix={arXiv},
      primaryClass={cs.CL},
      url={https://arxiv.org/abs/2104.09864}, 
}

@misc{zhang2019rootmeansquarelayer,
      title={Root Mean Square Layer Normalization}, 
      author={Biao Zhang and Rico Sennrich},
      year={2019},
      eprint={1910.07467},
      archivePrefix={arXiv},
      primaryClass={cs.LG},
      url={https://arxiv.org/abs/1910.07467}, 
}

@misc{takase2025spikemorestabilizingpretraining,
      title={Spike No More: Stabilizing the Pre-training of Large Language Models}, 
      author={Sho Takase and Shun Kiyono and Sosuke Kobayashi and Jun Suzuki},
      year={2025},
      eprint={2312.16903},
      archivePrefix={arXiv},
      primaryClass={cs.CL},
      url={https://arxiv.org/abs/2312.16903}, 
}

@misc{kingma2017adammethodstochasticoptimization,
      title={Adam: A Method for Stochastic Optimization}, 
      author={Diederik P. Kingma and Jimmy Ba},
      year={2017},
      eprint={1412.6980},
      archivePrefix={arXiv},
      primaryClass={cs.LG},
      url={https://arxiv.org/abs/1412.6980}, 
}

@misc{loshchilov2019decoupledweightdecayregularization,
      title={Decoupled Weight Decay Regularization}, 
      author={Ilya Loshchilov and Frank Hutter},
      year={2019},
      eprint={1711.05101},
      archivePrefix={arXiv},
      primaryClass={cs.LG},
      url={https://arxiv.org/abs/1711.05101}, 
}

@misc{everett2024scalingexponentsparameterizationsoptimizers,
      title={Scaling Exponents Across Parameterizations and Optimizers}, 
      author={Katie Everett and Lechao Xiao and Mitchell Wortsman and Alexander A. Alemi and Roman Novak and Peter J. Liu and Izzeddin Gur and Jascha Sohl-Dickstein and Leslie Pack Kaelbling and Jaehoon Lee and Jeffrey Pennington},
      year={2024},
      eprint={2407.05872},
      archivePrefix={arXiv},
      primaryClass={cs.LG},
      url={https://arxiv.org/abs/2407.05872}, 
}

@inproceedings{
wortsman2023stable,
title={Stable and low-precision training for large-scale vision-language models},
author={Mitchell Wortsman and Tim Dettmers and Luke Zettlemoyer and Ari S. Morcos and Ali Farhadi and Ludwig Schmidt},
booktitle={Thirty-seventh Conference on Neural Information Processing Systems},
year={2023},
url={https://openreview.net/forum?id=sqqASmpA2R}
}

@InProceedings{Zhai_2022_CVPR,
    author    = {Zhai, Xiaohua and Kolesnikov, Alexander and Houlsby, Neil and Beyer, Lucas},
    title     = {Scaling Vision Transformers},
    booktitle = {Proceedings of the IEEE/CVF Conference on Computer Vision and Pattern Recognition (CVPR)},
    month     = {June},
    year      = {2022},
    pages     = {12104-12113}
}

@InProceedings{pmlr-v202-geiping23a,
  title = 	 {Cramming: Training a Language Model on a single {GPU} in one day.},
  author =       {Geiping, Jonas and Goldstein, Tom},
  booktitle = 	 {Proceedings of the 40th International Conference on Machine Learning},
  pages = 	 {11117--11143},
  year = 	 {2023},
  editor = 	 {Krause, Andreas and Brunskill, Emma and Cho, Kyunghyun and Engelhardt, Barbara and Sabato, Sivan and Scarlett, Jonathan},
  volume = 	 {202},
  series = 	 {Proceedings of Machine Learning Research},
  month = 	 {23--29 Jul},
  publisher =    {PMLR},
  url = 	 {https://proceedings.mlr.press/v202/geiping23a.html}
}

@inproceedings{paperno2016lambada,
  title={The {L}{A}{M}{B}{A}{D}{A} Dataset: Word Prediction Requiring a Broad Discourse Context},
  author={Paperno, Denis and Kruszewski, Germ{\'a}n and Lazaridou, Angeliki and Pham, Ngoc-Quan and Bernardi, Raffaella and Pezzelle, Sandro and Baroni, Marco and Boleda, Gemma and Fern{\'a}ndez, Raquel},
  booktitle={Proceedings of the 54th Annual Meeting of the Association for Computational Linguistics},
  pages={1525--1534},
  year={2016}
}

@inproceedings{zellers2019hellaswag,
  title={Hella{S}wag: Can a Machine Really Finish Your Sentence?},
  author={Zellers, Rowan and Holtzman, Ari and Bisk, Yonatan and Farhadi, Ali and Choi, Yejin},
  booktitle ={Proceedings of the 57th Annual Meeting of the Association for Computational Linguistics},
  year={2019}
}

@article{sakaguchi2021winogrande,
  title={Winogrande: An Adversarial {W}inograd Schema Challenge at Scale},
  author={Sakaguchi, Keisuke and Bras, Ronan Le and Bhagavatula, Chandra and Choi, Yejin},
  journal={Communications of the ACM},
  volume={64},
  number={9},
  pages={99--106},
  year={2021},
  publisher={ACM New York, NY, USA}
}

@article{clark2018think,
  title={Think you have Solved Question Answering? Try {A}{R}{C}, the {A}{I}2 Reasoning Challenge},
  author={Clark, Peter and Cowhey, Isaac and Etzioni, Oren and Khot, Tushar and Sabharwal, Ashish and Schoenick, Carissa and Tafjord, Oyvind},
  journal={arXiv preprint arXiv:1803.05457},
  year={2018}
}

@inproceedings{bisk2020piqa,
  title={P{I}{Q}{A}: Reasoning about Physical Commonsense in Natural Language},
  author={Bisk, Yonatan and Zellers, Rowan and Gao, Jianfeng and Choi, Yejin and others},
  booktitle={Proceedings of the AAAI conference on Artificial Intelligence},
  volume={34},
  year={2020}
}

@misc{merity2016pointer,
    title={Pointer Sentinel Mixture Models},
    author={Stephen Merity and Caiming Xiong and James Bradbury and Richard Socher},
    year={2016},
    eprint={1609.07843},
    archivePrefix={arXiv},
    primaryClass={cs.CL}
}

@ARTICLE{1082819,
  author={Desoer, C. and Min-Yen Wu},
  journal={IEEE Transactions on Circuit Theory}, 
  title={Stability of Linear Time-Invariant Systems}, 
  year={1968},
  volume={15},
  number={3},
  pages={245-250},
  doi={10.1109/TCT.1968.1082819}}

@misc{Hinton2013TrainingRN,
  title={Training Recurrent Neural Networks},
  author={Geoffrey E. Hinton and Ilya Sutskever},
  year={2013},
  url={https://api.semanticscholar.org/CorpusID:61713861}
}

@misc{kaplan2020scalinglawsneurallanguage,
      title={Scaling Laws for Neural Language Models}, 
      author={Jared Kaplan and Sam McCandlish and Tom Henighan and Tom B. Brown and Benjamin Chess and Rewon Child and Scott Gray and Alec Radford and Jeffrey Wu and Dario Amodei},
      year={2020},
      eprint={2001.08361},
      archivePrefix={arXiv},
      primaryClass={cs.LG},
      url={https://arxiv.org/abs/2001.08361}, 
}

@misc{dettmers2023case4bitprecisionkbit,
      title={The case for 4-bit precision: k-bit Inference Scaling Laws}, 
      author={Tim Dettmers and Luke Zettlemoyer},
      year={2023},
      eprint={2212.09720},
      archivePrefix={arXiv},
      primaryClass={cs.LG},
      url={https://arxiv.org/abs/2212.09720}, 
}

@misc{lin2024awqactivationawareweightquantization,
      title={AWQ: Activation-aware Weight Quantization for LLM Compression and Acceleration}, 
      author={Ji Lin and Jiaming Tang and Haotian Tang and Shang Yang and Wei-Ming Chen and Wei-Chen Wang and Guangxuan Xiao and Xingyu Dang and Chuang Gan and Song Han},
      year={2024},
      eprint={2306.00978},
      archivePrefix={arXiv},
      primaryClass={cs.CL},
      url={https://arxiv.org/abs/2306.00978}, 
}

@misc{nanochat,
  author = {Andrej Karpathy},
  title = {nanochat: The best ChatGPT that \$100 can buy},
  year = {2025},
  publisher = {GitHub},
  url = {https://github.com/karpathy/nanochat}
}

@misc{touvron2023llamaopenefficientfoundation,
      title={LLaMA: Open and Efficient Foundation Language Models}, 
      author={Hugo Touvron and Thibaut Lavril and Gautier Izacard and Xavier Martinet and Marie-Anne Lachaux and Timothée Lacroix and Baptiste Rozière and Naman Goyal and Eric Hambro and Faisal Azhar and Aurelien Rodriguez and Armand Joulin and Edouard Grave and Guillaume Lample},
      year={2023},
      eprint={2302.13971},
      archivePrefix={arXiv},
      primaryClass={cs.CL},
      url={https://arxiv.org/abs/2302.13971}, 
}

@article{Csordas2024MoEUTMU,
  title={MoEUT: Mixture-of-Experts Universal Transformers},
  author={R'obert Csord'as and Kazuki Irie and J{\"u}rgen Schmidhuber and Christopher Potts and Christopher D. Manning},
  journal={ArXiv},
  year={2024},
  volume={abs/2405.16039},
  url={https://api.semanticscholar.org/CorpusID:270063139}
}

@article{Bae2024RelaxedRT,
  title={Relaxed Recursive Transformers: Effective Parameter Sharing with Layer-wise LoRA},
  author={Sangmin Bae and Adam Fisch and Hrayr Harutyunyan and Ziwei Ji and Seungyeon Kim and Tal Schuster},
  journal={ArXiv},
  year={2024},
  volume={abs/2410.20672},
  url={https://api.semanticscholar.org/CorpusID:273654907}
}

@misc{koishekenov2025encodethinkdecodescaling,
      title={Encode, Think, Decode: Scaling test-time reasoning with recursive latent thoughts}, 
      author={Yeskendir Koishekenov and Aldo Lipani and Nicola Cancedda},
      year={2025},
      eprint={2510.07358},
      archivePrefix={arXiv},
      primaryClass={cs.LG},
      url={https://arxiv.org/abs/2510.07358}, 
}

@inproceedings{
bai2022neural,
title={Neural Deep Equilibrium Solvers},
author={Shaojie Bai and Vladlen Koltun and J Zico Kolter},
booktitle={International Conference on Learning Representations},
year={2022},
url={https://openreview.net/forum?id=B0oHOwT5ENL}
}

@misc{wu2025parallellooptransformerefficient,
      title={Parallel Loop Transformer for Efficient Test-Time Computation Scaling}, 
      author={Bohong Wu and Mengzhao Chen and Xiang Luo and Shen Yan and Qifan Yu and Fan Xia and Tianqi Zhang and Hongrui Zhan and Zheng Zhong and Xun Zhou and Siyuan Qiao and Xingyan Bin},
      year={2025},
      eprint={2510.24824},
      archivePrefix={arXiv},
      primaryClass={cs.CL},
      url={https://arxiv.org/abs/2510.24824}, 
}

@misc{jacobs2026blockrecurrentdynamicsvisiontransformers,
      title={Block-Recurrent Dynamics in Vision Transformers}, 
      author={Mozes Jacobs and Thomas Fel and Richard Hakim and Alessandra Brondetta and Demba Ba and T. Andy Keller},
      year={2026},
      eprint={2512.19941},
      archivePrefix={arXiv},
      primaryClass={cs.CV},
      url={https://arxiv.org/abs/2512.19941}, 
}

@misc{alabdulmohsin2025recursiveinferencescalingwinning,
      title={Recursive Inference Scaling: A Winning Path to Scalable Inference in Language and Multimodal Systems}, 
      author={Ibrahim Alabdulmohsin and Xiaohua Zhai},
      year={2025},
      eprint={2502.07503},
      archivePrefix={arXiv},
      primaryClass={cs.AI},
      url={https://arxiv.org/abs/2502.07503}, 
}

@misc{jordan2024muon,
  author       = {Keller Jordan and Yuchen Jin and Vlado Boza and You Jiacheng and
                  Franz Cesista and Laker Newhouse and Jeremy Bernstein},
  title        = {Muon: An optimizer for hidden layers in neural networks},
  year         = {2024},
  url          = {https://kellerjordan.github.io/posts/muon/}
}

@misc{li2025datacomplmsearchgenerationtraining,
      title={DataComp-LM: In search of the next generation of training sets for language models}, 
      author={Jeffrey Li and Alex Fang and Georgios Smyrnis and Maor Ivgi and Matt Jordan and Samir Gadre and Hritik Bansal and Etash Guha and Sedrick Keh and Kushal Arora and Saurabh Garg and Rui Xin and Niklas Muennighoff and Reinhard Heckel and Jean Mercat and Mayee Chen and Suchin Gururangan and Mitchell Wortsman and Alon Albalak and Yonatan Bitton and Marianna Nezhurina and Amro Abbas and Cheng-Yu Hsieh and Dhruba Ghosh and Josh Gardner and Maciej Kilian and Hanlin Zhang and Rulin Shao and Sarah Pratt and Sunny Sanyal and Gabriel Ilharco and Giannis Daras and Kalyani Marathe and Aaron Gokaslan and Jieyu Zhang and Khyathi Chandu and Thao Nguyen and Igor Vasiljevic and Sham Kakade and Shuran Song and Sujay Sanghavi and Fartash Faghri and Sewoong Oh and Luke Zettlemoyer and Kyle Lo and Alaaeldin El-Nouby and Hadi Pouransari and Alexander Toshev and Stephanie Wang and Dirk Groeneveld and Luca Soldaini and Pang Wei Koh and Jenia Jitsev and Thomas Kollar and Alexandros G. Dimakis and Yair Carmon and Achal Dave and Ludwig Schmidt and Vaishaal Shankar},
      year={2025},
      eprint={2406.11794},
      archivePrefix={arXiv},
      primaryClass={cs.LG},
      url={https://arxiv.org/abs/2406.11794}, 
}

@misc{penedo2024finewebdatasetsdecantingweb,
      title={The FineWeb Datasets: Decanting the Web for the Finest Text Data at Scale}, 
      author={Guilherme Penedo and Hynek Kydlíček and Loubna Ben allal and Anton Lozhkov and Margaret Mitchell and Colin Raffel and Leandro Von Werra and Thomas Wolf},
      year={2024},
      eprint={2406.17557},
      archivePrefix={arXiv},
      primaryClass={cs.CL},
      url={https://arxiv.org/abs/2406.17557}, 
}

@misc{rajpurkar2016squad100000questionsmachine,
      title={SQuAD: 100,000+ Questions for Machine Comprehension of Text}, 
      author={Pranav Rajpurkar and Jian Zhang and Konstantin Lopyrev and Percy Liang},
      year={2016},
      eprint={1606.05250},
      archivePrefix={arXiv},
      primaryClass={cs.CL},
      url={https://arxiv.org/abs/1606.05250}, 
}

@misc{reddy2019coqaconversationalquestionanswering,
      title={CoQA: A Conversational Question Answering Challenge}, 
      author={Siva Reddy and Danqi Chen and Christopher D. Manning},
      year={2019},
      eprint={1808.07042},
      archivePrefix={arXiv},
      primaryClass={cs.CL},
      url={https://arxiv.org/abs/1808.07042}, 
}

@article{lbfgs,
 ISSN = {00255718, 10886842},
 URL = {http://www.jstor.org/stable/2006193},
 author = {Jorge Nocedal},
 journal = {Mathematics of Computation},
 number = {151},
 pages = {773--782},
 publisher = {American Mathematical Society},
 title = {Updating Quasi-Newton Matrices with Limited Storage},
 urldate = {2026-03-12},
 volume = {35},
 year = {1980}
}

@article{huber,
author = {Peter J. Huber},
title = {{Robust Estimation of a Location Parameter}},
volume = {35},
journal = {The Annals of Mathematical Statistics},
number = {1},
publisher = {Institute of Mathematical Statistics},
pages = {73 -- 101},
year = {1964},
doi = {10.1214/aoms/1177703732},
URL = {https://doi.org/10.1214/aoms/1177703732}
}

@misc{chowdhery2022palmscalinglanguagemodeling,
      title={PaLM: Scaling Language Modeling with Pathways}, 
      author={Aakanksha Chowdhery and Sharan Narang and Jacob Devlin and Maarten Bosma and Gaurav Mishra and Adam Roberts and Paul Barham and Hyung Won Chung and Charles Sutton and Sebastian Gehrmann and Parker Schuh and Kensen Shi and Sasha Tsvyashchenko and Joshua Maynez and Abhishek Rao and Parker Barnes and Yi Tay and Noam Shazeer and Vinodkumar Prabhakaran and Emily Reif and Nan Du and Ben Hutchinson and Reiner Pope and James Bradbury and Jacob Austin and Michael Isard and Guy Gur-Ari and Pengcheng Yin and Toju Duke and Anselm Levskaya and Sanjay Ghemawat and Sunipa Dev and Henryk Michalewski and Xavier Garcia and Vedant Misra and Kevin Robinson and Liam Fedus and Denny Zhou and Daphne Ippolito and David Luan and Hyeontaek Lim and Barret Zoph and Alexander Spiridonov and Ryan Sepassi and David Dohan and Shivani Agrawal and Mark Omernick and Andrew M. Dai and Thanumalayan Sankaranarayana Pillai and Marie Pellat and Aitor Lewkowycz and Erica Moreira and Rewon Child and Oleksandr Polozov and Katherine Lee and Zongwei Zhou and Xuezhi Wang and Brennan Saeta and Mark Diaz and Orhan Firat and Michele Catasta and Jason Wei and Kathy Meier-Hellstern and Douglas Eck and Jeff Dean and Slav Petrov and Noah Fiedel},
      year={2022},
      eprint={2204.02311},
      archivePrefix={arXiv},
      primaryClass={cs.CL},
      url={https://arxiv.org/abs/2204.02311}, 
}

@misc{olsson2022incontextlearninginductionheads,
      title={In-context Learning and Induction Heads}, 
      author={Catherine Olsson and Nelson Elhage and Neel Nanda and Nicholas Joseph and Nova DasSarma and Tom Henighan and Ben Mann and Amanda Askell and Yuntao Bai and Anna Chen and Tom Conerly and Dawn Drain and Deep Ganguli and Zac Hatfield-Dodds and Danny Hernandez and Scott Johnston and Andy Jones and Jackson Kernion and Liane Lovitt and Kamal Ndousse and Dario Amodei and Tom Brown and Jack Clark and Jared Kaplan and Sam McCandlish and Chris Olah},
      year={2022},
      eprint={2209.11895},
      archivePrefix={arXiv},
      primaryClass={cs.LG},
      url={https://arxiv.org/abs/2209.11895}, 
}

@misc{zhang2024relu2winsdiscoveringefficient,
      title={ReLU{$^2$} Wins: Discovering Efficient Activation Functions for Sparse LLMs},
      author={Zhengyan Zhang and Yixin Song and Guanghui Yu and Xu Han and Yankai Lin and Chaojun Xiao and Chenyang Song and Zhiyuan Liu and Zeyu Mi and Maosong Sun},
      year={2024},
      eprint={2402.03804},
      archivePrefix={arXiv},
      primaryClass={cs.LG},
      url={https://arxiv.org/abs/2402.03804},
}

@misc{tian2023resformerscalingvitsmultiresolution,
      title={ResFormer: Scaling ViTs with Multi-Resolution Training}, 
      author={Rui Tian and Zuxuan Wu and Qi Dai and Han Hu and Yu Qiao and Yu-Gang Jiang},
      year={2023},
      eprint={2212.00776},
      archivePrefix={arXiv},
      primaryClass={cs.CV},
      url={https://arxiv.org/abs/2212.00776}, 
}

@misc{henry2020querykeynormalizationtransformers,
      title={Query-Key Normalization for Transformers}, 
      author={Alex Henry and Prudhvi Raj Dachapally and Shubham Pawar and Yuxuan Chen},
      year={2020},
      eprint={2010.04245},
      archivePrefix={arXiv},
      primaryClass={cs.CL},
      url={https://arxiv.org/abs/2010.04245}, 
}

@misc{elfwing2017sigmoidweightedlinearunitsneural,
      title={Sigmoid-Weighted Linear Units for Neural Network Function Approximation in Reinforcement Learning}, 
      author={Stefan Elfwing and Eiji Uchibe and Kenji Doya},
      year={2017},
      eprint={1702.03118},
      archivePrefix={arXiv},
      primaryClass={cs.LG},
      url={https://arxiv.org/abs/1702.03118}, 
}

@article{radford2019language,
  title={Language Models are Unsupervised Multitask Learners},
  author={Radford, Alec and Wu, Jeff and Child, Rewon and Luan, David and Amodei, Dario and Sutskever, Ilya},
  year={2019}
}

@misc{amsel2025polarexpressoptimalmatrix,
      title={The Polar Express: Optimal Matrix Sign Methods and Their Application to the Muon Algorithm}, 
      author={Noah Amsel and David Persson and Christopher Musco and Robert M. Gower},
      year={2025},
      eprint={2505.16932},
      archivePrefix={arXiv},
      primaryClass={cs.LG},
      url={https://arxiv.org/abs/2505.16932}, 
}

@misc{si2025adamuonadaptivemuonoptimizer,
      title={AdaMuon: Adaptive Muon Optimizer}, 
      author={Chongjie Si and Debing Zhang and Wei Shen},
      year={2025},
      eprint={2507.11005},
      archivePrefix={arXiv},
      primaryClass={cs.LG},
      url={https://arxiv.org/abs/2507.11005}, 
}

@misc{chen2026cautiousweightdecay,
      title={Cautious Weight Decay}, 
      author={Lizhang Chen and Jonathan Li and Kaizhao Liang and Baiyu Su and Cong Xie and Nuo Wang Pierse and Chen Liang and Ni Lao and Qiang Liu},
      year={2026},
      eprint={2510.12402},
      archivePrefix={arXiv},
      primaryClass={cs.LG},
      url={https://arxiv.org/abs/2510.12402}, 
}

@misc{ding2024fewertruncationsimprovelanguage,
      title={Fewer Truncations Improve Language Modeling}, 
      author={Hantian Ding and Zijian Wang and Giovanni Paolini and Varun Kumar and Anoop Deoras and Dan Roth and Stefano Soatto},
      year={2024},
      eprint={2404.10830},
      archivePrefix={arXiv},
      primaryClass={cs.CL},
      url={https://arxiv.org/abs/2404.10830}, 
}

@misc{openai2024gpt4technicalreport,
      title={GPT-4 Technical Report}, 
      author={OpenAI},
      year={2024},
      eprint={2303.08774},
      archivePrefix={arXiv},
      primaryClass={cs.CL},
      url={https://arxiv.org/abs/2303.08774}, 
}

@incollection{wsc:2015,
title={The Winograd Schema Challenge and Reasoning about Correlation},
author={Daniel Bailey and Amelia Harrison and Yuliya Lierler and Vladimir Lifschitz and Julian Michael},
booktitle={Working Notes of the Symposium on Logical Formalizations of Commonsense Reasoning},
publisher={AAAI Press},
url="http://www.cs.utexas.edu/users/ai-lab?wsc15",
year={2015}
}

@misc{srivastava2023imitationgamequantifyingextrapolating,
      title={Beyond the Imitation Game: Quantifying and extrapolating the capabilities of language models}, 
    author = {Aarohi Srivastava and Abhinav Rastogi and Abhishek Rao and Abu Awal Md Shoeb and Abubakar Abid and Adam Fisch and Adam R. Brown and Adam Santoro and Aditya Gupta and Adrià Garriga-Alonso and Agnieszka Kluska and Aitor Lewkowycz and Akshat Agarwal and Alethea Power and Alex Ray and Alex Warstadt and Alexander W. Kocurek and Ali Safaya and Ali Tazarv and Alice Xiang and Alicia Parrish and Allen Nie and Aman Hussain and Amanda Askell and Amanda Dsouza et al.},
      year={2023},
      eprint={2206.04615},
      archivePrefix={arXiv},
      primaryClass={cs.CL},
      url={https://arxiv.org/abs/2206.04615}, 
}

@misc{kaggle200000Jeopardy,
  author = {kaggle200000Jeopardy},
  title  = {200,000+ {J}eopardy! {Q}uestions},
  year   = {2019},
  url    = {https://www.kaggle.com/datasets/tunguz/200000-jeopardy-questions}
}

@misc{hendrycks2021measuringmassivemultitasklanguage,
      title={Measuring Massive Multitask Language Understanding}, 
      author={Dan Hendrycks and Collin Burns and Steven Basart and Andy Zou and Mantas Mazeika and Dawn Song and Jacob Steinhardt},
      year={2021},
      eprint={2009.03300},
      archivePrefix={arXiv},
      primaryClass={cs.CY},
      url={https://arxiv.org/abs/2009.03300}, 
}

@inproceedings{gordon-etal-2012-semeval,
    title = "{S}em{E}val-2012 Task 7: Choice of Plausible Alternatives: An Evaluation of Commonsense Causal Reasoning",
    author = "Gordon, Andrew  and
      Kozareva, Zornitsa  and
      Roemmele, Melissa",
    editor = "Agirre, Eneko  and
      Bos, Johan  and
      Diab, Mona  and
      Manandhar, Suresh  and
      Marton, Yuval  and
      Yuret, Deniz",
    booktitle = "*{SEM} 2012: The First Joint Conference on Lexical and Computational Semantics {--} Volume 1: Proceedings of the main conference and the shared task, and Volume 2: Proceedings of the Sixth International Workshop on Semantic Evaluation ({S}em{E}val 2012)",
    month = "7-8 " # jun,
    year = "2012",
    address = "Montr{\'e}al, Canada",
    publisher = "Association for Computational Linguistics",
    url = "https://aclanthology.org/S12-1052/",
    pages = "394--398"
}

@misc{talmor2019commonsenseqaquestionansweringchallenge,
      title={CommonsenseQA: A Question Answering Challenge Targeting Commonsense Knowledge}, 
      author={Alon Talmor and Jonathan Herzig and Nicholas Lourie and Jonathan Berant},
      year={2019},
      eprint={1811.00937},
      archivePrefix={arXiv},
      primaryClass={cs.CL},
      url={https://arxiv.org/abs/1811.00937}, 
}

@misc{mihaylov2018suitarmorconductelectricity,
      title={Can a Suit of Armor Conduct Electricity? A New Dataset for Open Book Question Answering}, 
      author={Todor Mihaylov and Peter Clark and Tushar Khot and Ashish Sabharwal},
      year={2018},
      eprint={1809.02789},
      archivePrefix={arXiv},
      primaryClass={cs.CL},
      url={https://arxiv.org/abs/1809.02789}, 
}

@misc{sap2019socialiqacommonsensereasoningsocial,
      title={SocialIQA: Commonsense Reasoning about Social Interactions}, 
      author={Maarten Sap and Hannah Rashkin and Derek Chen and Ronan LeBras and Yejin Choi},
      year={2019},
      eprint={1904.09728},
      archivePrefix={arXiv},
      primaryClass={cs.CL},
      url={https://arxiv.org/abs/1904.09728}, 
}

@misc{zhong2023agieval, title={AGIEval: A Human-Centric Benchmark for Evaluating Foundation Models}, author={Wanjun Zhong and Ruixiang Cui and Yiduo Guo and Yaobo Liang and Shuai Lu and Yanlin Wang and Amin Saied and Weizhu Chen and Nan Duan}, year={2023}, eprint={2304.06364}, archivePrefix={arXiv}, primaryClass={cs.CL} }

@misc{zhong2021arlsat, title={AR-LSAT: Investigating Analytical Reasoning of Text}, author={Wanjun Zhong and Siyuan Wang and Duyu Tang and Zenan Xu and Daya Guo and Jiahai Wang and Jian Yin and Ming Zhou and Nan Duan}, year={2021}, eprint={2104.06598}, archivePrefix={arXiv}, primaryClass={cs.CL} }

@article{wang2022lsat, title={From lsat: The progress and challenges of complex reasoning}, author={Wang, Siyuan and Liu, Zhongkun and Zhong, Wanjun and Zhou, Ming and Wei, Zhongyu and Chen, Zhumin and Duan, Nan}, journal={IEEE/ACM Transactions on Audio, Speech, and Language Processing}, year={2022}, publisher={IEEE} }

@software{llmfoundry,
  author = {MosaicML},
  title = {llm-foundry: LLM training and evaluation framework},
  year = {2023},
  publisher = {GitHub},
  url = {https://github.com}
}

@misc{amini2019mathqa,
    title={MathQA: Towards Interpretable Math Word Problem Solving with Operation-Based Formalisms},
    author={Aida Amini and Saadia Gabriel and Peter Lin and Rik Koncel-Kedziorski and Yejin Choi and Hannaneh Hajishirzi},
    year={2019},
    eprint={1905.13319},
    archivePrefix={arXiv},
    primaryClass={cs.CL}
}

@misc{liu2020logiqa,
    title={LogiQA: A Challenge Dataset for Machine Reading Comprehension with Logical Reasoning},
    author={Jian Liu and Leyang Cui and Hanmeng Liu and Dandan Huang and Yile Wang and Yue Zhang},
    year={2020},
    eprint={2007.08124},
    archivePrefix={arXiv},
    primaryClass={cs.CL}
}

@inproceedings{clark2019boolq,
  title     = {BoolQ: Exploring the Surprising Difficulty of Natural Yes/No Questions},
  author    = {Clark, Christopher and Lee, Kenton and Chang, Ming-Wei and Kwiatkowski, Tom and Collins, Michael and Toutanova, Kristina},
  booktitle = {NAACL},
  year      = {2019},
}

@inproceedings{jin2019pubmedqa,
  title={PubMedQA: A Dataset for Biomedical Research Question Answering},
  author={Jin, Qiao and Dhingra, Bhuwan and Liu, Zhengping and Cohen, William and Lu, Xinghua},
  booktitle={Proceedings of the 2019 Conference on Empirical Methods in Natural Language Processing and the 9th International Joint Conference on Natural Language Processing (EMNLP-IJCNLP)},
  pages={2567--2577},
  year={2019}
}

@misc{rudinger2018genderbiascoreferenceresolution,
      title={Gender Bias in Coreference Resolution}, 
      author={Rachel Rudinger and Jason Naradowsky and Brian Leonard and Benjamin Van Durme},
      year={2018},
      eprint={1804.09301},
      archivePrefix={arXiv},
      primaryClass={cs.CL},
      url={https://arxiv.org/abs/1804.09301}, 
}

@misc{parrish2022bbqhandbuiltbiasbenchmark,
      title={BBQ: A Hand-Built Bias Benchmark for Question Answering}, 
      author={Alicia Parrish and Angelica Chen and Nikita Nangia and Vishakh Padmakumar and Jason Phang and Jana Thompson and Phu Mon Htut and Samuel R. Bowman},
      year={2022},
      eprint={2110.08193},
      archivePrefix={arXiv},
      primaryClass={cs.CL},
      url={https://arxiv.org/abs/2110.08193}, 
}

@misc{xu2019understandingimprovinglayernormalization,
      title={Understanding and Improving Layer Normalization}, 
      author={Jingjing Xu and Xu Sun and Zhiyuan Zhang and Guangxiang Zhao and Junyang Lin},
      year={2019},
      eprint={1911.07013},
      archivePrefix={arXiv},
      primaryClass={cs.LG},
      url={https://arxiv.org/abs/1911.07013}, 
}

@misc{moon2024lpulatencyoptimizedhighlyscalable,
      title={LPU: A Latency-Optimized and Highly Scalable Processor for Large Language Model Inference}, 
      author={Seungjae Moon and Jung-Hoon Kim and Junsoo Kim and Seongmin Hong and Junseo Cha and Minsu Kim and Sukbin Lim and Gyubin Choi and Dongjin Seo and Jongho Kim and Hunjong Lee and Hyunjun Park and Ryeowook Ko and Soongyu Choi and Jongse Park and Jinwon Lee and Joo-Young Kim},
      year={2024},
      eprint={2408.07326},
      archivePrefix={arXiv},
      primaryClass={cs.AR},
      url={https://arxiv.org/abs/2408.07326}, 
}

@misc{narayan2025minionscostefficientcollaborationondevice,
      title={Minions: Cost-efficient Collaboration Between On-device and Cloud Language Models}, 
      author={Avanika Narayan and Dan Biderman and Sabri Eyuboglu and Avner May and Scott Linderman and James Zou and Christopher Re},
      year={2025},
      eprint={2502.15964},
      archivePrefix={arXiv},
      primaryClass={cs.LG},
      url={https://arxiv.org/abs/2502.15964}, 
}
